\documentclass{article} 
\usepackage{iclr2023_conference,times}

\usepackage{amsmath,amsfonts,bm}

\def\eqref#1{equation~\ref{#1}}

\def\1{\bm{1}}

\def\rvx{{\mathbf{x}}}
\def\rvy{{\mathbf{y}}}

\DeclareMathAlphabet{\mathsfit}{\encodingdefault}{\sfdefault}{m}{sl}
\SetMathAlphabet{\mathsfit}{bold}{\encodingdefault}{\sfdefault}{bx}{n}

\def\gD{{\mathcal{D}}}

\def\gS{{\mathcal{S}}}
\def\gT{{\mathcal{T}}}

\def\gX{{\mathcal{X}}}

\newcommand{\E}{\mathbb{E}}
\newcommand{\Ls}{\mathcal{L}}
\newcommand{\R}{\mathbb{R}}

\DeclareMathOperator*{\argmin}{arg\,min}

\usepackage[colorlinks]{hyperref}
\usepackage{url}
\usepackage{graphicx}
\usepackage{caption}
\usepackage{subcaption}
\usepackage{xspace}
\usepackage{amsmath}
\usepackage{booktabs}
\usepackage{comment}
\usepackage{enumitem}
\usepackage[capitalise]{cleveref}
\usepackage{soul}
\usepackage{multirow}
\usepackage{authblk}
\usepackage{amssymb}
\usepackage{wrapfig}
\usepackage{pifont}
\usepackage[ruled]{algorithm2e}
\usepackage{url}
\usepackage{tcolorbox}

\def\tmp#1#2#3{%
  \definecolor{Hy#1color}{#2}{#3}%
  \hypersetup{#1color=Hy#1color}}
\tmp{link}{HTML}{800006}
\tmp{cite}{HTML}{2E7E2A}
\tmp{file}{HTML}{131877}
\tmp{url} {HTML}{8A0087}
\tmp{menu}{HTML}{727500}
\tmp{run} {HTML}{137776}
\def\tmp#1#2{%
  \colorlet{Hy#1bordercolor}{Hy#1color#2}%
  \hypersetup{#1bordercolor=Hy#1bordercolor}}
\tmp{link}{!60!white}
\tmp{cite}{!60!white}
\tmp{file}{!60!white}
\tmp{url} {!60!white}
\tmp{menu}{!60!white}
\tmp{run} {!60!white}

\usepackage{tikz}
\usetikzlibrary{decorations.pathreplacing, calligraphy}
\usetikzlibrary{shapes.callouts}
\usetikzlibrary{chains, fit, quotes, automata}
\usetikzlibrary{arrows}
\usetikzlibrary{matrix,positioning}
\usetikzlibrary{shapes.geometric, shapes.misc}
\usetikzlibrary{calc}

\newcommand{\method}{\texttt{RangeAugment}\xspace}

\definecolor{yesColor}{RGB}{51,160,44}
\definecolor{noColor}{RGB}{189,0,38}
\definecolor{qColor}{RGB}{253,141,60}

\definecolor{highlightColor}{RGB}{152, 204, 112}

\newcommand{\cmark}{\textcolor{yesColor}{\scalebox{2}{\ding{51}}}}%
\newcommand{\xmark}{\textcolor{noColor}{\scalebox{2}{\ding{55}}}}%
\newcommand{\qmark}{\textcolor{qColor}{\scalebox{2}{?}}}%
\newcommand{\xmarkApp}{\textcolor{noColor}{\scalebox{1}{\ding{55}}}}%

\newcommand{\cmarkA}{\textcolor{yesColor}{\scalebox{1}{\ding{51}}}}%
\newcommand{\xmarkA}{\textcolor{noColor}{\scalebox{1}{\ding{55}}}}%

\newtcolorbox{obsBox}[2][]{colback=green!5!white, colframe=green!75!black,fonttitle=\bfseries, colbacktitle=red!85!black,enhanced,
attach boxed title to top center={yshift=-2mm},
  title={#2},#1}

\title{\method: Efficient Online Augmentation with Range Learning}

\makeatletter
\renewcommand\AB@affilsepx{\quad\protect\Affilfont}
\makeatother

\author[1,$\ddagger$]{Sachin Mehta}
\author[2\thanks{Work done during summer internship at Apple.}]{Saeid Naderiparizi}
\author[1]{Fartash Faghri}
\author[1]{Maxwell Horton} 
\author[1]{Lailin Chen}
\author[1]{\\ Ali Farhadi}
\author[1]{Oncel Tuzel}
\author[1]{Mohammad Rastegari}

\affil[1]{Apple}
\affil[2]{University of British Columbia 
\protect \\
$^\ddagger$ Project lead and main contributor
\protect \\ 
\begin{tcolorbox}[width=4.1in,
                  colback=red!5!white,
                  colframe=red!75!black
                  ]
    \textbf{Source code}: \url{https://github.com/apple/ml-cvnets}
\end{tcolorbox}
}

\newcommand{\meanp}[2]{\E_{#1} \left[#2\right]}

\iclrfinalcopy
\begin{document}

\maketitle

\begin{table}[h!]
    \centering
    \begin{subtable}[b]{0.45\columnwidth}
    \centering
        \resizebox{!}{25px}{
            \begin{tabular}{lcccc}
                \toprule[1.5pt]
                \textbf{Model} & \textbf{Aug. method} & \textbf{Search space} & \textbf{Train time overhead$^\dagger$} & \textbf{Top-1} \\
                \midrule[1.25pt]
                \multirow{3}{*}{MobileNetv2-1.0} & AutoAugment & $O\left( \left(MNP\right)^{2L}\right)$ & $25 \% \pm 5 \%$ & 72.7\% \\
                 & RandAugment & $O(MN)$  & $25 \% \pm 5 \%$ & 72.1\% \\
                 & \method (Ours) & $O(1)$  & $2 \% \pm 1 \%$ & \textbf{73.0}\% \\
                \midrule
                \multirow{3}{*}{ResNet-101} & AutoAugment & $O\left( \left(MNP\right)^{2L}\right)$  & $25 \% \pm 5 \%$ & 81.4\% \\
                 & RandAugment & $O(MN)$  & $25 \% \pm 5 \%$ & 81.5\% \\
                 & \method (Ours) & $O(1)$  & $2\% \pm 1\%$ & \textbf{81.9}\% \\
                 \bottomrule[1.5pt]
            \end{tabular}
        }
        \caption{ImageNet classification}
        \label{tab:teaser_table_img1k}
    \end{subtable}
    \hfill
    \begin{subtable}[b]{0.5\columnwidth}
    \centering
    \resizebox{!}{25px}{
        \begin{tabular}{lcc}
            \toprule[1.5pt]
            \textbf{Backbone} & \textbf{\method} & \textbf{mIoU} \\
            \midrule[1.25pt]
            \multirow{2}{*}{MobileNetv2-1.0} & \xmark & 38.0\% \\
             & \cmark & \textbf{38.6}\% \\
             \midrule
             \multirow{2}{*}{ResNet-101} & \xmark & 45.7\% \\
             & \cmark & \textbf{46.5}\% \\
             \bottomrule[1.5pt]
        \end{tabular}
        }
        \caption{ADE20k semantic segmentation with DeepLabv3}
    \end{subtable}
    \vfill
    \begin{subtable}[b]{0.5\columnwidth}
    \centering
    \resizebox{!}{25px}{
        \begin{tabular}{lccc}
            \toprule[1.5pt]
            \textbf{Backbone} & \textbf{\method} & \textbf{BBox mAP} & \textbf{Seg. mAP} \\
            \midrule[1.25pt]
            \multirow{2}{*}{MobileNetv2-1.0} & \xmark & 37.2\% & 33.4\% \\
             & \cmark & \textbf{38.4}\% & \textbf{34.7}\% \\
             \midrule
             \multirow{2}{*}{ResNet-101} & \xmark & 45.6\% & 40.4\% \\
             & \cmark & \textbf{46.1}\% & \textbf{41.1}\% \\
             \bottomrule[1.5pt]
        \end{tabular}
        }
        \caption{MS-COCO object detection with Mask R-CNN}
    \end{subtable}
    \hfill
    \begin{subtable}[b]{0.45\columnwidth}
    \centering
    \resizebox{!}{25px}{
        \begin{tabular}{lccc}
            \toprule[1.5pt]
            \textbf{Model} & \textbf{Dataset size} & \textbf{\method} & \textbf{0-shot Top-1} \\
            \midrule[1.25pt]
            \multirow{3}{*}{CLIP ViT-B} & 302 M & \xmark & 67.1\% \\
             & 302 M & \cmark & \textbf{68.3}\% \\
             \midrule
             \multirow{3}{*}{CLIP ViT-H} & 2.0 B & \xmark & \textbf{78.0}\% \\
             & \textbf{1.1 B} & \cmark & 77.9\% \\
             \bottomrule[1.5pt]
        \end{tabular}
        }
        \caption{Foundation models}
    \end{subtable}
    \caption{\textbf{\method: Learning  \hl{model- and task-specific} augmentation policies \textcolor{cyan}{online} for visual recognition tasks \textcolor{highlightColor}{efficiently} \textcolor{orange}{at scale}}. Here, $M$, $N$, $P$, and $L$ denotes the number of magnitude bins, augmentation operations, probability bins, and augmentation policies respectively. $^\dagger$Training time overhead is computed with respect to the baseline model trained without any augmentation.}
    \label{tab:teaser_table}
\end{table}

\begin{abstract}
State-of-the-art automatic augmentation methods (e.g., AutoAugment and RandAugment) for visual recognition tasks diversify training data using a large set of augmentation operations. The range of magnitudes of many augmentation operations (e.g., brightness and contrast) is continuous. Therefore, to make search computationally tractable, these methods use \emph{fixed} and \emph{manually-defined} magnitude ranges for each operation, which may lead to sub-optimal policies. To answer the open question on the importance of magnitude ranges for each augmentation operation,  we introduce \method that allows us to efficiently learn the range of magnitudes for individual as well as composite augmentation operations. \method uses an auxiliary loss based on image similarity as a measure to control the range of magnitudes of augmentation operations. As a result, \method has a single scalar parameter for search, image similarity, which we simply optimize via linear search. \method integrates seamlessly with any model and learns model- and task-specific augmentation policies. With extensive experiments on the ImageNet dataset across different networks, we show that \method achieves competitive performance to state-of-the-art automatic augmentation methods with 4-5 times fewer augmentation operations. Experimental results on semantic segmentation, object detection, foundation models and knowledge distillation further shows \method's effectiveness.
\end{abstract}

\section{Introduction}

Data augmentation is a widely used regularization method for training deep neural networks \citep{lecun1998gradient, krizhevsky2017imagenet, szegedy2015going, perez2017effectiveness, steiner2021train}. These methods apply carefully designed augmentation (or image transformation) operations (e.g., color transforms) to increase the quantity and diversity of training data, which in turn helps improve the generalization ability of models. However, these methods rely heavily on expert knowledge and extensive trial-and-error experiments. 

\begin{figure}[b!]
    \centering
    \begin{subfigure}[b]{\columnwidth}
        \centering
        \resizebox{0.95\columnwidth}{!}{
            \tikzset{>=latex}

\tikzstyle{framebox} = [rectangle, minimum width=18em, minimum height=1cm, draw=black, fill=white, rounded corners, thick, line width=0.5mm]
\tikzstyle{textbox} = [rectangle, minimum width=15em, minimum height=1cm, draw=black, fill=gray!10, 
text width =17em, line width=0.25mm]
\tikzstyle{textboxPlain} = [rectangle, minimum width=18em, minimum height=1cm, draw=white, fill=white, text width =17em, line width=0.25mm]

\newcommand{\overviewStd}{
    \begin{tikzpicture}[%
    auto,
  ]
    \node[framebox, fill=purple!20] (ss) {\scalebox{1.75}{Search space}};
    \node[textbox, below=0cm of ss] () {
        Set of augmentations $\mathcal{T}$ with their parameters
        \begin{itemize}[leftmargin=*]
            \item Probability of applying augmentation operation
            \item Augmentation magnitude, etc.
        \end{itemize}
    };
    \node[framebox, right=2cm of ss, fill=cyan!20] (sm) {\scalebox{1.75}{Search strategy}};
    \node[textbox, below=0cm of sm] () {
        \begin{itemize}[leftmargin=*]
            \item Reinforcement learning
            \item Grid search
            \item Bayesian optimization
            \item Gradient matching, etc.
        \end{itemize}
    };
    \node[framebox, right=5cm of sm, fill=yellow!20] (perf) {\scalebox{1.75}{Performance estimation}};
    \node[textbox, below=0cm of perf] () {
        Validation accuracy
    };

    \node[textboxPlain, right=1.5cm of perf] () {
        
        \begin{itemize}[leftmargin=*]
            \item[\cmark] Improves generalization ability
            \item[\qmark] Model-specific policy
            \item[\qmark] Task-specific policy
            \item[\xmark] Manual range of magnitudes
            \item[\xmark] Augmentation-dependent search space
        \end{itemize}
    };

    \draw[thick, line width=0.5mm] (sm.north east) edge[->, bend left] node[align=center, text width=6.5cm] {Train with a subset of augmentations with their parameters $\mathcal{S}$} (perf.north west);

    \draw[thick, line width=0.5mm] (perf.south west) edge[->, bend left] node [align=center, text width=2.5cm] {Performance feedback of $\mathcal{S}$} (sm.south east);
    
    \draw[->, thick, line width=0.5mm] (ss) -- (sm);
    \end{tikzpicture}
}

\newcommand{\overviewNA}{
    \begin{tikzpicture}[%
    auto,
  ]
    \node[framebox, fill=purple!20] (ss) {\scalebox{1.75}{Search space}};
    \node[textbox, below=0cm of ss] () {
        Target image similarity $\Delta$
    };
    \node[framebox, right=2cm of ss, fill=cyan!20] (sm) {\scalebox{1.75}{Search strategy}};
    \node[textbox, below=0cm of sm] () {
        Linear search over discrete values of $\Delta$
    };
    \node[framebox, right=5cm of sm, fill=yellow!20] (perf) {\scalebox{1.75}{Performance estimation}};
    \node[textbox, below=0cm of perf] () {
        Validation accuracy
    };
    \node[textboxPlain, right=1.5cm of perf] () {
        
        \begin{itemize}[leftmargin=*]
            \item[\cmark] Improves generalization ability
            \item[\cmark] Model-specific policy
            \item[\cmark] Task-specific policy
            \item[\cmark] Learned range of magnitudes
            \item[\cmark] Constant search space
        \end{itemize}
    };

    \draw[thick, line width=0.5mm] (sm.north east) edge[->, bend left] node [align=center, text width=6cm] {Train with all augmentations in $\mathcal{T}$ for given $\Delta$} (perf.north west);

    \draw[thick, line width=0.5mm] (perf.south west) edge[->, bend left] node [align=center, text width=2.5cm] {Performance feedback of $\Delta$} (sm.south east);

    \draw[->, thick, line width=0.5mm] (ss) -- (sm);
        
    \end{tikzpicture}
}\overviewStd
        }
        \caption{Standard automatic augmentation}
    \end{subfigure}
    \vfill
    \begin{subfigure}[b]{\columnwidth}
        \centering
        \resizebox{0.95\columnwidth}{!}{
            \tikzset{>=latex}

\tikzstyle{framebox} = [rectangle, minimum width=18em, minimum height=1cm, draw=black, fill=white, rounded corners, thick, line width=0.5mm]
\tikzstyle{textbox} = [rectangle, minimum width=15em, minimum height=1cm, draw=black, fill=gray!10, 
text width =17em, line width=0.25mm]
\tikzstyle{textboxPlain} = [rectangle, minimum width=18em, minimum height=1cm, draw=white, fill=white, text width =17em, line width=0.25mm]

\newcommand{\overviewStd}{
    \begin{tikzpicture}[%
    auto,
  ]
    \node[framebox, fill=purple!20] (ss) {\scalebox{1.75}{Search space}};
    \node[textbox, below=0cm of ss] () {
        Set of augmentations $\mathcal{T}$ with their parameters
        \begin{itemize}[leftmargin=*]
            \item Probability of applying augmentation operation
            \item Augmentation magnitude, etc.
        \end{itemize}
    };
    \node[framebox, right=2cm of ss, fill=cyan!20] (sm) {\scalebox{1.75}{Search strategy}};
    \node[textbox, below=0cm of sm] () {
        \begin{itemize}[leftmargin=*]
            \item Reinforcement learning
            \item Grid search
            \item Bayesian optimization
            \item Gradient matching, etc.
        \end{itemize}
    };
    \node[framebox, right=5cm of sm, fill=yellow!20] (perf) {\scalebox{1.75}{Performance estimation}};
    \node[textbox, below=0cm of perf] () {
        Validation accuracy
    };

    \node[textboxPlain, right=1.5cm of perf] () {
        
        \begin{itemize}[leftmargin=*]
            \item[\cmark] Improves generalization ability
            \item[\qmark] Model-specific policy
            \item[\qmark] Task-specific policy
            \item[\xmark] Manual range of magnitudes
            \item[\xmark] Augmentation-dependent search space
        \end{itemize}
    };

    \draw[thick, line width=0.5mm] (sm.north east) edge[->, bend left] node[align=center, text width=6.5cm] {Train with a subset of augmentations with their parameters $\mathcal{S}$} (perf.north west);

    \draw[thick, line width=0.5mm] (perf.south west) edge[->, bend left] node [align=center, text width=2.5cm] {Performance feedback of $\mathcal{S}$} (sm.south east);
    
    \draw[->, thick, line width=0.5mm] (ss) -- (sm);
    \end{tikzpicture}
}

\newcommand{\overviewNA}{
    \begin{tikzpicture}[%
    auto,
  ]
    \node[framebox, fill=purple!20] (ss) {\scalebox{1.75}{Search space}};
    \node[textbox, below=0cm of ss] () {
        Target image similarity $\Delta$
    };
    \node[framebox, right=2cm of ss, fill=cyan!20] (sm) {\scalebox{1.75}{Search strategy}};
    \node[textbox, below=0cm of sm] () {
        Linear search over discrete values of $\Delta$
    };
    \node[framebox, right=5cm of sm, fill=yellow!20] (perf) {\scalebox{1.75}{Performance estimation}};
    \node[textbox, below=0cm of perf] () {
        Validation accuracy
    };
    \node[textboxPlain, right=1.5cm of perf] () {
        
        \begin{itemize}[leftmargin=*]
            \item[\cmark] Improves generalization ability
            \item[\cmark] Model-specific policy
            \item[\cmark] Task-specific policy
            \item[\cmark] Learned range of magnitudes
            \item[\cmark] Constant search space
        \end{itemize}
    };

    \draw[thick, line width=0.5mm] (sm.north east) edge[->, bend left] node [align=center, text width=6cm] {Train with all augmentations in $\mathcal{T}$ for given $\Delta$} (perf.north west);

    \draw[thick, line width=0.5mm] (perf.south west) edge[->, bend left] node [align=center, text width=2.5cm] {Performance feedback of $\Delta$} (sm.south east);

    \draw[->, thick, line width=0.5mm] (ss) -- (sm);
        
    \end{tikzpicture}
}\overviewNA
        }
        \caption{\method}
    \end{subfigure}
    \caption{\textbf{\method vs. standard automatic augmentation methods}. \method's search space is independent of augmentation parameters, allowing us to learn model- and task-specific policies in a constant time.}
    \label{fig:standard_vs_rangeAug}
\end{figure}
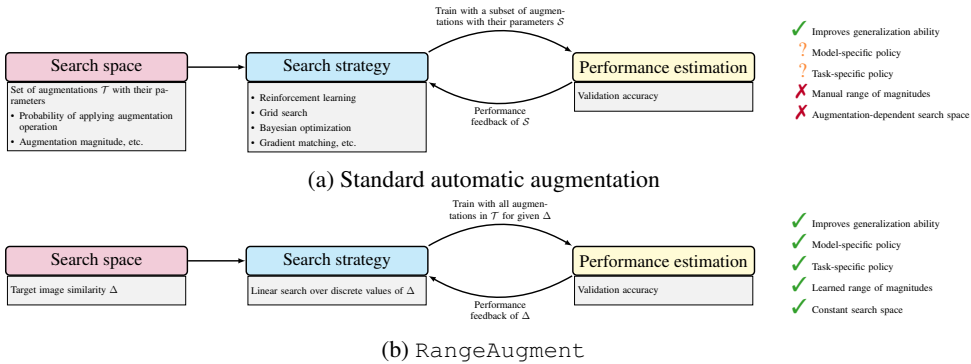

Recently, automatic augmentation methods have gained attention because of their ability to search for augmentation policy (e.g., combinations of different augmentation operations) that maximizes validation performance \citep{cubuk2019autoaugment, cubuk2020randaugment, lim2019fast, hataya2020faster, zheng2021deep}. In general, most augmentation operations (e.g., brightness and contrast) are parameterized by two variables: (1) the probability of applying them and (2) their range of magnitudes. These methods take a set of augmentation operations with a \emph{fixed} (often discretized) range of magnitudes as an input, and produce a policy of applying some or all augmentation operations along with their parameters (\cref{fig:standard_vs_rangeAug}). As an example, AutoAugment \citep{cubuk2019autoaugment} discretizes the range of magnitudes and probabilities of 16 augmentation operations, and searches for sub-policies (i.e., compositions of two augmentation operations along with their probability and magnitude) in a space of about $10^{32}$ possible combinations. These methods empirically show that automatic augmentation policies help improve performance of downstream networks. For example, AutoAugment improves the validation top-1 accuracy of ResNet-50 \citep{he2016deep} by about 1.3\% on the ImageNet dataset \citep{deng2009imagenet}. In other words, these methods underline the importance of automatic composition of augmentation operations in improving validation performance. However, policies generated using these networks may be sub-optimal because they use hand-crafted magnitude ranges. The importance of magnitude ranges for each augmentation operation is still an open question. An obstacle in answering this question is the range of magnitudes for most augmentation operations is continuous, which makes the search computationally intractable.

\begin{figure}[t!]
    \centering
    \begin{subfigure}[b]{0.49\columnwidth}
        \centering
        \resizebox{!}{49px}{
        \begin{tabular}{ccc}
            \toprule[1pt]
            \textbf{Brightness} & \textbf{Contrast} & \textbf{Noise} \\
            \midrule[0.75pt]
            \includegraphics[width=0.33\columnwidth]{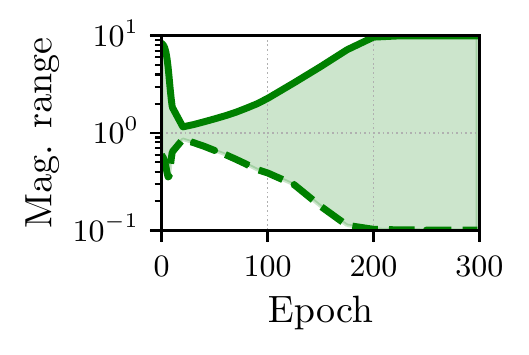} & \includegraphics[width=0.33\columnwidth]{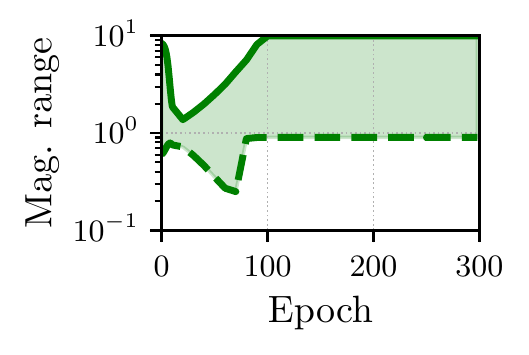}
             & \includegraphics[width=0.33\columnwidth]{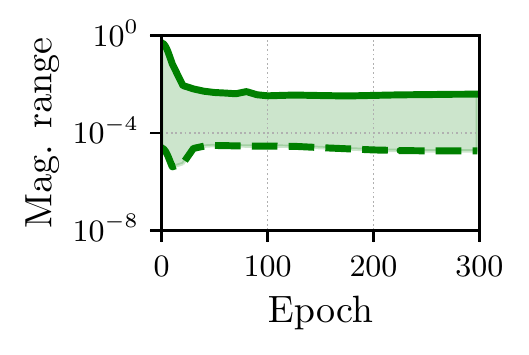} \\
             \textbf{Model}: SwinTransformer & \textbf{Task}: Classification & \textbf{Dataset}: ImageNet \\
             \midrule[0.75pt]
             \includegraphics[width=0.33\columnwidth]{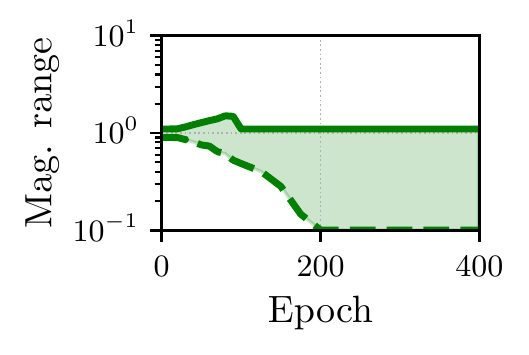}& \includegraphics[width=0.33\columnwidth]{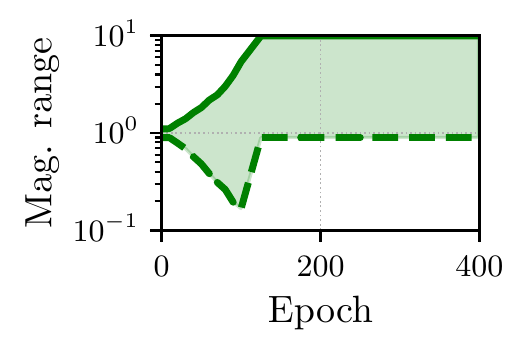}
             & \includegraphics[width=0.33\columnwidth]{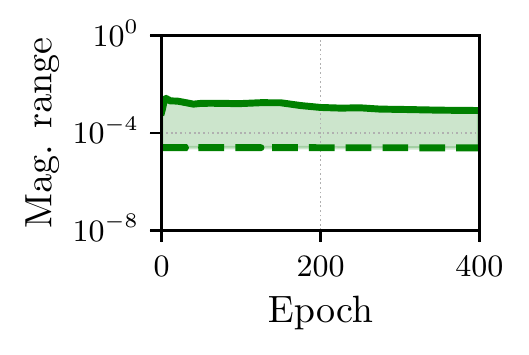} \\
             \textbf{Model}: EfficientNet & \textbf{Task}: Classification & \textbf{Dataset}: ImageNet \\
            \bottomrule[1pt]
        \end{tabular}
        }
        \caption{Model-specific policy}
        \label{fig:learned_range_diff_arch}
    \end{subfigure}
    \hfill
    \begin{subfigure}[b]{0.49\columnwidth}
        \centering
        \resizebox{!}{49px}{
        \begin{tabular}{ccc}
            \toprule[1pt]
            \textbf{Brightness} & \textbf{Contrast} & \textbf{Noise} \\
            \midrule[0.75pt]
            \includegraphics[width=0.33\columnwidth]{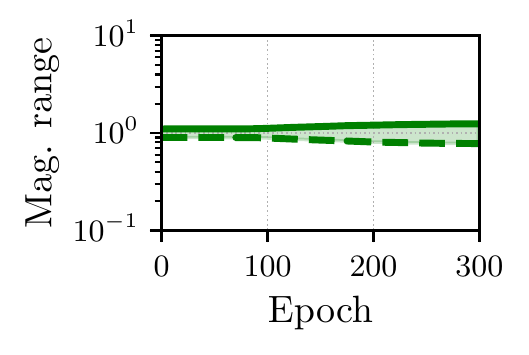} & \includegraphics[width=0.33\columnwidth]{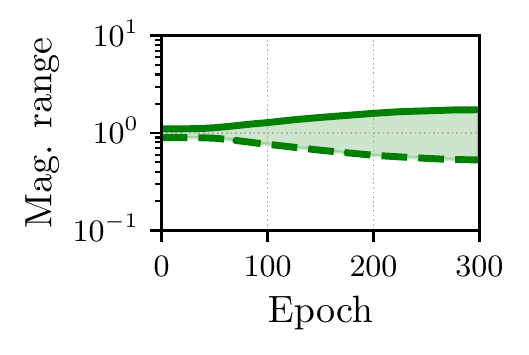}
             & \includegraphics[width=0.33\columnwidth]{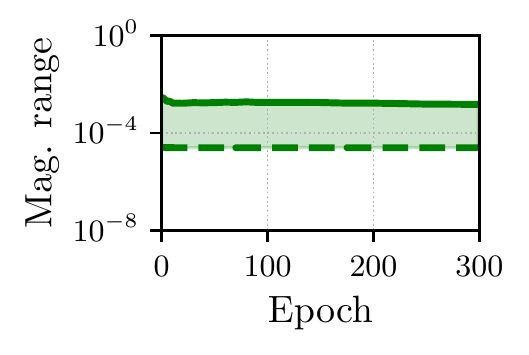} \\
             \textbf{Model}: MobileNetv3 & \textbf{Task}: Classification & \textbf{Dataset}: ImageNet \\
             \midrule[0.75pt]
             \includegraphics[width=0.33\columnwidth]{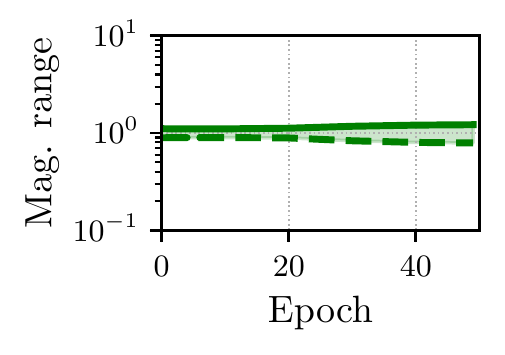}& \includegraphics[width=0.33\columnwidth]{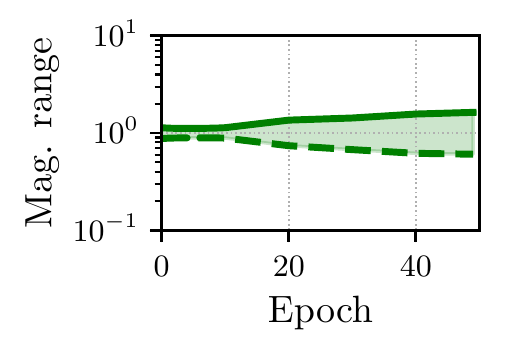}
             & \includegraphics[width=0.33\columnwidth]{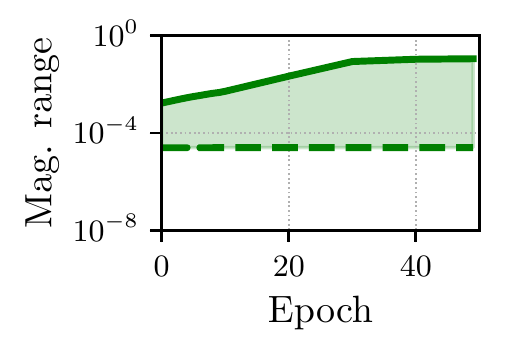} \\
             \textbf{Model}: MobileNetv3 & \textbf{Task}: Segmentation & \textbf{Dataset}: ADE20k \\
            \bottomrule[1pt]
        \end{tabular}
        }
        \caption{Task-specific policy}
        \label{fig:neural_aug_data_policy}
    \end{subfigure}
    \caption{\textbf{\method learns model- and task-specific policies.} (a) shows magnitude ranges for two different models on the same task and dataset while (b) shows magnitude ranges for the same model on two different tasks. All models are trained end-to-end with \method. The target image similarity $\Delta$ (PSNR) is annealed from 40 to 5 in (a) and from 40 to 20 in (b). Training epochs and magnitude range (log scale) of augmentation operations are plotted in x- and y-axis respectively.}
    \label{fig:learn_model_task_policy}
\end{figure}

This paper introduces \method, a simple and efficient method to learn the range of magnitudes for each augmentation operation. Inspired by image similarity metrics \citep{hore2010image}, \method introduces an auxiliary augmentation loss that allows us to learn the range of magnitudes for each augmentation operation. We realize this by controlling the similarity between the input and the augmented image for a given model and task. Rather than directly specifying the parameters for each augmentation operation, \method takes a target image similarity value as an input. The loss function is then formulated as a combination of the empirical loss and an augmentation loss. The objective of the augmentation loss is to match the target image similarity value. Therefore, the search objective in \method is to find the target similarity value that provides a good trade-off between minimizing the augmentation loss (i.e., matching the target similarity value) and the empirical loss. As a result, the augmentation policy learning in \method reduces to searching for a single scalar parameter, \emph{target image similarity}, that maximizes the downstream model's validation performance. We search for this target image similarity value via linear search. Empirically, we observe that this trade-off between the augmentation and empirical loss leads to better generalization ability of downstream model. Compared to existing automatic augmentation methods that require a large set of augmentation operations (usually 14-16 operations), \method is able to achieve competitive performance with only three simple operations (brightness, contrast, and additive Gaussian noise). Because \method's search space is independent of augmentation parameters and is fully differentiable (\cref{fig:standard_vs_rangeAug}), it can be trained end-to-end with any downstream model to learn model- and task-specific policies (\cref{fig:learn_model_task_policy}).

We empirically demonstrate in \cref{sec:imagenet_neural_aug} that \method allows us to learn model-specific policies when trained end-to-end with downstream models on the ImageNet dataset (\cref{fig:learned_range_diff_arch}). Importantly, \method achieves competitive performance to existing automatic augmentation methods (e.g., AutoAugment) while having a constant search space (\cref{tab:teaser_table_img1k}). In \cref{sec:task_gen_na}, we apply \method to variety of visual recognition tasks, including semantic segmentation (\cref{ssec:semantic_seg}), object detection (\cref{ssec:maskrcnn_coco}), foundation models (\cref{ssec:clip}), and distillation (\cref{ssec:distillation}), to demonstrate its simplicity and seamless integration ability to different tasks. We further show that \method learn task-specific policies (\cref{fig:neural_aug_data_policy}). To the best of our knowledge, \method is the first automatic augmentation method that learns model- and task-specific magnitude range for each augmentation operation at scale.

\section{Related Work}
\label{sec:related_work}

Data augmentation combines different augmentation operations (e.g., random brightness, random contrast, random Gaussian noise, and data mixing) to synthesize additional training data. Traditional augmentation methods rely heavily on expert knowledge and extensive trial-and-error experiments. In practice, these manual augmentation methods have been used to train different models on a variety of datasets and tasks \citep[e.g.,][]{szegedy2015going, he2016deep, zhao2017pyramid, howard2019searching}. However, these policies may not be optimal for all models. 

Motivated by neural architecture search \citep{zoph2017neural},  recent methods focus on finding optimal augmentation policies automatically from data. AutoAugment formulates automatic augmentation as a reinforcement learning problem, and uses model's validation performance as a reward to find an  augmentation policy leading to optimal validation performance. Because AutoAugment searches for several augmentation policy parameters, the search space is enormous and computationally intractable on large datasets and models. Therefore, in practice, policies found for smaller datasets are transferred to larger datasets. Since then, many follow-up works have focused on reducing the search space while delivering a similar performance to AutoAugment \citep{ratner2017learning,lemley2017smart,lingchen2020uniformaugment,li2020differentiable,zheng2021deep, liu2021direct}. The first line of research reduces the search time by introducing different hyper-parameter optimization methods, including population-based training \citep{ho2019population}, density matching \citep{lim2019fast, hataya2020faster}, and gradient matching \citep{zheng2021deep}. The second line of research reduces the search space by making practical assumptions \citep{cubuk2020randaugment, muller2021trivialaugment}. For instance, RandAugment \citep{cubuk2020randaugment} applies two transforms randomly with uniform probability. With these assumptions, RandAugment reduces AutoAugment's search space from $10^{32}$ to $10^2$ while maintaining downstream networks performance. 

One common characteristic among these different automatic augmentation methods is that they use \emph{fixed} and \emph{manually-defined} range of magnitudes for different augmentation operations, and focus on diversifying training data by using a large set of augmentation operations (e.g., 14 transforms). This is because the range of magnitudes for most augmentation operations is continuous and large, and searching over this large range is practically infeasible. Unlike these works, \method focuses on \emph{learning the magnitude range} of each augmentation operation (\cref{fig:standard_vs_rangeAug,fig:learn_model_task_policy}). We show in \cref{sec:imagenet_neural_aug} that \method is able to learn model- and task-specific policies while delivering competitive performance to previous automatic augmentation methods across different models.

\section{\method}
\label{sec:neural_aug}
Existing automatic augmentation methods search for composite augmentations over a large set of augmentation operations, but each augmentation operation has a manually-defined range of magnitudes.  This paper introduces \method, a method for learning a range of magnitude for each augmentation operation (\cref{fig:overview}). \method uses image similarity between the input and augmented image to learn the range of magnitudes for each augmentation operation. In the rest of this section, we first formulate the problem (\cref{ssec:formulation}), then elaborate on \method's policy learning method (\cref{ssec:policy-learning}), then we give implementation details ( \cref{ssec:implementation-details}). 

\begin{figure}[t!]
    \centering
    \resizebox{0.55\columnwidth}{!}{
        \input{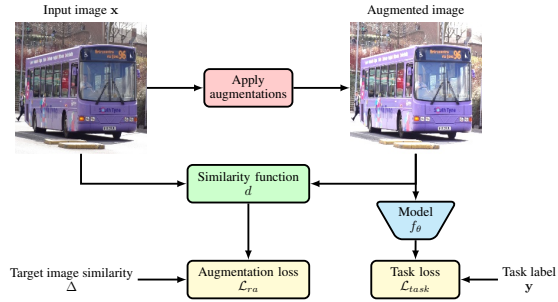}\rangeAugment
    }
    \caption{\textbf{\method: End-to-end learning of model- and task-specific augmentation policy.} The target image similarity value, $\Delta$, in \method controls the diversity of augmented samples. Therefore, the augmentation policy learning in \method reduces to linear search over discrete values of $\Delta$.}
    \label{fig:overview}
\end{figure}

\def\augop{T}
\def\setops{\gT}
\def\augment{\gS}
\def\magnitude{m}

\subsection{Problem Formulation}
\label{ssec:formulation}

Let $\setops = \{\augop_1, \cdots, \augop_N \}$ be a set of $N$ differentiable augmentation operations. Each augmentation operation $\augop \in \setops$ is parameterized by a scalar magnitude parameter $\magnitude \in \R$ such that ${T(\cdot; \magnitude): \gX \rightarrow \gX}$ is defined on the image space $\gX$. Let $\pi_\phi$ be an augmentation policy that defines a distribution over sub-policies $\augment \sim \pi_\phi$ in \method such that ${\augment = \{T_i(\cdot; \magnitude_i)\}_{i=1}^N}$.
A sub-policy $\gS$ applies augmentation operations to an input image $\rvx$ with uniform probability as
\begin{equation}
    \augment(\rvx) := \rvx^{(N)}, \quad
    \rvx^{(i)} = T_i(\rvx^{(i-1)}; \magnitude_i), \quad
    \quad \rvx^{(0)} = \rvx.
\end{equation}
For any given model and task, the goal of automatic augmentation is to find the augmentation policy $\pi_\phi$ that diversifies training data, and helps improve model's generalization ability the most. \method learns the range of magnitudes for each augmentation operation in $\setops$. Formally, the policy parameters in \method are $\phi = \{(a_i, b_i)\}_{i=1}^N$
and the magnitude parameter $m_i \sim U(a_i, b_i)$ for the $i$-th augmentation operation in a sub-policy $\augment = \{T_i(\cdot; m_i)\}_{i=1}^N$ is uniformly sampled, where $a_i \in \mathbb{R}$ and $b_i \in \mathbb{R}$ are learned parameters.

\subsection{Policy learning}
\label{ssec:policy-learning}
Diverse training data can be produced by using wider range of magnitudes, $(a_i, b_i)$, for each augmentation operation. However, directly searching for the optimal values of $(a_i, b_i)$ for each model and dataset is challenging because of its continuous nature. To address this, \method introduces an auxiliary loss which, in conjunction with the task-specific empirical loss, enables learning model-specific range of magnitudes for each augmentation operation in an end-to-end fashion. 

Let $d: \gX \times \gX \rightarrow \R$ be a differentiable image similarity function that measures the similarity between the input and the augmented image. To control the range of magnitudes for each augmentation operation, \method minimizes the distance between the expected value of $d(\rvx, \augment(\rvx))$ and a target image similarity value $\Delta \in \R$ using an augmentation loss function $\Ls_{\text{ra}}$ (e.g., smooth L1 loss or L2 loss). An example of $d$ and $\Delta$ are peak signal to noise ratio (PSNR) and target PSNR value respectively. When target PSNR value is small, the difference between the input and augmented image obtained after applying an augmentation operation (say brightness) will be large. In other words, for a smaller target PSNR value, the range of magnitudes for brightness operation will be wider and vice-versa.

For a given value of $\Delta$ and parameterized model $f_\theta$ with parameters $\theta$, the overall loss function to learn model- and task-specific augmentation policy is a weighted sum of the augmentation loss $\Ls_{\text{ra}}$ and the task-specific empirical loss $\mathcal{L}_{\text{task}}$: 
\begin{equation}
    \theta^*, \phi^* = \argmin_{\theta, \phi} \meanp{(\rvx, \rvy) \sim \gD_{\text{train}}}{
        \meanp{\augment \sim \pi_{\phi}}{
            \mathcal{L}_{\text{task}}(f_\theta(\augment(\rvx)), \rvy) + \lambda \Ls_{\text{ra}}(\rvx, \augment(\rvx); \Delta)
        }
    },
    \label{eq:optim_objective}
\end{equation}
where $\lambda$ and $\gD_{\text{train}}$ represent weight term and training set respectively. Note that, in \cref{eq:optim_objective}, re-parameterization trick on uniform distributions \citep{kingma2013auto} is applied to back-propagate through the expectation over $\gS \sim \pi_\phi$.

The $\Delta$ in \cref{eq:optim_objective} allows \method to control diversity of augmented samples. Therefore, the augmentation policy learning in \method reduces to searching a single scalar parameter, $\Delta$, that maximizes downstream model's validation performance. \method finds the optimal value of $\Delta$ using a linear search.

\subsection{Implementation details}
\label{ssec:implementation-details}

We use PSNR as the image similarity function $d$ in our experiments because it is (1) a standard image quality metric, (2) differentiable, and (3) fast to compute. Across different downstream networks, we observe a 0.5\%-3\% training overhead over the empirical risk minimization baseline. 

\begin{wrapfigure}[12]{r}{0.45\textwidth}
    \centering
    \resizebox{0.45\columnwidth}{!}{
    \begin{tabular}{ccc}
        \includegraphics[width=0.3\columnwidth]{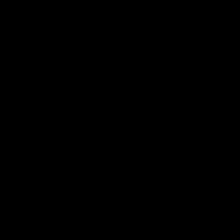} &
       \includegraphics[width=0.3\columnwidth]{figs/extreme_samples/orig.jpg}  & \includegraphics[width=0.3\columnwidth]{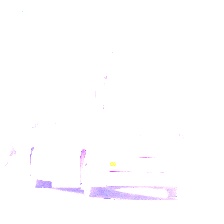} \\
        $\magnitude=0.1$ & $\magnitude=1.0$ & $\magnitude=10.0$ 
    \end{tabular}
    }
    \addtocounter{figure}{0} 
    \caption{Example outputs of brightness operation, $T(\rvx; m) = \magnitude \rvx $, at different values of magnitude parameter $\magnitude$. At extremes (i.e., $\magnitude=0.1$ or $10$), the bus is hardly identifiable.}
    \label{fig:extreme-augmentations}
\end{wrapfigure}

To find an optimal value of $\Delta$ in \cref{eq:optim_objective}, we study two approaches: (1) fixed target PSNR ($\Delta \in \{5, 10, 20, 30\}$) and (2) target PSNR with a curriculum, where the value of $\Delta$ is annealed from 40 to $\delta$ and $\delta \in \{5, 10, 20, 30\} $. The learned ranges of magnitudes, $(a, b)$, can scale beyond the image space (e.g., negative values) and result in training instability. To prevent this, we clip the range of magnitudes if they are beyond  extreme bounds of augmentation operations. We choose these extreme bounds such that objects in an image are hardly identifiable at or beyond the extreme points of the bounds  (see \cref{fig:extreme-augmentations}). Also, because the focus of \method is to learn the range of magnitudes for each augmentation operation, we apply all augmentation operations in $\mathcal{T}$ with uniform probability. We use smooth L1 loss as augmentation loss function $\Ls_{\text{ra}}$.

To demonstrate the importance of magnitude ranges, we study \method with three basic operations (brightness, contrast, and additive Gaussian noise), and show empirically in  \cref{sec:imagenet_neural_aug} that \method can achieve competitive performance to existing methods with 4 to 5 times fewer augmentation operations.

\section{Evaluating \method on Image Classification}
\label{sec:imagenet_neural_aug}
\method can learn model-specific augmentation policies. To evaluate this, we first study the importance of single and composite augmentation operations using ResNet-50 on the ImageNet dataset (\cref{ssec:neural_aug_op_char_res50}). We then study model-level generalization of \method (\cref{ssec:neural_aug_model_gen}).

\subsection{Experimental Set-up}
\label{ssec:exp_setup_imagenet}

\paragraph{Dataset} For image classification, we use the ImageNet dataset that has 1.28M training and 50k validation images spanning across 1000  categories. We use top-1 accuracy to measure performance.

\paragraph{Baseline models} To evaluate the effectiveness of \method, we study different CNN- and transformer-based models. We group these models into two categories based on their complexity: (1) \textbf{\textit{mobile:}} MobileNetv1 \citep{howard2017mobilenets}, MobileNetv2 \citep{sandler2018mobilenetv2}, MobileNetv3 \citep{howard2019searching}, and MobileViT \citep{mehta2021mobilevit} and (2) \textbf{\textit{non-mobile:}} ResNet-50 \citep{he2016deep}, ResNet-101, EfficientNet \citep{tan2019efficientnet}, and SwinTransformer \citep{liu2021swin}. We implement \method using the CVNets library \citep{mehta2022cvnets} and use their baseline model implementations and training recipes. Additional experiments, including the effect of individual and joint learning on \method's performance, along with training details are given in \cref{sec:appendix_ablations,sec:appendix_training_details} respectively.

\paragraph{Baseline augmentation methods} For an apples to apples comparison, each baseline model is trained with three different random seeds and the following baseline augmentation strategies: (1) \textbf{\textit{Baseline}} - standard Inception-style pre-processing (random resized cropping and random horizontal flipping) \citep{szegedy2015going}, (2) \textbf{\textit{RandAugment}} - Baseline pre-processing followed by RandAugment policy of \citet{cubuk2020randaugment}, (3) \textbf{\textit{AutoAugment}} - Baseline pre-processing followed by AutoAugment policy of \citet{cubuk2019autoaugment}, and (4) \textbf{\textit{\method}} - Baseline pre-processing following by the proposed augmentation policy in \cref{sec:neural_aug}. See \cref{ssec:sota_imagenet_perf} for comparison with other augmentation methods.

\begin{figure}[t!]
    \centering
    \begin{subfigure}[b]{\columnwidth}
        \includegraphics[width=\columnwidth]{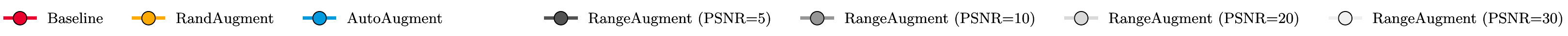}
    \end{subfigure}
    \vfill
    \begin{subfigure}[b]{0.32\columnwidth}
        \includegraphics[width=\columnwidth]{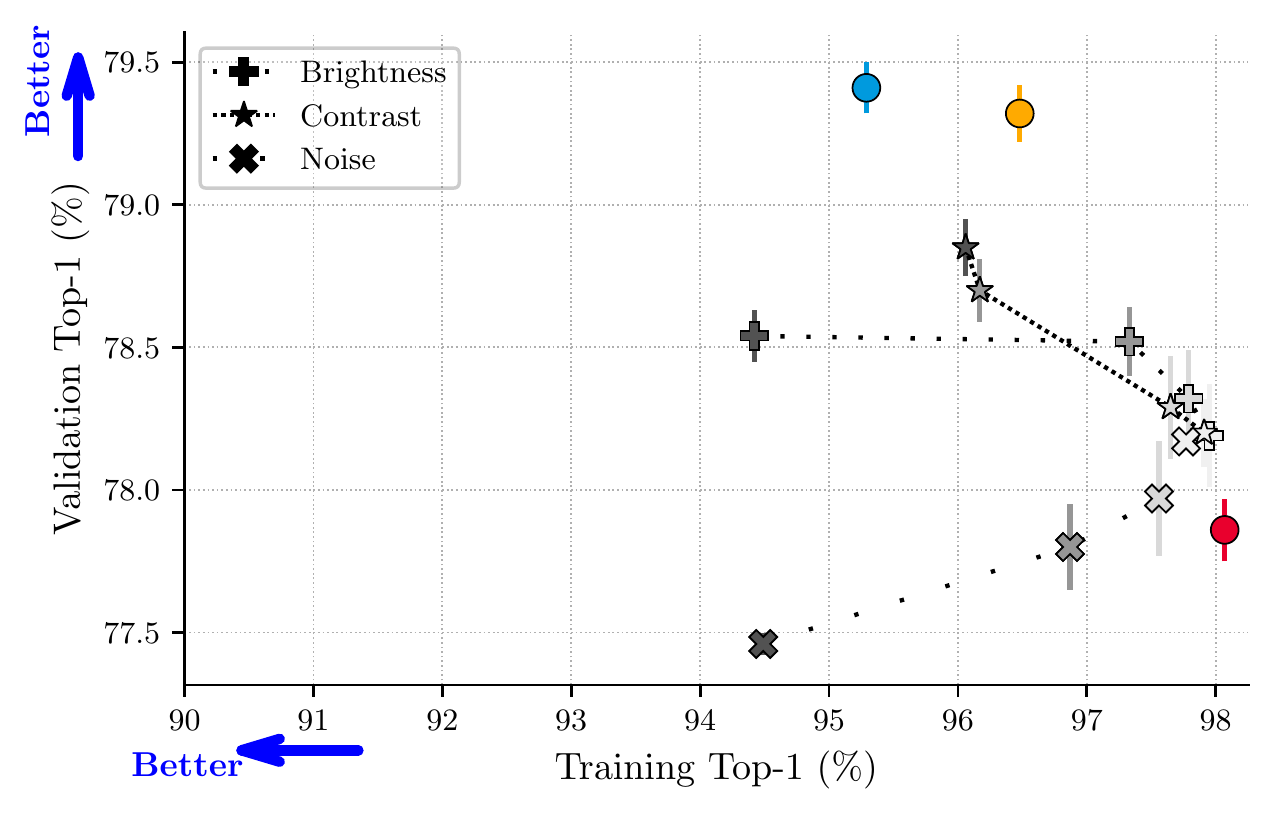}
        \caption{$N=1$}
        \label{fig:aug_effect_perf_char_n1}
    \end{subfigure}
    \hfill
    \begin{subfigure}[b]{0.32\columnwidth}
        \includegraphics[width=\columnwidth]{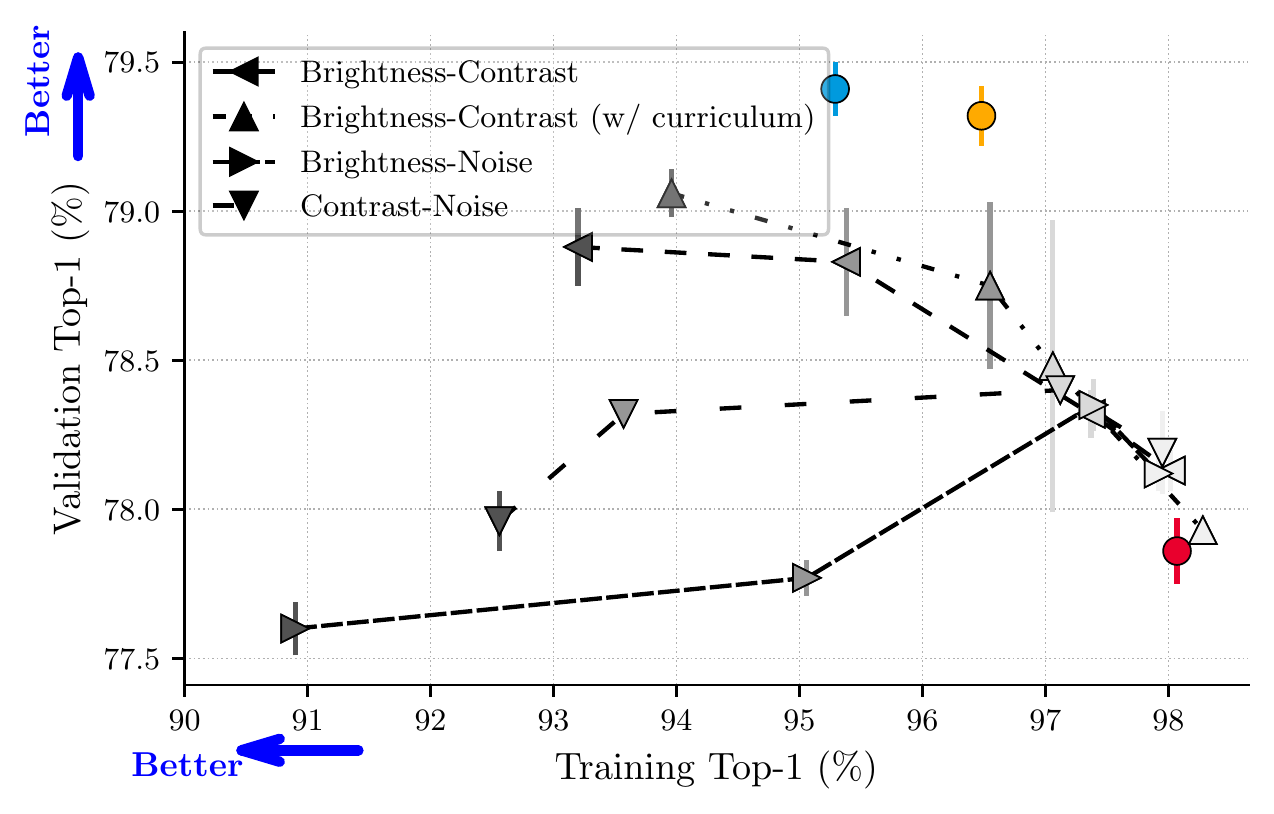}
        \caption{$N=2$}
        \label{fig:aug_effect_perf_char_n2}
    \end{subfigure}
    \hfill
    \begin{subfigure}[b]{0.32\columnwidth}
        \includegraphics[width=\columnwidth]{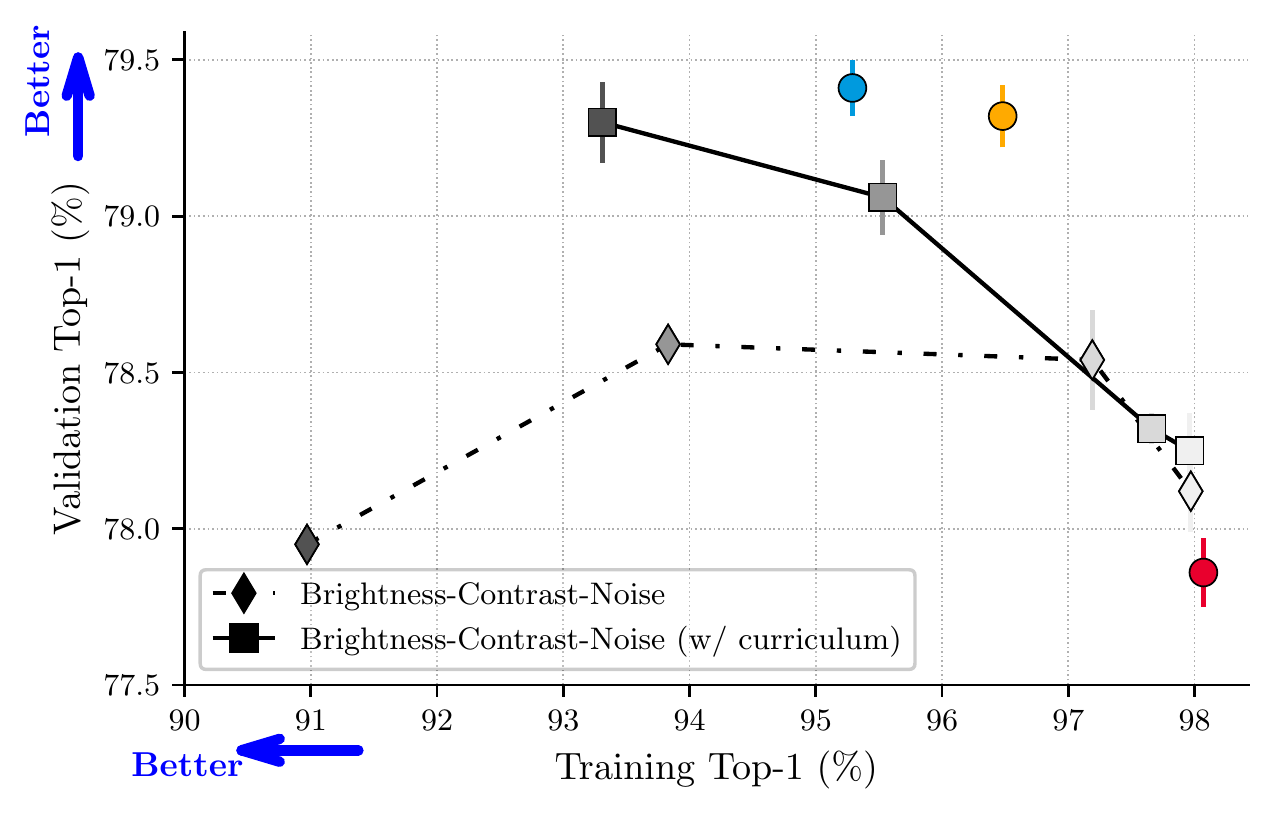}
        \caption{$N=3$}
        \label{fig:aug_effect_perf_char_n3}
    \end{subfigure}
    \caption{\textbf{Augmentation operation characterization for ResNet-50 using \method}. In (a-c), the performance of ResNet-50 on the ImageNet dataset is shown when data diversity is increased by learning the range of magnitudes for single ($N=1$) and composite augmentation operations ($N>1$) using \method. For curriculum learning, target PSNR value is annealed from 40 to the value mentioned in the legend. For \textbf{generalization}, we are interested in the gap between training and validation accuracy \citep{jiang2020Fantastic}. \hl{Given two models with similar validation accuracy, the model with a larger gap between training and validation accuracy demonstrates better generalization as training is much harder for this model.}
    }
    \label{fig:aug_effect_perf_char}
\end{figure}

\begin{figure}[t!]
    \centering
    \begin{subfigure}[b]{0.49\columnwidth}
        \centering
        \resizebox{\columnwidth}{!}{
            \begin{tabular}{cccc}
                \toprule[1.5pt]
                \textbf{$\Delta$} & \textbf{Brightness} & \textbf{Contrast} & \textbf{Noise} \\
                \midrule[1.25pt]
                $5$ & \raisebox{-0.5\totalheight}{\includegraphics[width=0.3\columnwidth]{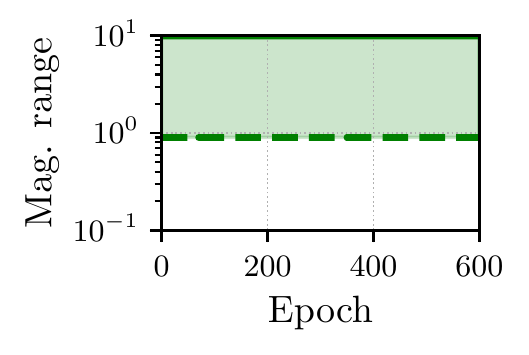}} & \raisebox{-0.5\totalheight}{\includegraphics[width=0.3\columnwidth]{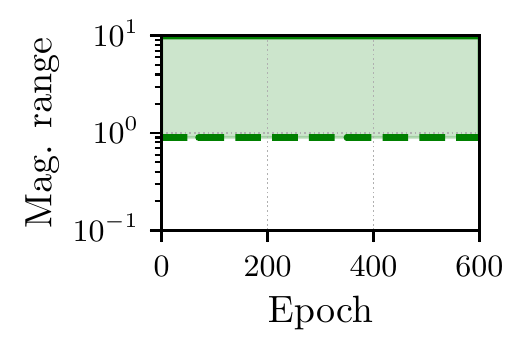}}
                 & \raisebox{-0.5\totalheight}{\includegraphics[width=0.3\columnwidth]{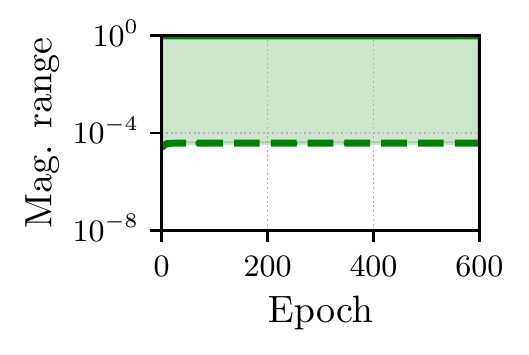}} \\
                 \midrule
                 $15$ & \raisebox{-0.5\totalheight}{\includegraphics[width=0.3\columnwidth]{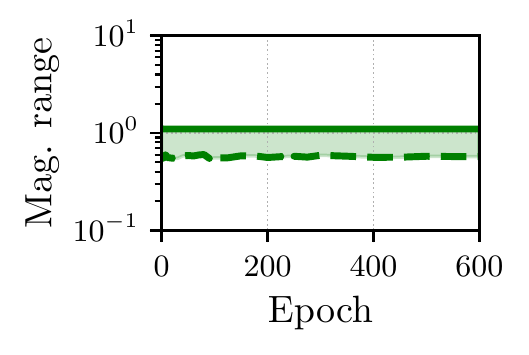}} & \raisebox{-0.5\totalheight}{\includegraphics[width=0.3\columnwidth]{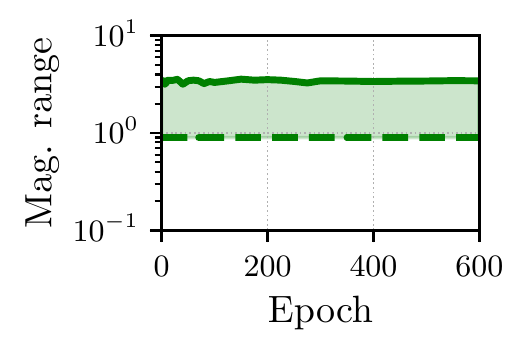}}
                 & \raisebox{-0.5\totalheight}{\includegraphics[width=0.3\columnwidth]{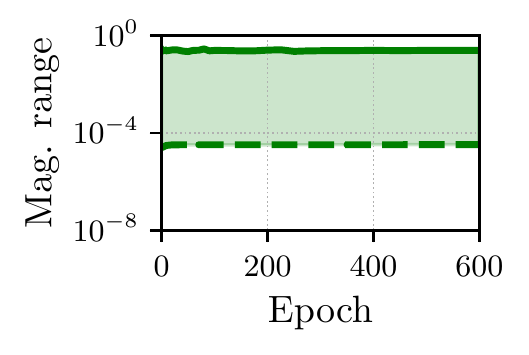}} \\
                \bottomrule[1.5pt]
            \end{tabular}
        }
        \caption{$N=1$ (without curriculum)}
        \label{fig:learned_range_r50_n1}
    \end{subfigure}
    \hfill
    \begin{subfigure}[b]{0.49\columnwidth}
        \centering
        \resizebox{\columnwidth}{!}{
            \begin{tabular}{cccc}
                \toprule[1.5pt]
                \textbf{$\Delta$} & \textbf{Brightness} & \textbf{Contrast} & \textbf{Noise} \\
                \midrule[1.25pt]
                5 & \raisebox{-0.5\totalheight}{\includegraphics[width=0.3\columnwidth]{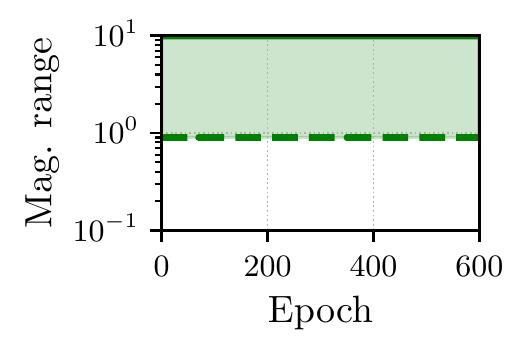}} & \raisebox{-0.5\totalheight}{\includegraphics[width=0.3\columnwidth]{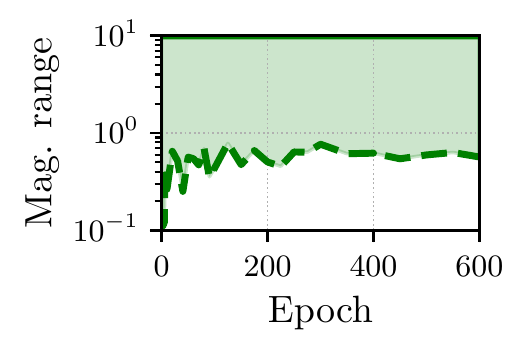}}
                 & \raisebox{-0.5\totalheight}{\includegraphics[width=0.3\columnwidth]{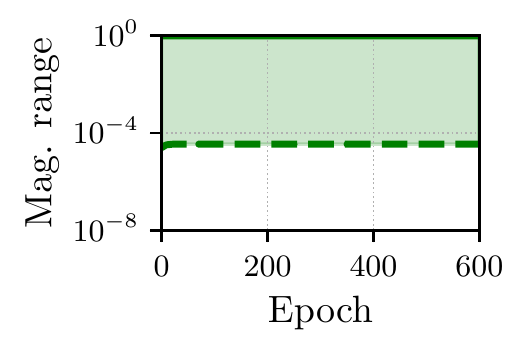}} \\
                 \midrule
                 15 & \raisebox{-0.5\totalheight}{\includegraphics[width=0.3\columnwidth]{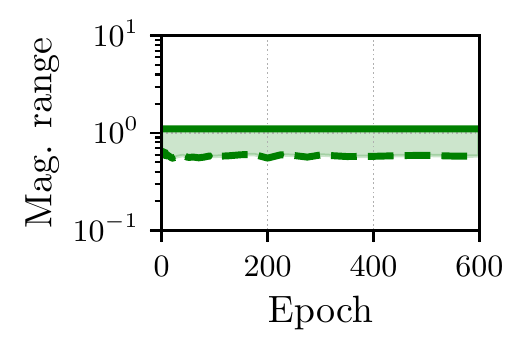}} & \raisebox{-0.5\totalheight}{\includegraphics[width=0.3\columnwidth]{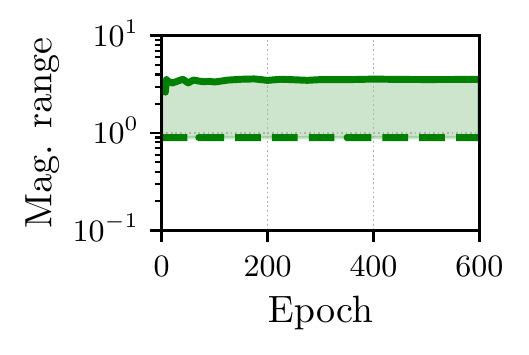}}
                 & \raisebox{-0.5\totalheight}{\includegraphics[width=0.3\columnwidth]{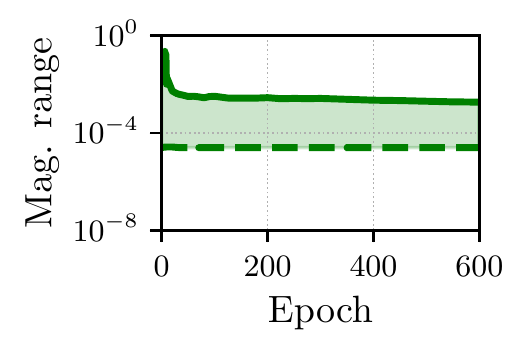}} \\
                \bottomrule[1.5pt]
            \end{tabular}
        }
        \caption{$N=3$ (without curriculum)}
        \label{fig:learned_range_r50_n3}
    \end{subfigure}
    \vfill
    \begin{subfigure}[b]{\columnwidth}
        \centering
        \resizebox{0.54\columnwidth}{!}{
            \begin{tabular}{cccc}
                \toprule[1.5pt]
                \textbf{$\Delta$} & \textbf{Brightness} & \textbf{Contrast} & \textbf{Noise} \\
                \midrule[1.25pt]
                $40 \rightarrow 5$ &
                \raisebox{-0.5\totalheight}{\includegraphics[width=0.3\columnwidth]{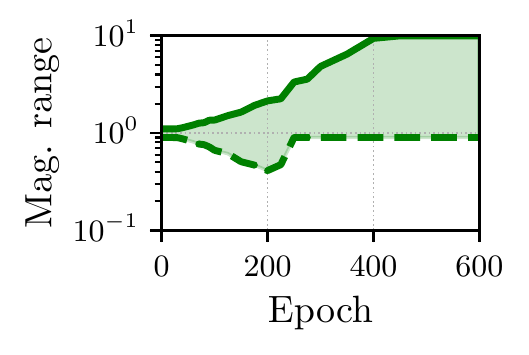}} & \raisebox{-0.5\totalheight}{\includegraphics[width=0.3\columnwidth]{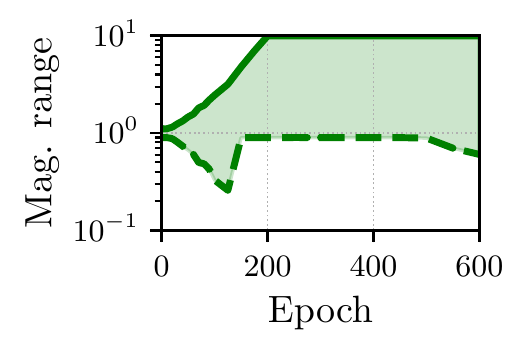}}
                 & \raisebox{-0.5\totalheight}{\includegraphics[width=0.3\columnwidth]{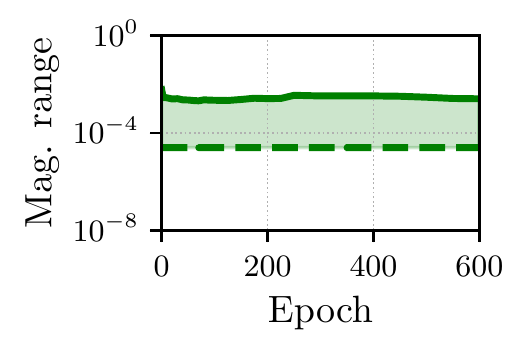}} \\
                 \midrule
                 $40 \rightarrow 15$ & \raisebox{-0.5\totalheight}{\includegraphics[width=0.3\columnwidth]{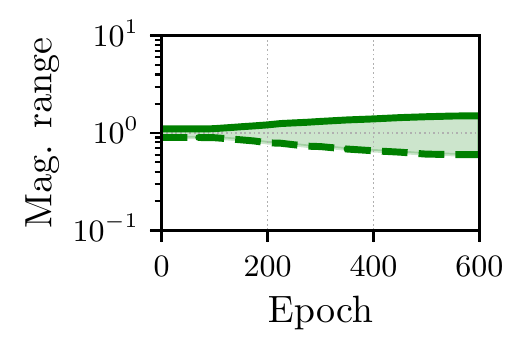}} & \raisebox{-0.5\totalheight}{\includegraphics[width=0.3\columnwidth]{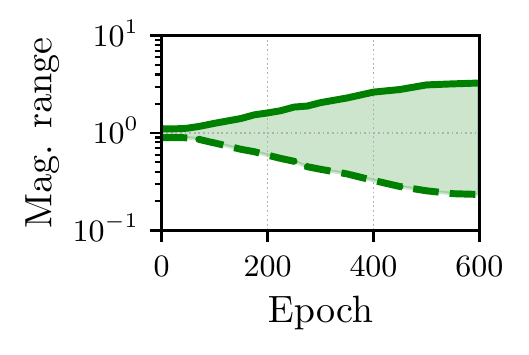}}
                 & \raisebox{-0.5\totalheight}{\includegraphics[width=0.3\columnwidth]{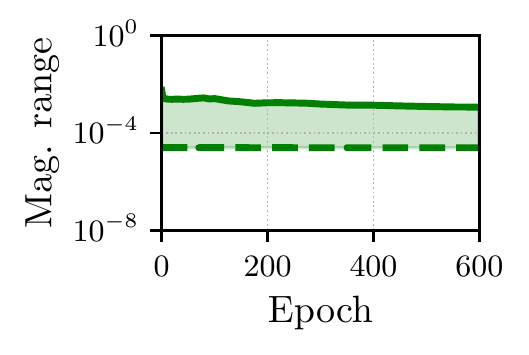}} \\
                \bottomrule[1.5pt]
            \end{tabular}
        }
        \caption{$N=3$ (with curriculum)}
        \label{fig:learned_range_r50_n3_curr}
    \end{subfigure}
    \caption{\textbf{Learned range of magnitudes for different augmentation operations when training ResNet-50 jointly with \method.} In (a) and (b), the target image similarity value (PSNR; $\Delta$) is fixed while in (c), $\Delta$ is annealed using cosine curriculum. Training epochs and magnitude range (log scale) of augmentation operations are plotted in x- and y-axis respectively.}
    \label{fig:learned_range_r50}
\end{figure}

\subsection{Augmentation Operation Characterization  using \method}
\label{ssec:neural_aug_op_char_res50}

The quantity and diversity of training data can be increased by (1) increasing the magnitude range of each augmentation operation and (2) composite augmentation operations. \cref{fig:aug_effect_perf_char} characterizes the effect of these variables on ResNet-50's performance on the ImageNet dataset. We can make the following observations:
\begin{enumerate}[leftmargin=*]
    \item \textbf{Single augmentation operation improves baseline accuracy.} \cref{fig:learned_range_r50_n1} shows that single augmentation operation ($N=1$) with wider magnitude ranges (e.g., the range of magnitudes at target PSNR of 5 are wider than the ones at target PSNR of 15) helped in improving ResNet-50's validation accuracy and reducing training accuracy, thereby improving its generalization capability (\cref{fig:aug_effect_perf_char_n1}). Particularly, increasing the magnitude range of contrast (or brightness) operation increased ResNet-50's validation performance over baseline by 1.0\% (or 0.7\%) while decreasing the training performance by 2\% (or 3\%). This is likely because wider magnitude ranges of an augmentation operation increases diversity of training data. On the other hand, the additive Gaussian noise operation with a wider magnitude range slightly dropped the validation accuracy, but still reduces the training accuracy. In other words, it improves ResNet-50's generalization ability. 
    \item \textbf{Composite augmentation operations improve model's generalization.} Composite augmentation operations ($N > 1$; \cref{fig:aug_effect_perf_char_n2,fig:aug_effect_perf_char_n3}) reduces the training accuracy significantly while having a validation performance similar to single augmentation operation. This is expected as composite operations further increases training data diversity.
    \item \textbf{Curriculum matters.} \cref{fig:learned_range_r50_n3_curr} shows that progressively learning to increase the diversity of augmented samples (i.e., narrower to wider magnitude ranges\footnote{Wider magnitude ranges produce diverse augmented samples and vice versa (\cref{sec:appendix_sample_vis}).}) using a cosine curriculum further improves the performance (\cref{fig:aug_effect_perf_char_n3}). A plausible explanation is that the learned magnitude ranges of different augmentation operations using \method with fixed target PSNR may get stuck near poor solutions. Because the range of magnitudes  are wider for these solutions (e.g., the range of magnitudes at target PSNR value of 5 in  \cref{fig:learned_range_r50_n1} \& \cref{fig:learned_range_r50_n3}), \method samples more diverse data from the beginning of the training, making training difficult. Gradually annealing target PSNR from high to low (e.g., 40 to 5 in \cref{fig:learned_range_r50_n3_curr}) allows \method to increase the data diversity slowly, thereby helping model to learn better representations. Importantly, it also allows \method to identify useful ranges for each augmentation operation. For example, the range of magnitudes for the additive Gaussian noise operation is relatively narrower when optimizing with curriculum (e.g., annealing target PSNR from 40 to 5; \cref{fig:learned_range_r50_n3_curr}) compared to optimizing with a fixed target PSNR of 5 (\cref{fig:learned_range_r50_n3}). This indicates that noise operation with narrow magnitude range is favorable for training ResNet-50 on the ImageNet classification task. This concurs with results in \cref{fig:aug_effect_perf_char_n1,fig:aug_effect_perf_char_n2} where we observed that noise operation does not improve ResNet-50's validation performance. Moreover, our findings with progressively increasing data diversity are consistent  with previous works \citep[e.g.,][]{bengio2009curriculum, tan2021efficientnetv2} that shows scheduling training samples from easy to hard helps improve model's performance.

    Interestingly, ResNet-50 with \method ($N=3$) achieves comparable validation accuracy and a smaller generalization gap as compared to state-of-the-art methods which use more augmentation operations ($N > 14$) to increase data diversity during training. We conjecture that the difference in the generalization gap between existing methods and \method is probably caused by insufficient policy search in existing methods as they use manually-defined magnitude ranges for each augmentation operation during search.
\end{enumerate}

\begin{tcolorbox}[width=\columnwidth,
                  colback=green!5!white,
                  colframe=green!75!black
                  ]
\small{
\textbf{Observation 1:} Composite augmentation operations with \textbf{wider magnitude ranges} are important for improving downstream model's generalization ability.}
\end{tcolorbox}

In the rest of experiments, we will use all three augmentations ($N=3$) with cosine curriculum.

\begin{figure}[b!]
    \centering
    \begin{subfigure}[b]{\columnwidth}
        \centering
        \includegraphics[width=\columnwidth]{gen_graphs/legend.pdf}
    \end{subfigure}
    \vfill
    \begin{subfigure}[b]{\columnwidth}
        \centering
        \begin{tabular}{cc}
           \includegraphics[width=0.35\columnwidth]{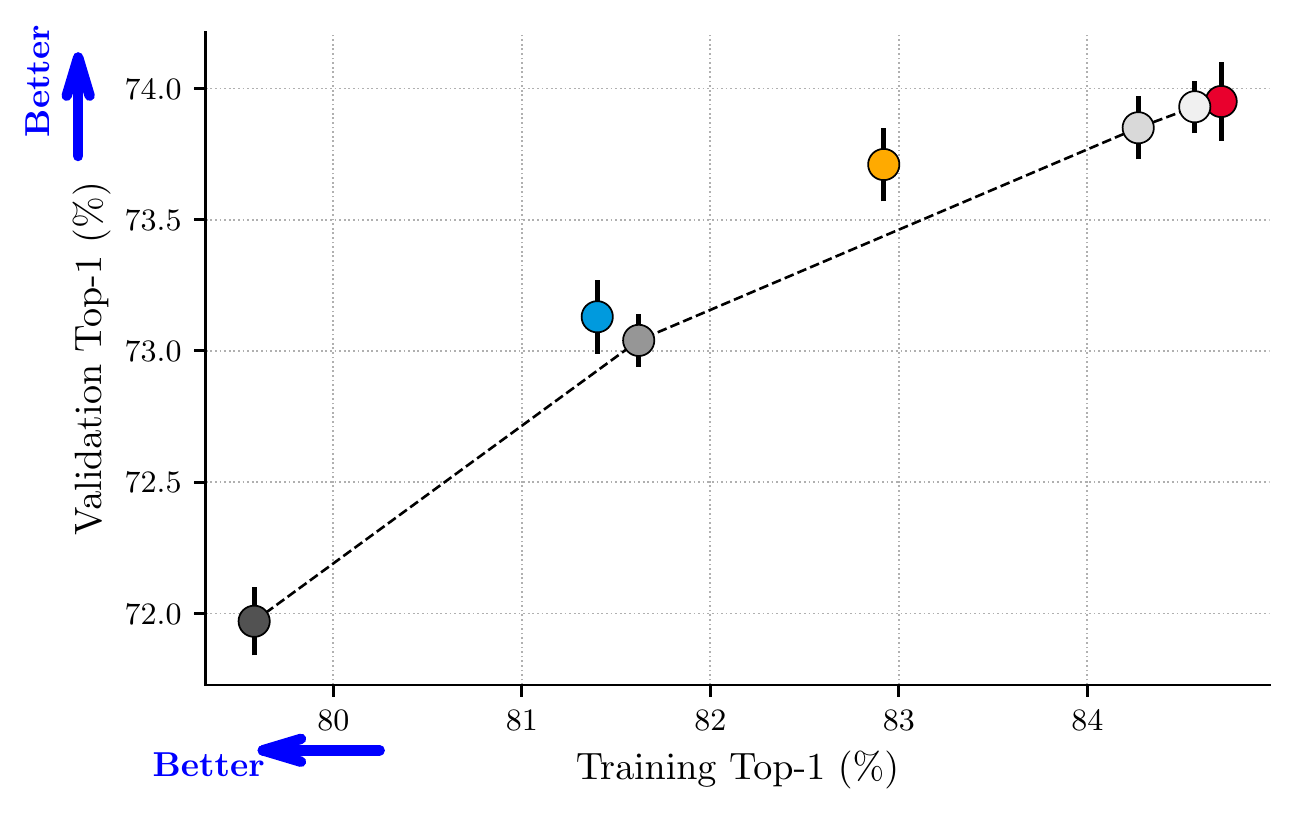}  &
           \includegraphics[width=0.35\columnwidth]{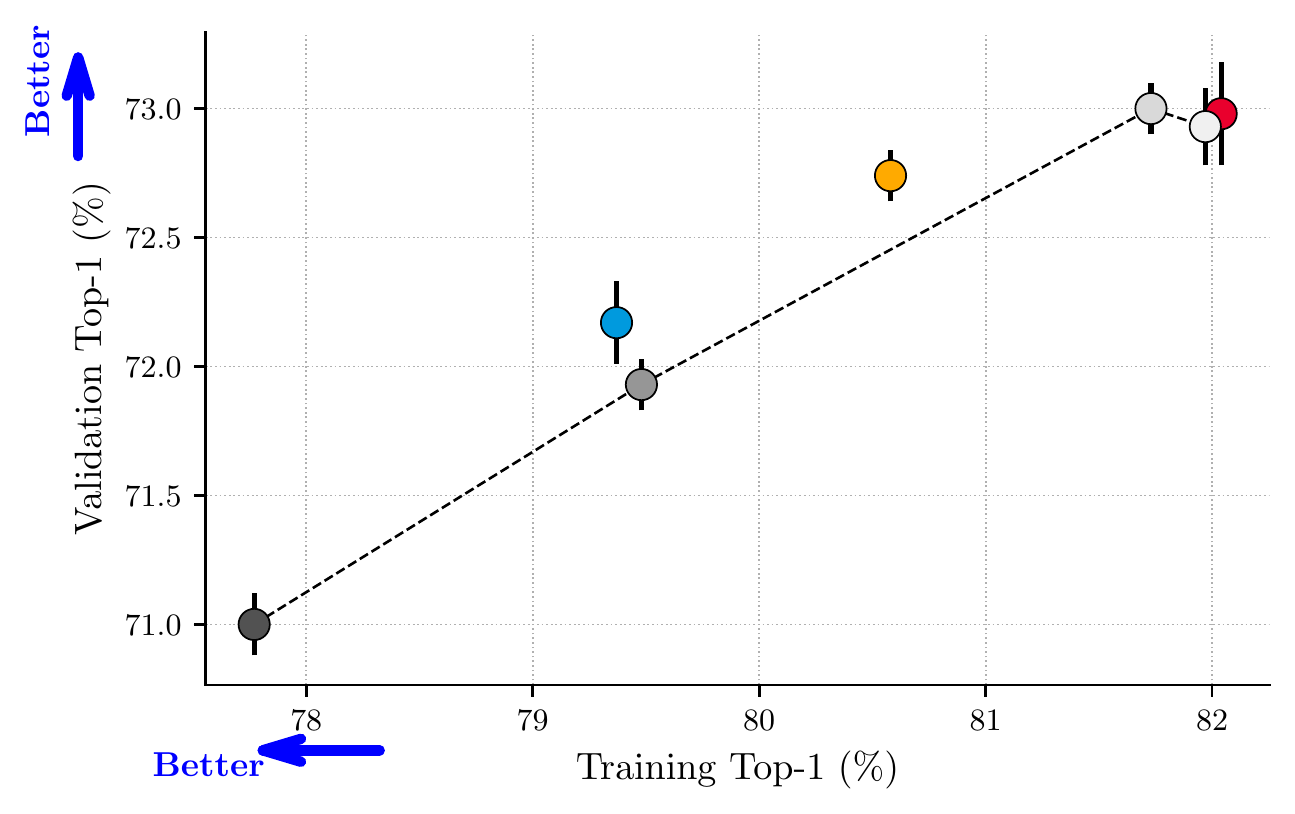} \\
           MobileNetv1 & MobileNetv2 \\
           \includegraphics[width=0.35\columnwidth]{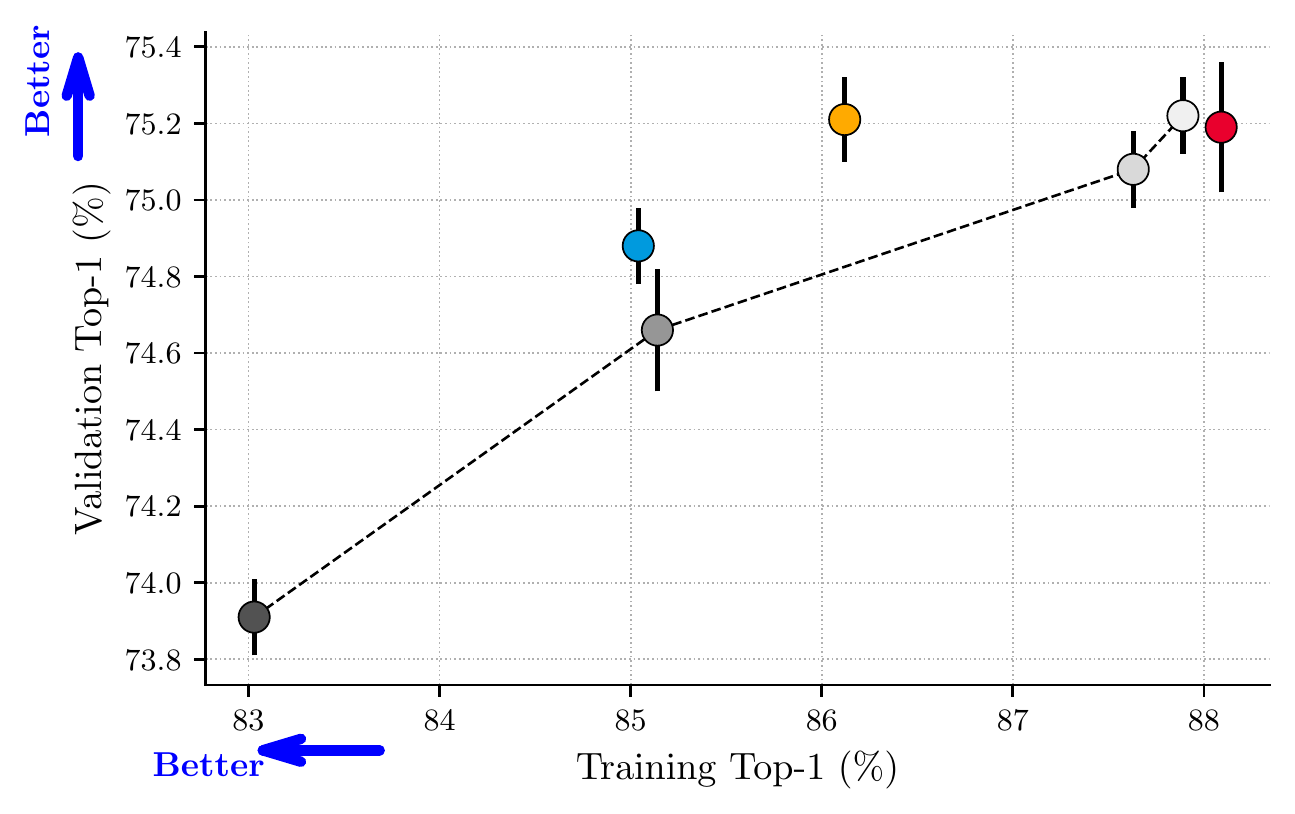} & 
           \includegraphics[width=0.35\columnwidth] {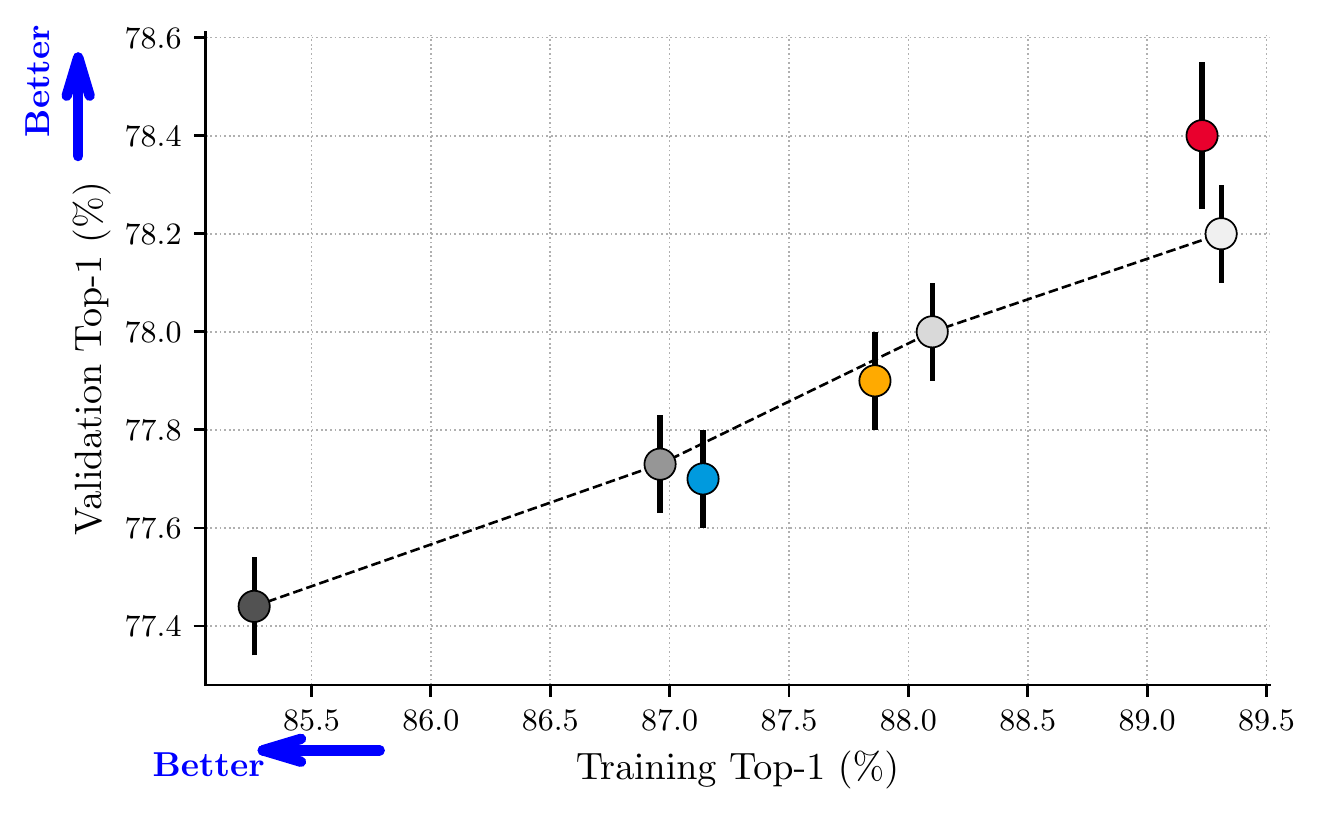} \\
             MobileNetv3 & MobileViT \\
        \end{tabular}
        \caption{Mobile models}
    \end{subfigure}
    \vfill
    \begin{subfigure}[b]{\columnwidth}
        \centering
        \begin{tabular}{cc}
           \includegraphics[width=0.35\columnwidth]{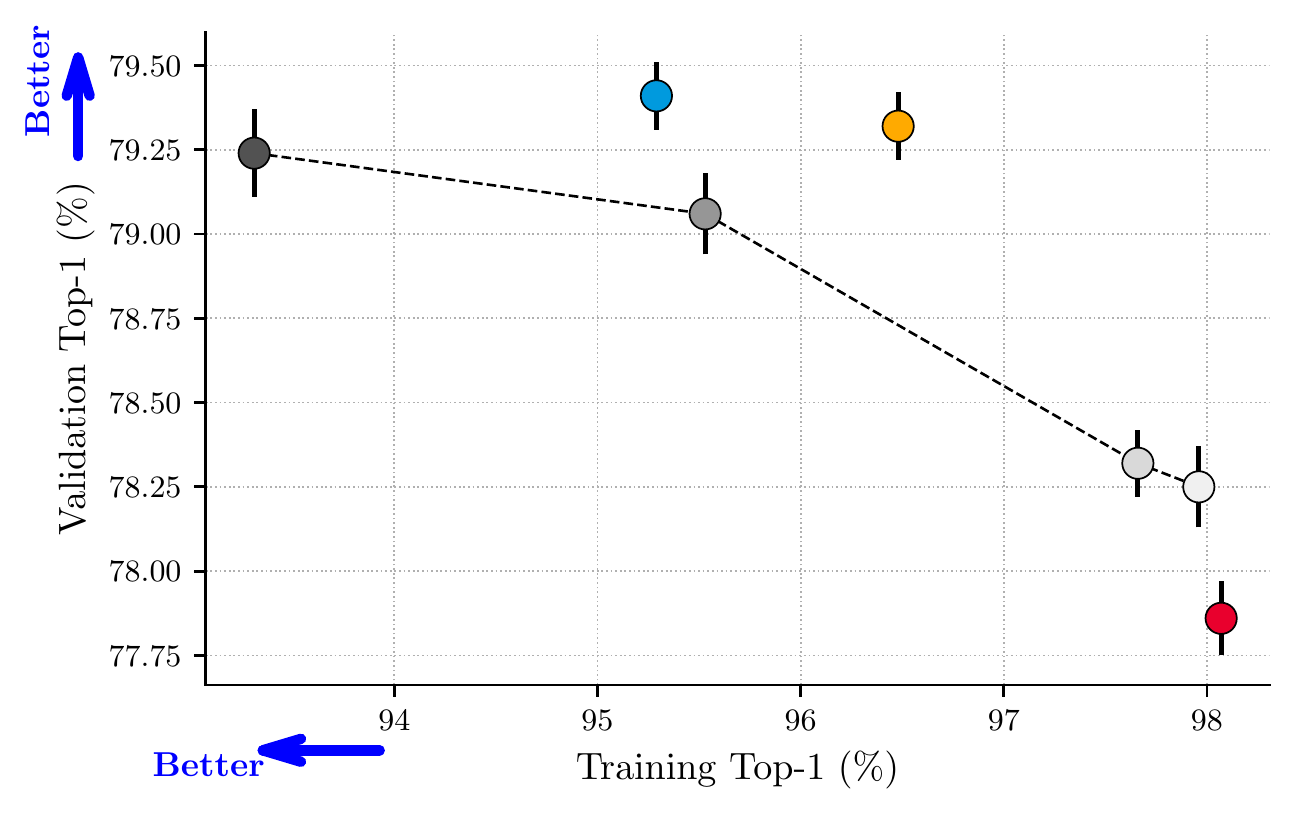}  &
           \includegraphics[width=0.35\columnwidth]{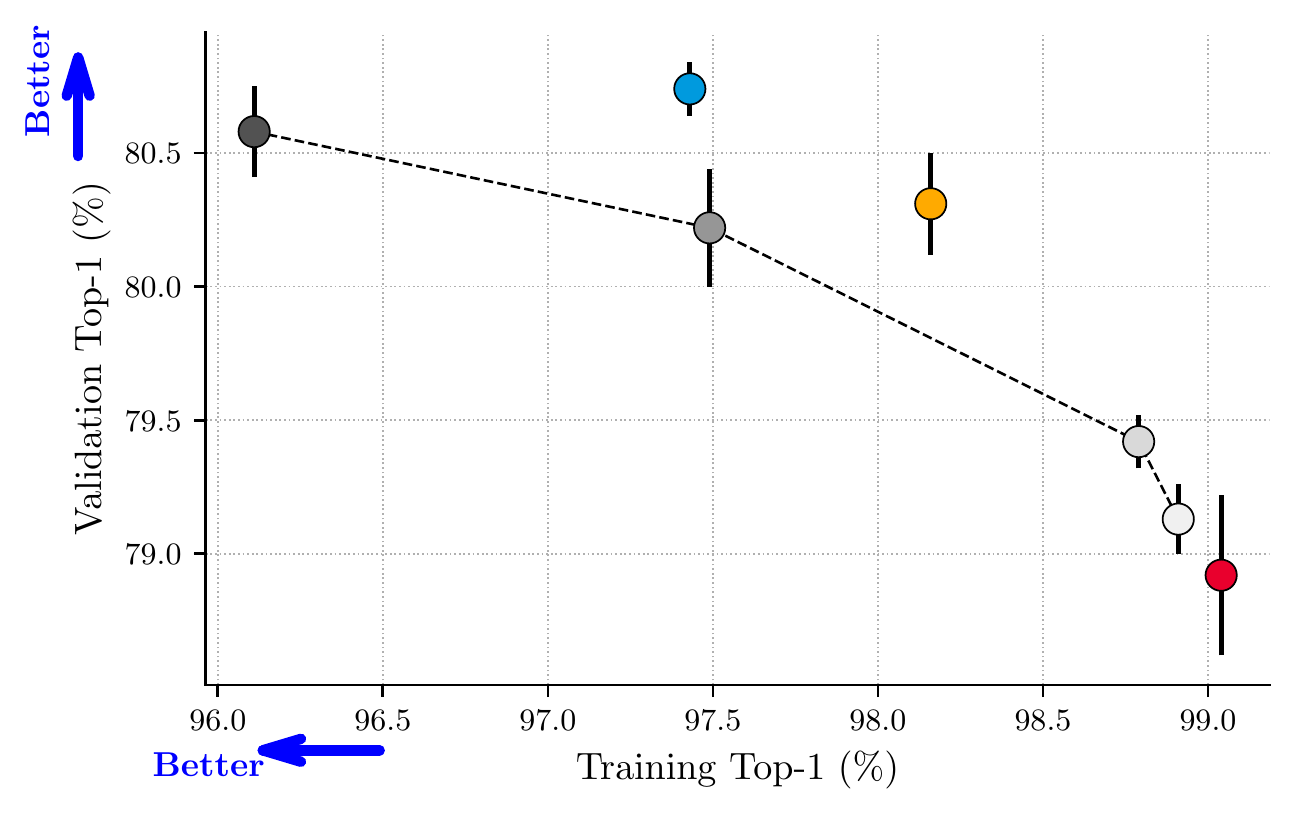} \\
           ResNet-50 & ResNet-101 \\
           \includegraphics[width=0.35\columnwidth]{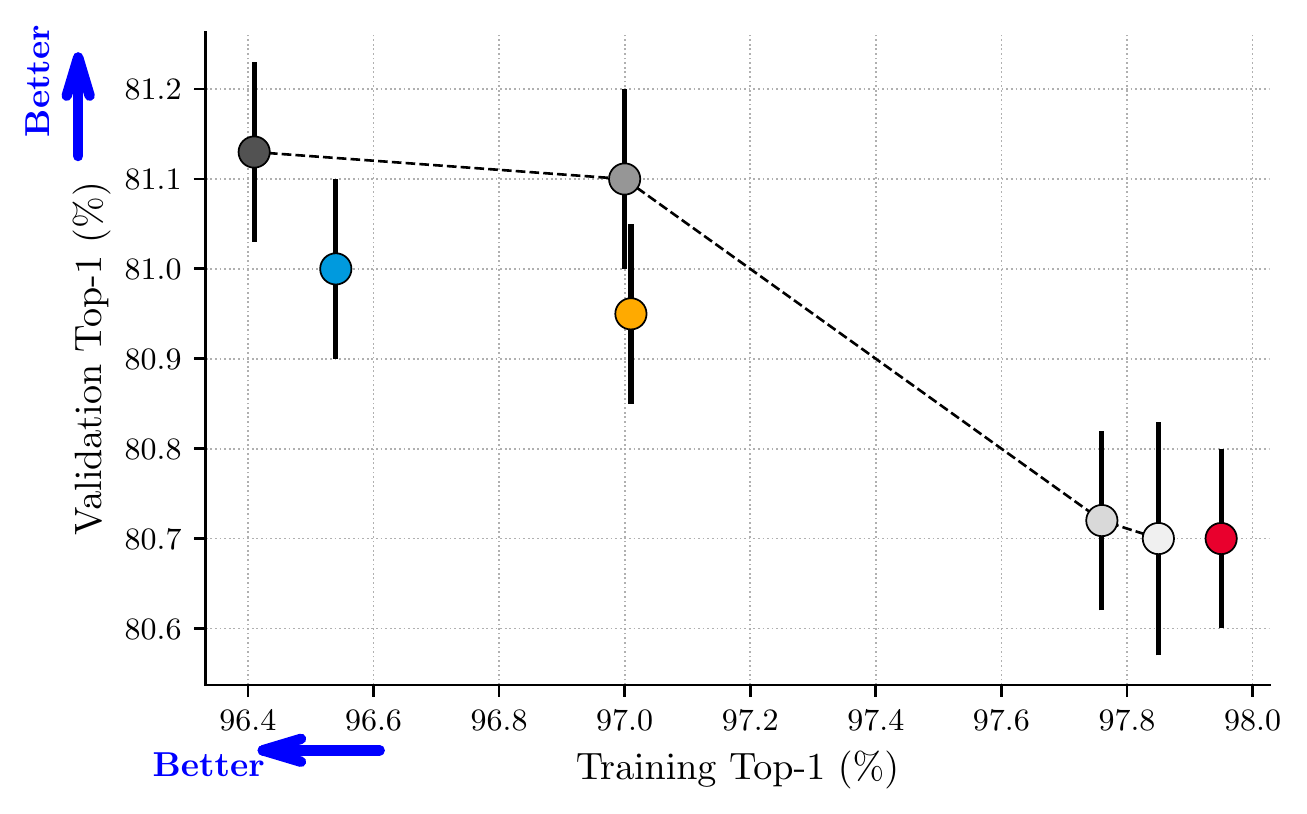} & 
           \includegraphics[width=0.35\columnwidth]{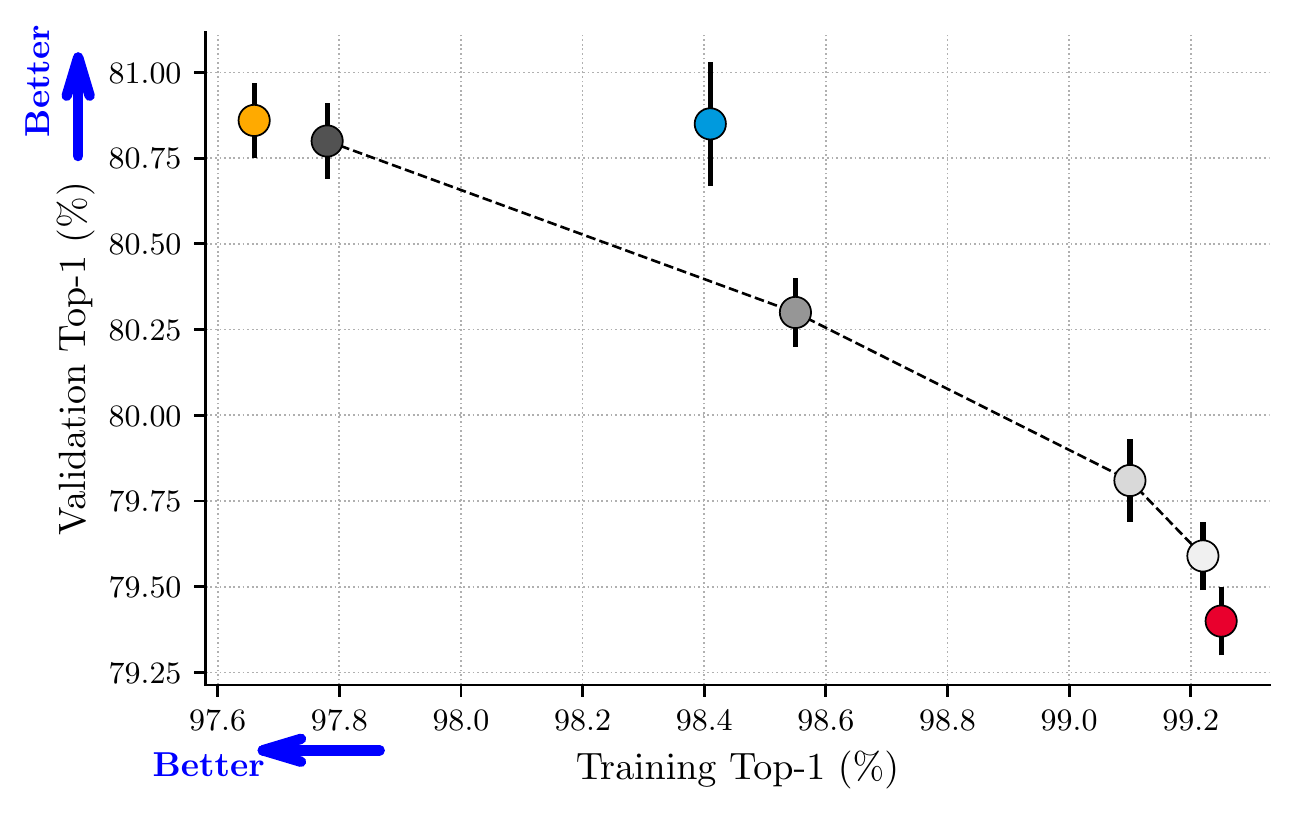} \\
             EfficientNet-B3 & SwinTransformer-Tiny \\
        \end{tabular}
        \caption{Non-mobile models}
    \end{subfigure}
    \caption{\textbf{\method is model agnostic.} Performance of different models on the ImageNet dataset using \method. Here, the target PSNR value is annealed from 40 to the value mentioned in the legend.}
    \label{fig:generic_neural_aug}
\end{figure}

\subsection{Model-level Generalization of \method}
\label{ssec:neural_aug_model_gen}

\cref{fig:aug_effect_perf_char} shows \method is effective for ResNet-50. Natural questions that arise are:
\begin{enumerate}[leftmargin=*]
    \item \textbf{Can \method be applied to other vision models?} \method's seamless integration ability with little training overhead (0.5\% to 3\%) over the baseline model allows us to study the generalization capability of different vision models easily. \cref{fig:generic_neural_aug} shows the performance of different models with \method. When data regularization is increased for mobile models by decreasing the target PSNR value from 40 to 5, the training as well as validation accuracy of different mobile models is decreased significantly as compared to the baseline. This is likely because of the limited capacity of these models. On the other hand, data regularization improved the performance of non-mobile models significantly. Consistent with our observations for ResNet-50 in  \cref{fig:aug_effect_perf_char}, we found that non-mobile models trained with \method are able to achieve competitive performance to state-of-the-art automatic augmentation methods, such as AutoAugment ($N=16$), but with fewer augmentations ($N=3$).
    \item \textbf{Does \method learn architecture-specific augmentations?} \cref{fig:learned_range_diff_arch} shows the learned magnitude ranges for different augmentation operations for a transformer- (SwinTransformer) and a CNN-based (EfficientNet) model. Though both of these architectures use the same curriculum in \method (i.e., target PSNR is annealed from 40 to 5), they learn different magnitude ranges  for each augmentation operation. This shows that \method is capable of learning model-specific magnitude ranges for each augmentation operation.
    \item \textbf{Does \method increase variance on model performance?} Because of the stochastic training and presence of randomness during different stages of training including \method, there may be some variability in model's performance. To measure the variability in model's performance, we run each experiment with three different random seeds. For different models, the standard deviation of model's validation accuracy in \cref{fig:generic_neural_aug} is between $0.01$ and $0.2$, and is in accordance with previous works on the ImageNet dataset \citep{radosavovic2020designing, wightman2021resnet}. These results show that training models with \method leads to a stable model performance.
\end{enumerate}

\subsection{Comparison with Existing Methods}
\label{ssec:sota_imagenet_perf}

\paragraph{Comparison for ResNet-50} Most previous automatic augmentation methods use ResNet-50 as a baseline model. For apples to apples comparison, we train ResNet-50 with different augmentation methods using the same training recipe (\cref{sec:appendix_training_details}). \cref{tab:compare_sota_imagenet_res50} shows that \method achieves similar performance as previous methods, but with $4-5 \times$ fewer augmentation operations. Note that, unlike previous methods, the search space of \method is independent of the augmentation operations and is constant (\cref{tab:teaser_table_img1k} and \cref{fig:standard_vs_rangeAug}). 

\begin{table}[t!]
    \centering
    \resizebox{0.6\columnwidth}{!}{
        \begin{tabular}{lcc}
            \toprule[1.5pt]
           \textbf{Aug. method} & \textbf{\# of operations}  & \textbf{Top-1 accuracy} \\
           \midrule[1.25pt]
            AutoAugment \citep{cubuk2019autoaugment} & 16 & 77.6\% \\
            RandAugment \cite{cubuk2020randaugment} & 14 & 77.6\% \\
            Fast AutoAugment \citep{dollar2021fast} & 16 & 77.6\% \\
            TrivialAugment \citep{muller2021trivialaugment}& 14 & 78.1\% \\
            Deep AutoAugment \citep{zheng2022deep} & 16 & 78.3\% \\
            \midrule
            RandAugment + MixUp \citep{wightman2021resnet} & 14 + 2 & \textbf{80.4}\% \\
            RandAugment + MixUp + RandonErase \citep{touvron2021training} & 14 + 2 + 1 & 79.5\% \\
            AutoAugment + MixUp \citep{touvron2019fixing} & 16 + 2 & 79.5\% \\
            \midrule
            AutoAugment + Mixup (Our repro.) & 16 + 2 & \textbf{80.2}\% \\
            RandAugment + Mixup (Our repro.) & 14 + 2 & \textbf{80.4}\% \\
            TrivialAugment + Mixup (Our repro.) & 14 + 2 & \textbf{80.4}\% \\
            Deep AutoAugment + Mixup (Our repro.) & 16 + 2 & 80.1\% \\
            \method + Mixup (Ours) & \textbf{3 + 2} & \textbf{80.2}\% \\
            \bottomrule[1.5pt]
        \end{tabular}
    }
    \caption{\textbf{Accuracy of ResNet-50 on the ImageNet validation set.} \method compares favorably to previous methods, but with $4-5 \times$ fewer augmentation operations. Best results within ImageNet's standard deviation range of $\pm 0.2$ are highlighted in bold.}
    \label{tab:compare_sota_imagenet_res50}
\end{table}

\begin{table}[t!]
    \centering
    \resizebox{0.75\columnwidth}{!}{
        \begin{tabular}{llcccc}
            \toprule[1.5pt]
           \multirow{2}{*}{\textbf{Model}} & \multirow{2}{*}{\bfseries Source} & \multicolumn{3}{c}{\bfseries Data augmentation methods} & \textbf{Top-1} \\ 
           \cmidrule[1.25pt]{3-5}
            & & \textbf{Auto aug. method} & \textbf{Random erase} & \textbf{Mixup} & \textbf{accuracy} \\
           \midrule[1.25pt]
            \multicolumn{6}{c}{\bfseries Mobile models} \\
           \midrule
            \multirow{4}{*}{\bfseries MobileNetV1-1.0} & Orig. \citep{howard2017mobilenets} & \xmarkA & \xmarkA & \xmarkA & 70.6\% \\
           & Our repro. & RandAugment ($N=14$) & \xmarkA & \xmarkA & \textbf{73.7\%} \\
           & Our repro. & AutoAugment  ($N=16$) & \xmarkA & \xmarkA & 73.1\% \\
           & Ours & \method  ($N=3$) & \xmarkA & \xmarkA & \textbf{73.8\%} \\
            \midrule
            \multirow{4}{*}{\bfseries MobileNetV2-1.0} & Orig. \citep{sandler2018mobilenetv2} & \xmarkA & \xmarkA & \xmarkA & 72.0\% \\
           & Our repro. & RandAugment ($N=14$) & \xmarkA & \xmarkA & 72.7\% \\
           & Our repro. & AutoAugment ($N=16$) & \xmarkA & \xmarkA & 72.1\% \\
           & Ours & \method ($N=3$) & \xmarkA & \xmarkA & \textbf{73.0\%} \\
            \midrule
            \multirow{4}{*}{\bfseries MobileNetV3-Large} & Orig. \citep{howard2019searching} & \xmarkA & \xmarkA & \xmarkA & 74.6\% \\
           & Our repro. & RandAugment ($N=14$) & \xmarkA & \xmarkA & \textbf{75.2\%} \\
           & Our repro. & AutoAugment ($N=16$) & \xmarkA & \xmarkA & 74.9\% \\
           & Ours & \method ($N=3$) & \xmarkA & \xmarkA & \textbf{75.1\%} \\
            \midrule
            \multirow{4}{*}{\bfseries MobileViT-Small} & Orig. \citep{mehta2021mobilevit} & \xmarkA & \xmarkA & \xmarkA & \textbf{78.4\%} \\
            & Our repro. & RandAugment ($N=14$) & \xmarkA & \xmarkA & 77.9\% \\
           & Our repro. & AutoAugment ($N=16$) & \xmarkA & \xmarkA & 77.7\% \\
           & Ours & \method ($N=3$) & \xmarkA & \xmarkA & \textbf{78.2\%} \\
            \midrule
            \multicolumn{6}{c}{\bfseries Non-mobile models} \\
            \midrule
           \multirow{5}{*}{\bfseries ResNet-50} & Orig. \citep{he2016deep} & \xmarkA & \xmarkA & \xmarkA & 76.2\% \\
            & TIMM \citep{wightman2021resnet} & RandAugment ($N=14$) & \xmarkA & \cmarkA & \textbf{80.4\%} \\
            & Our repro. & RandAugment ($N=14$) & \xmarkA & \cmarkA & \textbf{80.4\%} \\
            & Our repro. & AutoAugment ($N=16$) & \xmarkA & \cmarkA & \textbf{80.2\%} \\
            & Ours & \method ($N=3$) & \xmarkA & \cmarkA & \textbf{80.2\%} \\
           \midrule
           \multirow{5}{*}{\bfseries ResNet-101} & Orig. \citep{he2016deep} & \xmarkA & \xmarkA & \xmarkA & 77.4\% \\
           & TIMM \citep{wightman2021resnet} & RandAugment ($N=14$) & \xmarkA & \cmarkA & 81.5\% \\
           & Our repro. & RandAugment ($N=14$) & \xmarkA & \cmarkA & 81.4\% \\
           & Our repro. & AutoAugment ($N=16$) & \xmarkA & \cmarkA & 81.6\% \\
           & Ours & \method ($N=3$) & \xmarkA & \cmarkA & \textbf{82.0\%} \\
           \midrule
           \multirow{5}{*}{\bfseries EfficientNet-B0} & Orig. \citep{tan2019efficientnet} & AutoAugment ($N=16$) & \cmarkA & \cmarkA & \textbf{77.1\%} \\
           & TIMM \citep{wightman2021resnet} & RandAugment ($N=14$) & \xmarkA & \cmarkA & 77.0\% \\
           & Our repro. & RandAugment ($N=14$) & \xmarkA & \cmarkA & 77.0\% \\
           & Our repro. & AutoAugment ($N=16$) & \xmarkA & \cmarkA & \textbf{77.1}\% \\
           & Ours & \method ($N=3$) & \xmarkA & \cmarkA & \textbf{77.3\%} \\
           \midrule
           \multirow{5}{*}{\bfseries EfficientNet-B1} & Orig. \citep{tan2019efficientnet} & AutoAugment ($N=16$) & \cmarkA & \cmarkA & 79.1\% \\
           & TIMM \citep{wightman2021resnet} & RandAugment ($N=14$) & \xmarkA & \cmarkA & 79.2\% \\
           & Our repro. & RandAugment ($N=14$) & \xmarkA & \cmarkA & 79.0\% \\
           & Our repro. & AutoAugment ($N=16$) & \xmarkA & \cmarkA & 79.2\% \\
           & Ours & \method ($N=3$) & \xmarkA & \cmarkA & \textbf{79.5\%} \\
           \midrule
           \multirow{5}{*}{\bfseries EfficientNet-B2} & Orig. \citep{tan2019efficientnet} & AutoAugment ($N=16$) & \cmarkA & \cmarkA & 80.1\% \\
           & TIMM \citep{wightman2021resnet} & RandAugment ($N=14$) & \xmarkA & \cmarkA & 80.4\% \\
           & Our repro. & RandAugment ($N=14$) & \xmarkA & \cmarkA & 80.8\% \\
           & Our repro. & AutoAugment ($N=16$) & \xmarkA & \cmarkA & 80.9\% \\
           & Ours & \method ($N=3$) & \xmarkA & \cmarkA & \textbf{81.3\%} \\
           \midrule
           \multirow{5}{*}{\bfseries EfficientNet-B3} & Orig. \citep{tan2019efficientnet} & AutoAugment ($N=16$) & \cmarkA & \cmarkA & 81.6\% \\
           & TIMM \citep{wightman2021resnet} & RandAugment ($N=14$) & \xmarkA & \cmarkA & 81.4\% \\
           & Our repro. & RandAugment ($N=14$) & \xmarkA & \cmarkA & 81.5\% \\
           & Our repro. & AutoAugment ($N=16$) & \xmarkA & \cmarkA & 81.2\% \\
           & Ours & \method ($N=3$) & \xmarkA & \cmarkA & \textbf{81.9\%} \\
           \midrule
           \multirow{4}{*}{\bfseries Swin-Tiny} & Orig. \citep{liu2021swin} & RandAugment ($N=14$) & \cmarkA & \cmarkA & \textbf{81.3\%} \\
           & Our repro. & RandAugment ($N=14$) & \xmarkA & \cmarkA & 80.8\% \\
           & Our repro. & AutoAugment ($N=16$) & \xmarkA & \cmarkA & 81.0\% \\
           & Ours & \method ($N=3$) & \xmarkA & \cmarkA & \textbf{81.1\%} \\
            \midrule
           \multirow{4}{*}{\bfseries Swin-Small} & Orig. \citep{liu2021swin} & RandAugment ($N=14$) & \cmarkA & \cmarkA & \textbf{83.0\%} \\
           & Our repro. & RandAugment ($N=14$) & \xmarkA & \cmarkA & 82.7\% \\
           & Our repro. & AutoAugment ($N=16$) & \xmarkA & \cmarkA & 82.7\% \\
           & Ours & \method ($N=3$) & \xmarkA & \cmarkA & \textbf{82.8\%} \\
           \bottomrule[1.5pt]
        \end{tabular}
    }
    \caption{\textbf{\method is competitive to previous methods}. As per our observations in \cref{ssec:exp_setup_imagenet}, we anneal $\Delta$ from 40 to 30 and 40 to 5 for mobile and non-mobile models respectively. Methods whose performance is within ImageNet's standard deviation range of $\pm0.2$ of the best model are highlighted in bold. Here, $N$ denotes the number of augmentation operations. We reproduce models with RangeAugment and AutoAugment to align low-level details.}
    \label{tab:compare_sota_imagenet_models}
\end{table}

\paragraph{Comparison with other models} To demonstrate the effectiveness of \method across different models, we compare with AutoAugment and RandAugment. This is because (1) most state-of-the-art models use either RandAugment or AutoAugment along with Mixup transforms \citep{zhang2018mixup, yun2019cutmix} and (2) different augmentation methods deliver similar performance (\cref{tab:compare_sota_imagenet_res50}). \cref{tab:compare_sota_imagenet_models} shows that \method delivers competitive performance to existing automatic methods with significantly fewer augmentation operations. 

\begin{tcolorbox}[width=\columnwidth,
                  colback=green!5!white,
                  colframe=green!75!black
                  ]
\small{
\textbf{Observation 2:} With three basic augmentation operations, \method delivers competitive performance to existing methods across different models.}
\tcblower
\small{
On the ImageNet dataset, we recommend to train non-mobile models by annealing $\Delta$ from high to low similarity between input and augmented images (e.g., anneal $\Delta$ from 40 to 5 (or 10)).
}
\end{tcolorbox}

\section{Task-level Generalization of \method}
\label{sec:task_gen_na}

\cref{sec:imagenet_neural_aug} shows the effectiveness of \method on different downstream models on the ImageNet dataset. However, one might ask whether \method can be used for tasks other than image classification. To evaluate this, we study \method different tasks: (1) semantic segmentation (\cref{ssec:semantic_seg}), (2) object detection (\cref{ssec:maskrcnn_coco}), (3) foundation models (\cref{ssec:clip}), and (4) knowledge distillation (\cref{ssec:distillation}). 

\subsection{Semantic Segmentation with DeepLabv3 on ADE20k}
\label{ssec:semantic_seg}

\paragraph{Dataset and baseline models} We use the ADE20k dataset \citep{zhou2017scene} that has 20k training and 2k validation images across 150 semantic classes. We report the segmentation performance in terms of mean intersection over union (mIoU) on the validation set. 

We integrate mobile and non-mobile classification models with the Deeplabv3 segmentation head \citep{chen2018encoder} and finetune each model for 50 epochs. See \cref{sec:appendix_training_details} for training details. We do not study SwinTransformer for semantic segmentation because it is not compatible with Deeplabv3's segmentation head design as it adjusts the atrous rate of convolutions to control the output stride of backbone network.

\paragraph{Baseline augmentation methods} For semantic segmentation, experts have hand-crafted augmentation policies, and these manual policies are used to train state-of-the-art semantic segmentation methods \citep[e.g.,][]{chen2017rethinking, zhao2017pyramid, xie2021segformer, liu2021swin}. For an apples to apples comparison, we train each segmentation model with three different random seeds and compare with the following baselines: (1) \textbf{\textit{Baseline}} - standard pre-processing (randomly resize short image dimension, random horizontal flip, and random crop), (2) \textbf{\textit{Manual}} - baseline pre-processing with hand-crafted augmentation operations (color jittering using photometric distortion, random rotation, and random Gaussian noise), and (3) \textbf{\textit{\method}} - baseline pre-processing with learnable range of magnitudes for brightness, contrast, and noise (i.e., $N=3$). For reference, we include DeepLabv3 results (if available) from a popular segmentation library \citep{mmseg2020}.

\paragraph{Comparison with segmentation-specific augmentation methods} \cref{tab:seg_res_na_compare_manual_na} shows that \method improves the performance of different models in comparison to other augmentation methods. Interestingly, for semantic segmentation, the magnitude range of additive Gaussian noise is wider compared to image classification (\cref{fig:neural_aug_data_policy}). This concurs with previous manual augmentation methods which also found that Gaussian noise is important for semantic segmentation \citep{zhao2017pyramid,asiedu2022decoder}. Overall, these results suggest that \method learns task-specific augmentation policies.

\paragraph{DeepLabv3 strikes back} \cref{tab:seg_res_na_compare_backbone,tab:seg_res_na_compare_recent} shows that \method improves the performance of DeepLabv3 significantly across different backbones, and is able to deliver better performance than UPerNet \citep{xiao2018unified}, which has been widely adopted for semantic segmentation tasks in recent state-of-the-art models, including ConvNext \citep{liu2022convnet}.

\begin{table}[t!]
    \centering
    \begin{subtable}[b]{\columnwidth}
        \centering
        \resizebox{0.95\columnwidth}{!}{
        \begin{tabular}{lccccccc}
            \toprule[1.5pt]
            \multirow{3}{*}{\textbf{Backbone}} & \multicolumn{6}{c}{\textbf{Augmentation method}} \\
            \cmidrule[1pt]{2-7}
            & \multirow{2}{*}{\textbf{Baseline}} & \multirow{2}{*}{\textbf{Manual}} & \multicolumn{4}{c}{\textbf{\method (Ours)}} \\
            \cmidrule[0.5pt]{4-7}
            & & & (40, 5) & (40, 10) & (40, 20) & (40, 30)\\
            \midrule[1.25pt]
            MobileNetv1 & 38.77 $\pm$ 0.20 & 38.12 $\pm$ 0.16 & 36.46 $\pm$ 0.19 & 38.20 $\pm$ 0.26 & 38.96 $\pm$ 0.34 & \textbf{39.37 $\pm$ 0.19} \\
            MobileNetv2 & 37.74 $\pm$ 0.29 & 37.10 $\pm$ 0.27 & 35.58 $\pm$ 0.30 & 37.43 $\pm$ 0.73 & 38.06 $\pm$ 0.17 & \textbf{38.23 $\pm$ 0.36} \\
            MobileNetv3 & 37.58 $\pm$ 0.67 & 36.68 $\pm$ 0.34 & 34.80 $\pm$ 0.16 & 36.54 $\pm$ 0.12 & 37.77 $\pm$ 0.24 & \textbf{38.10 $\pm$ 0.13} \\
            MobileViT & 37.69 $\pm$ 0.50 & 37.19 $\pm$ 0.47 & 35.04 $\pm$ 0.83 & 36.70 $\pm$ 0.19 & 37.94 $\pm$ 0.45 & \textbf{38.41 $\pm$ 0.22} \\
            \midrule
            ResNet-50 & 42.27 $\pm$ 0.54 & 43.29 $\pm$ 0.27 & 41.56 $\pm$ 0.21 & 42.95 $\pm$ 0.22 & \textbf{43.31 $\pm$ 0.16} & 43.00 $\pm$ 0.28  \\
            ResNet-101 & 43.29 $\pm$ 0.17 & 44.04 $\pm$ 0.52 & 43.22 $\pm$ 0.52 & 43.89 $\pm$ 0.43 & \textbf{44.77 $\pm$ 0.28} & 43.95 $\pm$ 0.31 \\
            EfficientNet-B3 & 40.86 $\pm$ 0.55 & 41.15 $\pm$ 0.65 & 39.42 $\pm$ 0.29 & 40.39 $\pm$ 0.48 & \textbf{41.43 $\pm$ 0.36} & 41.08 $\pm$ 0.36 \\
            \midrule
            ResNet-50$^\star$ & 42.57 $\pm$ 0.43 & 43.49 $\pm$ 0.21 & & & \textbf{43.90 $\pm$ 0.10} \\
            ResNet-101$^\star$ & 44.19 $\pm$ 0.17 & 45.24 $\pm$ 0.27 & &  & \textbf{46.42 $\pm$ 0.19} \\
            EfficientNet-B3$^\star$ & 41.19 $\pm$ 0.16 & 42.35 $\pm$ 0.35 & & & \textbf{43.68 $\pm$ 0.16}  \\
            \bottomrule[1.5pt]
        \end{tabular}
    }
    \caption{\method vs. segmentation-specific manual augmentation methods. All models are reproduced by us to align low-level implementation details. Here, $^\star$ indicates that these backbones are trained with \method and mixup transforms.}
    \label{tab:seg_res_na_compare_manual_na}
    \end{subtable}
    \begin{subtable}[b]{\columnwidth}
        \centering
        \resizebox{0.65\columnwidth}{!}{
        \begin{tabular}{llccc}
            \toprule[1.5pt]
           \textbf{Backbone} & \textbf{Seg. architecture} & \textbf{mIoU} \\
           \midrule[1.25pt]
           \multirow{4}{*}{\bfseries MobileNetv2} & DeepLabv3$^\dagger$ & 34.0 \\
             & PSPNet$^\ddagger$ & 35.8 \\
             & DeepLabv3 (Our repro.)  & $37.74 \pm 0.29$  \\
            & DeepLabv3 w/ \method (Ours) &  $\mathbf{38.23 \pm 0.36}$ \\
            \midrule
            \multirow{6}{*}{\bfseries ResNet-50} & UPerNet$^\ddagger$ & 40.4 \\
            & UPerNet$^\dagger$ & 42.1 \\
             & PSPNet$^\dagger$ & 42.5 \\
             & DeepLabv3$^\dagger$ & 42.7 \\
             & DeepLabv3 (Our repro.) & $43.49 \pm 0.21$ \\
             & DeepLabv3 w/ \method (Ours) & $\mathbf{43.90 \pm 0.10}$ \\
            \midrule
            \multirow{6}{*}{\bfseries ResNet-101} & UPerNet$^\ddagger$ & 42.0 
            \\
            & PSPNet$^\dagger$ & 42.5 \\
             & UPerNet$^\dagger$ & 43.8 \\
             & DeepLabv3$^\dagger$ & 45.0 \\
             & DeepLabv3 (Our repro.) & $45.24 \pm 0.27$ \\
             & DeepLabv3 w/ \method (Ours)  & $\mathbf{46.42 \pm 0.19}$ \\
            \bottomrule[1.5pt]
        \end{tabular}
    }
    \caption{Comparison with different architectures for the same backbone. $^\dagger$ and $^\ddagger$ represents that these results are from MMSeg \citep{mmseg2020} and CSAIL \citep{zhou2017scene} segmentation libraries, respectively. Our training schedules, including batch size, are similar to MMSeg and CSAIL.}
    \label{tab:seg_res_na_compare_backbone}
    \end{subtable}
    \begin{subtable}[b]{\columnwidth}
        \centering
        \resizebox{0.8\columnwidth}{!}{
        \begin{tabular}{lccc}
            \toprule[1.5pt]
           \textbf{Seg. model} & \textbf{\# Params.} & \textbf{FLOPs}  & \textbf{mIoU} \\
           \midrule[1.25pt]
           Swin w/ UPerNet \citep{liu2021swin} & \textbf{60 M} & 945 G & 45.8 \\ 
           ConvNext w/ UPerNet \citep{liu2022convnet} & \textbf{60 M} & 939 G & \textbf{46.7} \\
           ResNet-101 w/ DeepLabv3 (\method; ours)  & 77 M & \textbf{303 G} & $\mathbf{46.42 \pm 0.19}$ \\
            \bottomrule[1.5pt]
        \end{tabular}
        }
        \caption{Comparison with recent segmentation models}
        \label{tab:seg_res_na_compare_recent}
    \end{subtable}
    \caption{\textbf{Semantic segmentation on the ADE20k dataset.}}
    \label{tab:seg_res_na}
\end{table}

\begin{table}[t!]
    \centering
    \resizebox{0.4\columnwidth}{!}{
    \begin{tabular}{lcc}
        \toprule[1.5pt]
       \textbf{Backbone}  & \textbf{mIoU} & \textbf{Prev. best} \\
        \midrule[1.25pt]
        MobileNetV1 & 77.2 & -- \\
        MobileNetV2 & 76.7 & 75.3 \citep{sandler2018mobilenetv2} \\
        MobileNetv3 & 77.0 & -- \\
        \midrule
        EfficientNet-B3 & 82.0 & --\\
        ResNet-50 & 81.2 & -- \\
        ResNet-101 & 84.0 & 82.7$^\dagger$ \citep{chen2017rethinking} \\
        \bottomrule[1.5pt]
    \end{tabular}
    }
    \caption{\textbf{\method's curriculum are transferable from ADE20k to PASCAL VOC}. Here, each backbone is trained with DeepLabv3 and \method. For mobile backbones, $\Delta$ is annealed from 40 to 30 while for non-mobile backbones, $\Delta$ is annealed from 40 to 20 (as noted in \cref{tab:seg_res_na_compare_manual_na}). $^\dagger$Model is evaluated with multiple scales and left-right flips during inference. Here, -- represents that results are not available.}
    \label{tab:transfer_ade20k_pascal_voc}
\end{table}

\paragraph{\method's curriculum is transferable to other segmentation datasets} \cref{tab:seg_res_na_compare_manual_na} shows that mobile and non-mobile backbones prefer different curriculum for the task of semantic segmentation. For mobile models, annealing the value of $\Delta$ from 40 to 30 yields best results while for non-mobile models, annealing the value of $\Delta$ from 40 to 20 yields the best results. \emph{Can we use the same curriculum for semantic segmentation on other datasets?} To validate this, we train DeepLabv3 with different backbones on the PASCAL VOC dataset \citep{everingham2010pascal}. Table \ref{tab:transfer_ade20k_pascal_voc} shows that \method's curriculum are transferable from ADE20k to PASCAL VOC.  

\subsection{Object detection with Mask R-CNN on MS-COCO}
\label{ssec:maskrcnn_coco}

\paragraph{Dataset and baseline models} We use MS COCO dataset \citep{lin2014microsoft} that has about 118k training and 5k validation images across 81 catogries (including background). We report the instance detection and segmentation performance in terms of standar COCO mean average precision (mAP) on the validation set.

We integrate mobile and non-mobile backbones with the Mask R-CNN \citep{he2017mask}. Following Detectron2 \citep{wu2019detectron2} which finetunes Mask R-CNN for about 270k iterations, we also finetune our model for similar number of iterations (230k iterations) with multi-scale sampler of \citep{mehta2021mobilevit}. 

\paragraph{Baseline augmentation methods} We study Mask R-CNN with two augmentation methods: (1) \textbf{Baseline} - standard image resizing as in \citep{he2017mask}, and (2) \textbf{RangeAugment} - standard image resizing with learnable range of magnitudes for brightness, contrast, and  noise. 

\begin{figure}
    \centering
    \begin{subfigure}[b]{0.4\columnwidth}
        \includegraphics[width=0.9\columnwidth]{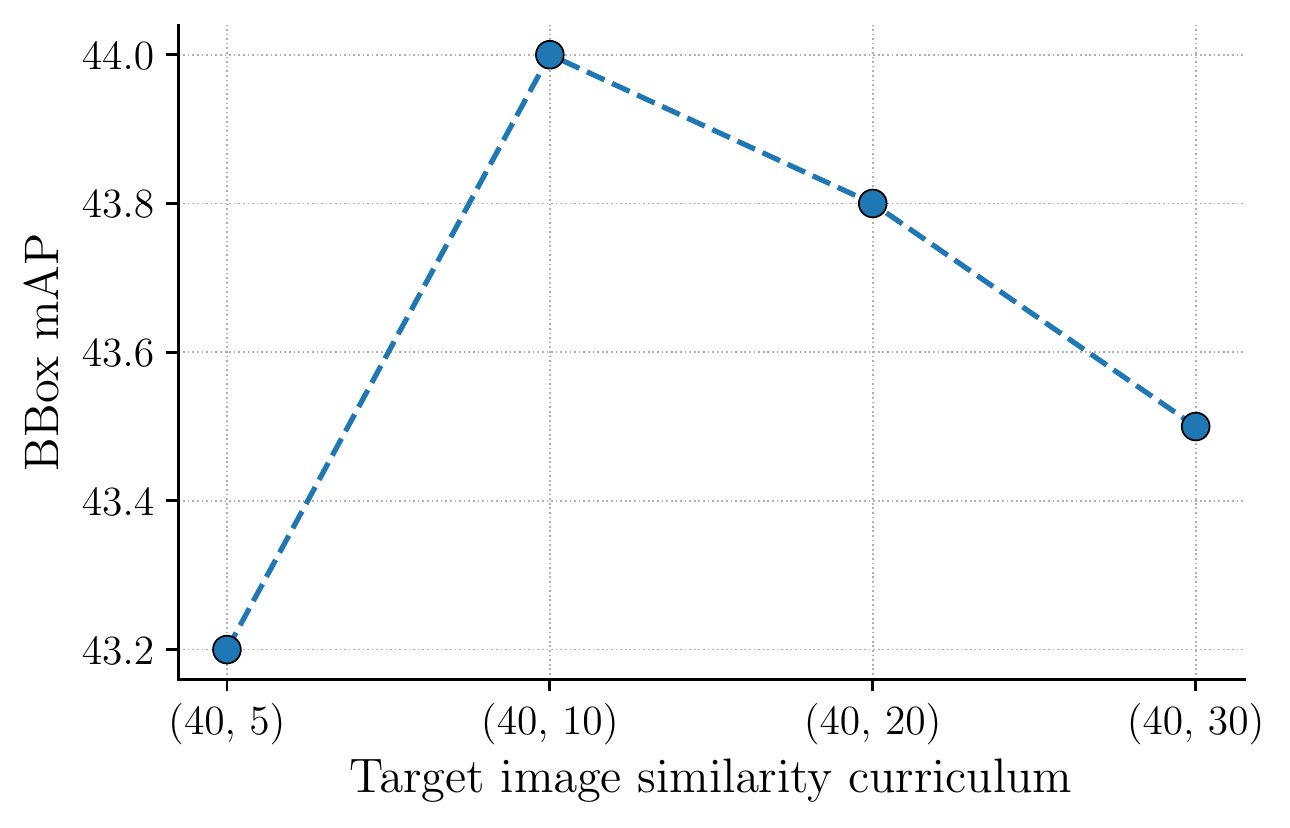}
        \caption{}
        \label{fig:maskrcnn_curr_res50}
    \end{subfigure}
    \hfill
    \begin{subfigure}[b]{0.55\columnwidth}
        \centering
        \resizebox{\columnwidth}{!}{
            \begin{tabular}{lccccc}
                \toprule[1.5pt]
                \textbf{Backbone} & \multicolumn{2}{c}{\bfseries mAP (w/o \method)} & & \multicolumn{2}{c}{\bfseries mAP (w/ \method)} \\
                \cmidrule[1.25pt]{2-3} \cmidrule[1.25pt]{5-6}
                 & \textbf{BBox} & \textbf{Segm} & & \textbf{BBox} & \textbf{Segm} \\
                \midrule[1.25pt]
                \multicolumn{6}{c}{Mobile backbones} \\
                \midrule
                 MobileNetv1 & 38.2 & 34.5 && \textbf{39.4} & \textbf{35.6}\\
                 MobileNetv2 & 37.7 & 34.1 && \textbf{38.4} & \textbf{34.7}\\
                 MobileNetv3 & 35.1 & 31.9 && \textbf{35.6} & \textbf{32.5}\\
                 MobileViT & 40.9 & 36.7 && \textbf{42.0} & \textbf{37.7}\\
                 \midrule[1.0pt]
                 \multicolumn{6}{c}{Non-mobile backbones} \\
                 \midrule
                 EfficientNet-B3 & 44.0 & 39.0 && \textbf{44.5} & \textbf{39.5}\\
                 ResNet-50 & 43.4 & 39.0 && \textbf{44.0} & \textbf{39.5}\\
                 ResNet-101 & 45.5 & 40.5 && \textbf{46.1} & \textbf{41.1}\\
                 \bottomrule[1.5pt]
            \end{tabular}
        }
        \caption{}
        \label{fig:maskrcnn_curr_diff_back}
    \end{subfigure}
    \caption{\textbf{Enhanced object detection results of Mask R-CNN with RangeAugment on COCO dataset.} (a) Effect of varying the target image similarity curriculumn on the performance of Mask R-CNN with ResNet50. Annealing the PSNR value from 40 to 10 yields best performance. (b) Effect of transferring the best curriculum in (a) to different backbones. We observe that \method improves detection performance across backbones. All models are trained by us to align low-level implementation details.}
    \label{fig:maskrcnn_results}
\end{figure}

\paragraph{Searching for the curriculum} We train Mask R-CNN with ResNet-50 for different target image similarity value, $\Delta$. \cref{fig:maskrcnn_curr_res50} shows that Mask R-CNN achieves the best performance when we anneal $\Delta$ from 40 to 10.

\paragraph{Transferring curriculum to other backbones} We integrate different backbones with Mask R-CNN and finetune it by annealing $\Delta$ from 40 to 10. \cref{fig:maskrcnn_curr_diff_back} shows that \method improves the detection performance across backbones significantly.

\subsection{Learning Augmentation for Foundation Models at Scale}
\label{ssec:clip}

A common perception in the community is that training on extremely large datasets does not require data augmentation \citep{radford2021learning, zhai2022scaling}. In this section, we challenge this perception and shows that \method improves the accuracy and data efficiency of large scale foundation models.

\paragraph{Dataset and baseline models} We train the CLIP \citep{radford2021learning} model with two different dataset sizes (304M and 1.1B image-text pairs) and evaluate their 0-shot top-1 accuracy on ImageNet's validation set using the same language prompts as as OpenCLIP \citep{ilharco_gabriel_2021_5143773}.

We train CLIP \citep{radford2021learning} with \method from scratch (\cref{sec:appendix_training_details}). The model uses ViT-B/16 (or ViT-H/16) \citep{dosovitskiy2020image} as its image encoder and transformer as its text encoder, and minimizes contrastive loss during training. We use multi-scale sampler of \citet{mehta2022cvnets} to make CLIP more robust to input scale changes. Because less data regularization is required at a scale of 100M+ samples, we anneal the target PSNR in \method from 40 to 20. We compare the performance with CLIP and OpenCLIP.

\paragraph{Results} \cref{tab:zero_shot_clip} compares the zero-shot performance of different models. Compared to OpenCLIP, \method achieves (1) 1.2\% better accuracy for similar dataset size and (2)  similar accuracy with $2\times$ fewer data samples. This shows that \method is effective when training foundation models on large-scale datasets.

\begin{table}[t!]
    \centering
    \begin{subtable}[b]{\columnwidth}
        \centering
    \resizebox{0.75\columnwidth}{!}{
        \begin{tabular}{cccc}
            \toprule[1.5pt]
            \textbf{Model} & \textbf{Dataset size}  & \textbf{Test resolution} & \textbf{Top-1 Accuracy (\%)} \\
            \midrule[1.25pt]
            CLIP of \citet{radford2021learning} & 400M &$224 \times 224$  & 68.3\% \\
            OpenCLIP of \citet{ilharco_gabriel_2021_5143773} & 400M & $224 \times 224$ & 67.1\% \\ 
            \midrule
            \multirow{5}{*}{CLIP  w/ \method (Ours)} & \multirow{5}{*}{302M (1.1B)} & $160 \times 160$ & 65.8\% (69.2\%)\\
            &  & $192 \times 192$ & 67.4\% (71.1\%) \\
            &  & $224 \times 224$ & 68.3\% (71.9\%) \\
            &  & $256 \times 256$ & 68.8\% (72.4\%) \\
            &  & $288 \times 288$ & \textbf{69.0\% (72.8\%)} \\
            \bottomrule[1.5pt]
        \end{tabular}
    }
    \caption{ViT-B/16 (Network parameters: 150M)}
    \label{tab:clip_vit_b16}
    \end{subtable}
    \vfill
    \begin{subtable}[b]{\columnwidth}
        \centering
    \resizebox{0.75\columnwidth}{!}{
        \begin{tabular}{cccc}
            \toprule[1.5pt]
            \textbf{Model} & \textbf{Dataset size}  & \textbf{Test resolution} & \textbf{Top-1 Accuracy (\%)} \\
            \midrule[1.25pt]
            OpenCLIP of \citet{ilharco_gabriel_2021_5143773} & 2.0B & $224 \times 224$ & 78.0\% \\ 
            \midrule
            \multirow{5}{*}{CLIP  w/ \method (Ours)} & \multirow{5}{*}{1.1B} & $160 \times 160$ & 76.1\% \\
            &  & $192 \times 192$ & 77.4\% \\
            &  & $224 \times 224$ & 77.9\% \\
            &  & $256 \times 256$ & 78.4\% \\
            &  & $288 \times 288$ & \textbf{78.6\%} \\
            \bottomrule[1.5pt]
        \end{tabular}
    }
    \caption{ViT-H/16 (Network parameters: 756M)}
    \label{tab:clip_vit_h16}
    \end{subtable}
    \caption{\textbf{\method learns augmentation efficiently for foundation models at scale}. With \method, we (a) improve accuracy by 1.2\% for similar dataset size and (b) achieve similar accuracy with approx. $2\times$ fewer data samples as compared to OpenCLIP. Each entry of CLIP with \method is the same model, but evaluated at different resolutions. Results within parenthesis in (a) are for CLIP ViT-B/16 when trained on 1.1B image-text pairs.}
    \label{tab:zero_shot_clip}
\end{table}

\subsection{Learning Augmentation for Student Model during Distillation}
\label{ssec:distillation}
Besides empirical risk minimization (ERM), another popular approach for training accurate mobile models is knowledge distillation (KD). In a standard distillation set-up, student and teacher receives the same input augmentations. Also, because of low complexity of mobile models, the augmentation strength is less. Because teachers are trained with extensive augmentations, \emph{it remains an open question to find an optimal range of magnitudes for students}. This section shows that \method can be used to learn the magnitude range for student model.

\paragraph{Dataset and teacher details} We distill knowledge from ResNet-101 to mobile models on the task of image classification on the ImageNet dataset. We use the same hyper-parameters during distillation as the ERM models.

\begin{figure}[t!]
    \centering
    \begin{subfigure}[b]{0.49\columnwidth}
        \centering
        \includegraphics[width=0.7\columnwidth]{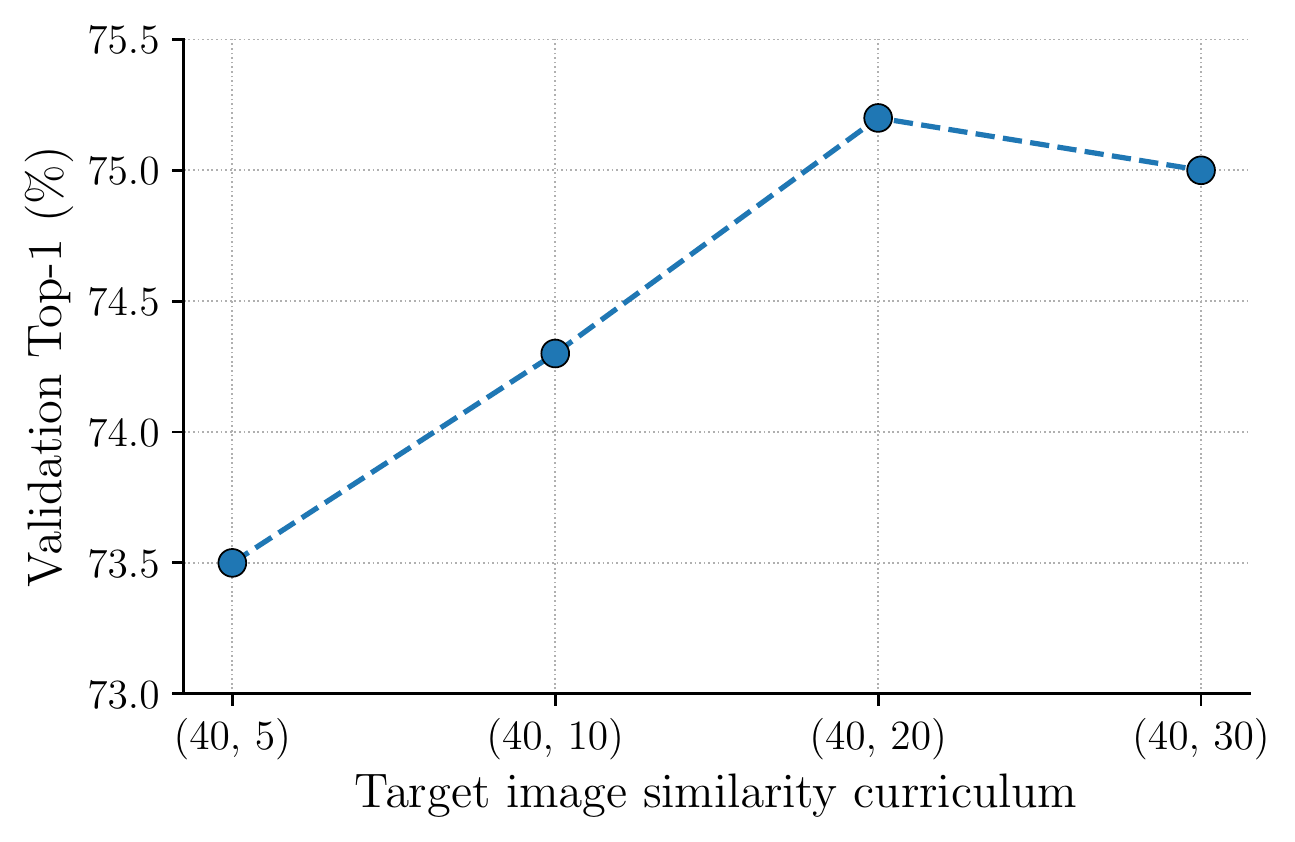}
        \caption{}
        \label{fig:distill_finding_delta}
    \end{subfigure}
    \hfill
    \begin{subfigure}[b]{0.49\columnwidth}
        \centering
        \resizebox{0.5\columnwidth}{!}{
            \begin{tabular}{lcc}
                \toprule[1.5pt]
                \textbf{Model} & \multicolumn{2}{c}{\textbf{Top-1 Accuracy}} \\
                \cmidrule[1.25pt]{2-3}
                 & \textbf{ERM} & \textbf{KD} \\
                 \midrule[1.25pt]
                 MobileNetv1 & 73.8\% & \textbf{75.2\%} \\
                 MobileNetv2 & 73.0\% & \textbf{73.4\%} \\
                 MobileNetv3 & 75.1\% & \textbf{76.0\%} \\
                 MobileViT & 78.2\% & \textbf{79.4\%} \\
                 \bottomrule[1.5pt]
            \end{tabular}
        }
        \caption{}
        \label{fig:distill_res}
    \end{subfigure}
    \begin{subfigure}[b]{\columnwidth}
        \centering
        \resizebox{0.85\columnwidth}{!}{
            \begin{tabular}{cccc}
                \toprule[1.5pt]
                \textbf{$\Delta$} & \textbf{Brightness} & \textbf{Contrast} & \textbf{Noise} \\
                \midrule[1.25pt]
                \multicolumn{4}{c}{MobileNetv1 trained using ERM (ImageNet Top-1: 73.8\%) } \\
                \midrule
                $40 \rightarrow 30$ &
                \raisebox{-0.5\totalheight}{\includegraphics[width=0.3\columnwidth]{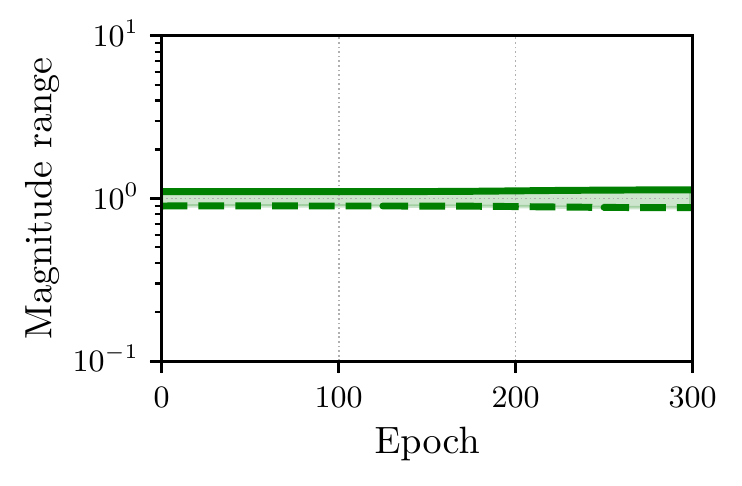}} & \raisebox{-0.5\totalheight}{\includegraphics[width=0.3\columnwidth]{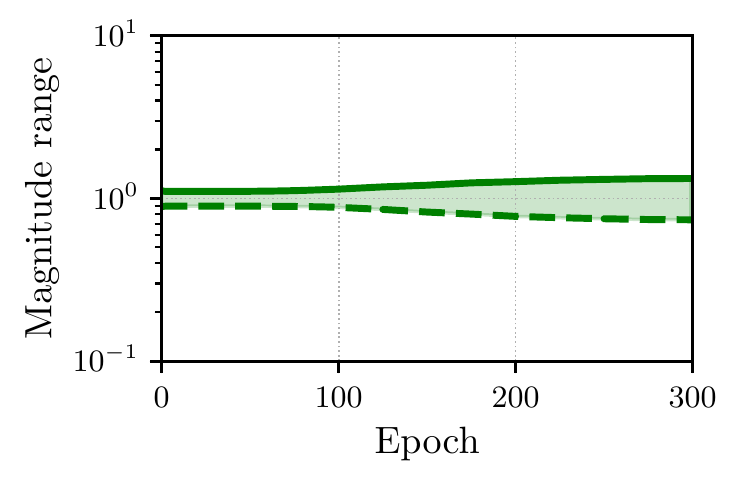}}
                 & \raisebox{-0.5\totalheight}{\includegraphics[width=0.3\columnwidth]{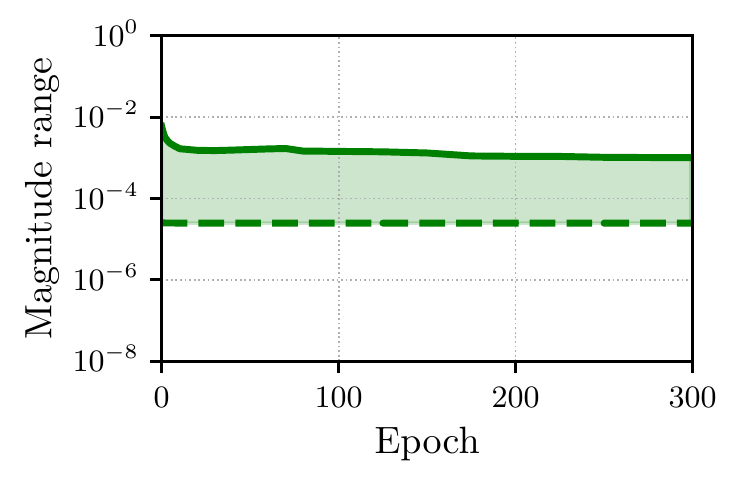}} \\
                 \midrule[1.25pt]
                 \multicolumn{4}{c}{MobileNetv1 trained using KD (ImageNet Top-1: 75.2\%) } \\
                 \midrule
                 $40 \rightarrow 20$ & \raisebox{-0.5\totalheight}{\includegraphics[width=0.3\columnwidth]{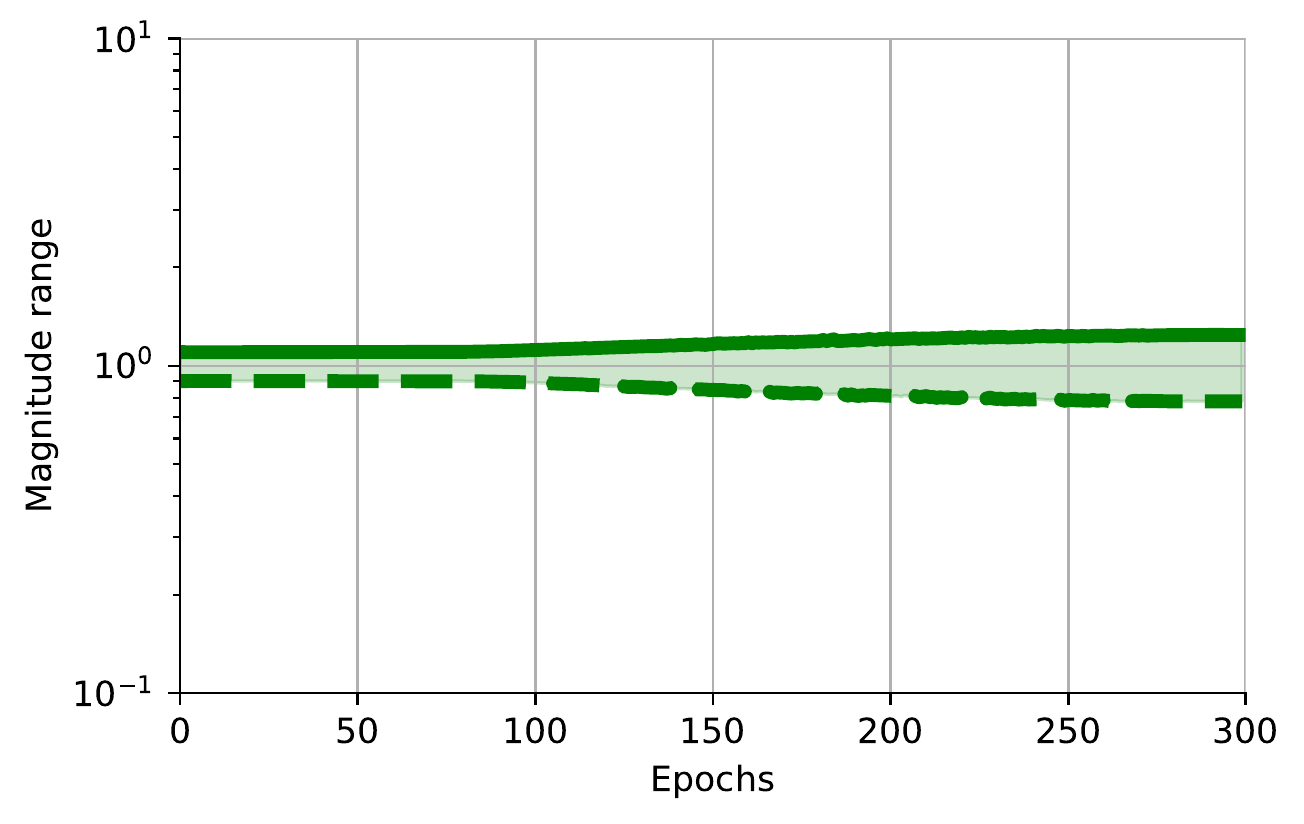}} & \raisebox{-0.5\totalheight}{\includegraphics[width=0.3\columnwidth]{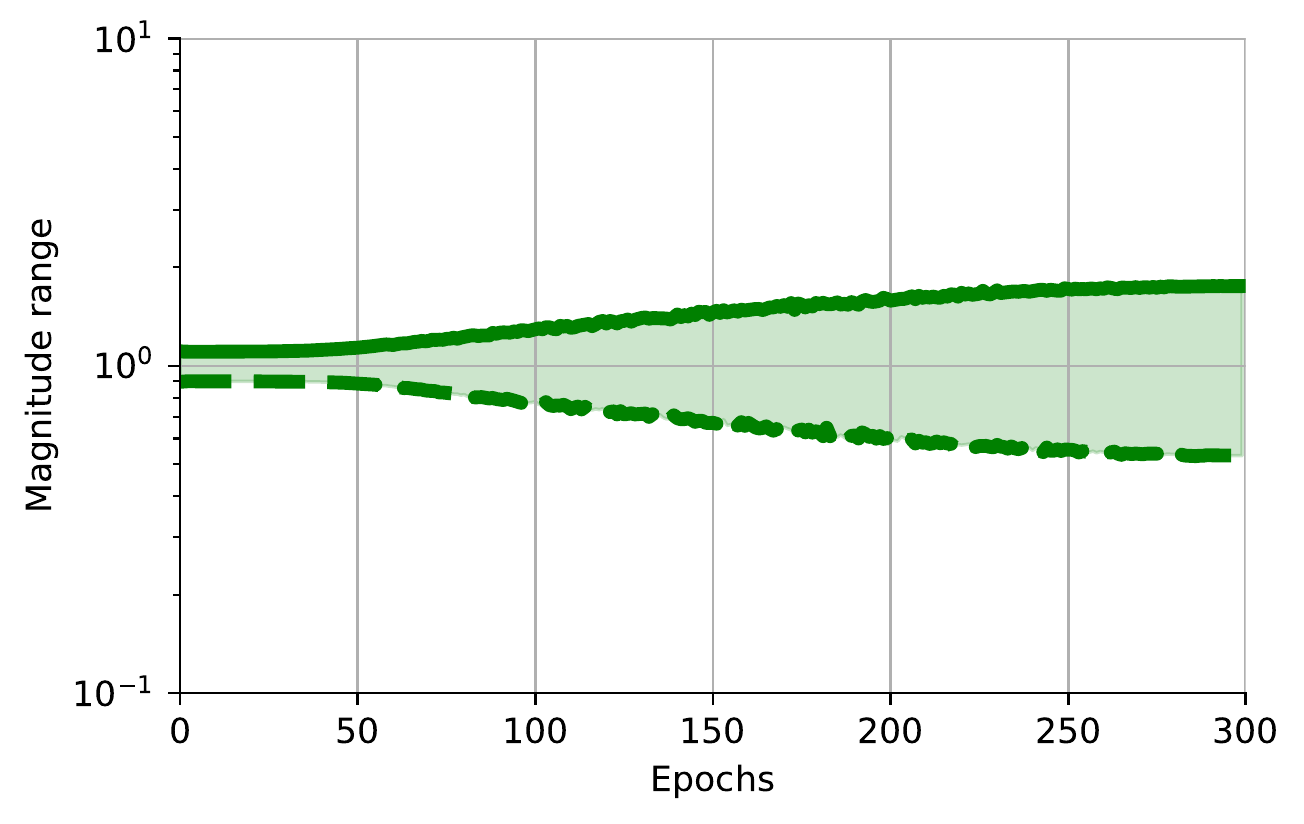}}
                 & \raisebox{-0.5\totalheight}{\includegraphics[width=0.3\columnwidth]{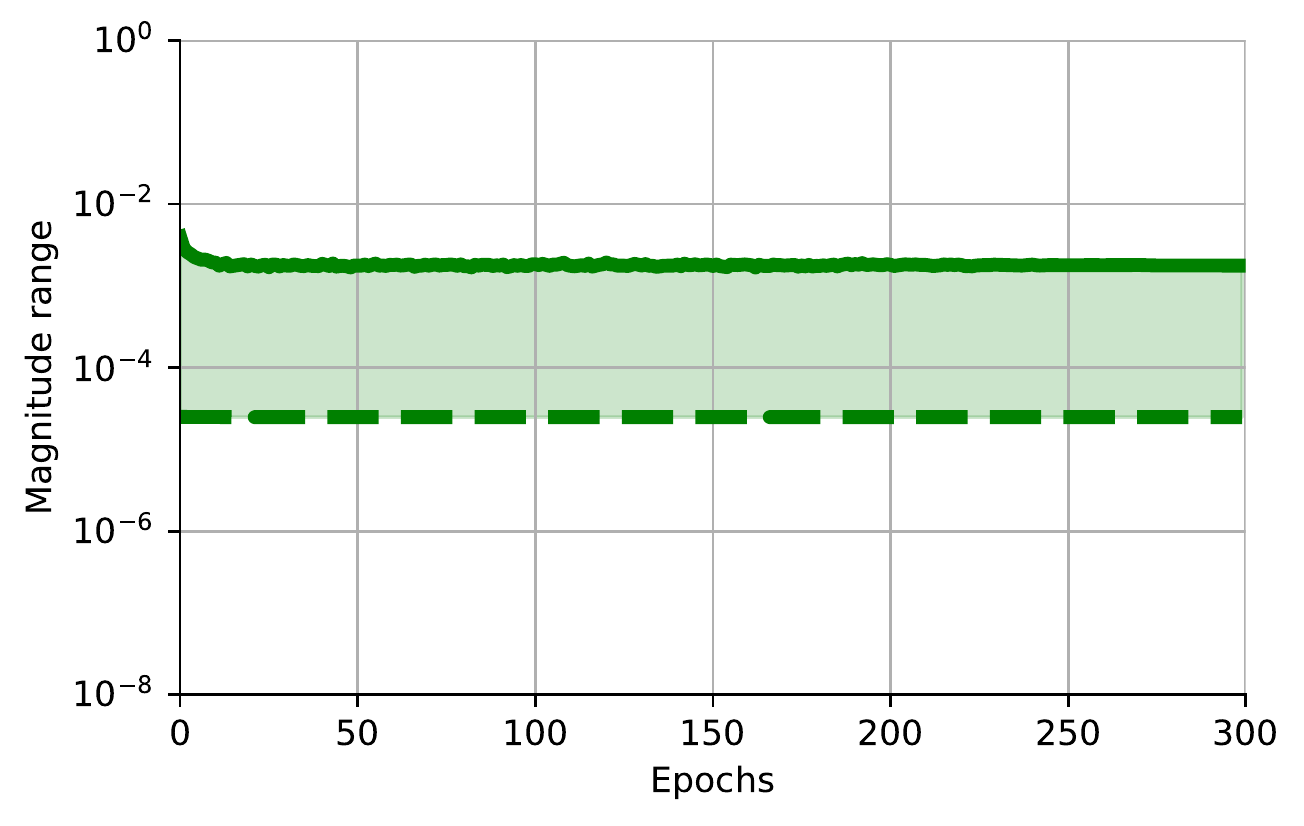}} \\
                \bottomrule[1.5pt]
            \end{tabular}
        }
        \caption{}
        \label{fig:distill_policy}
    \end{subfigure}
    \caption{\textbf{\method learns the magnitude range of augmentation operations for the task of distillation.} In (a), we search for the curriculum for target image similarity value. Here, student is MobileNetv1 and teacher is ResNet-101. In (b), we distill knowledge from ResNet-101 to different mobile backbones. We use the same curriculum values that gives the best accuracy in (a); demonstrating the curriculum is transferable. In (c), we visualize the magnitude range of different augmentation operations for the MobileNetv1 model trained with ERM and KD. During distillation, MobileNetv1 preferred wider magnitude ranges as compared to ERM.}
    \label{fig:my_label}
\end{figure}

\paragraph{Searching for the curriculum} We distill knowledge from ResNet-101 to MobileNetv1 on the ImageNet dataset at different target image similarity values, $\Delta$. \cref{fig:distill_finding_delta} shows that MobileNetv1 achieves the best performance when we anneal $\Delta$ from 40 to 20. This curriculum is different from our observations in \cref{tab:compare_sota_imagenet_models}, where we found that MobileNetv1 trained with ERM achieve best performance when we anneal $\Delta$ from 40 to 30. Moreover, \cref{fig:distill_policy} shows that MobileNetv1 prefers wider magnitude ranges during KD as compared to ERM. 

\paragraph{\method's curriculum is transferable for other mobile models} To see if the curriculum found for MobileNetv1 for KD is transferrable to other mobile backbones, we distilled knowledge from ResNet-101 to other backbones. During distillation, we annealed the $\Delta$ from 40 to 20. 
\cref{fig:distill_res} shows that KD with \method improves the performance of different models.

\begin{tcolorbox}[width=\columnwidth,
                  colback=green!5!white,
                  colframe=green!75!black
                  ]
\small{
\textbf{Observation 3:} The range of magnitudes for each augmentation operation varies across models and tasks. \method allows us to learn model- and task-specific magnitude range of each augmentation operation efficiently in an end-to-end manner.
}
\end{tcolorbox}

\section{Conclusion}
This paper introduces an end-to-end method for learning model- and task-specific automatic augmentation policies with a constant search time. We demonstrated that  \method delivers competitive performance to existing methods across different downstream models and tasks. This is despite the fact that \method uses only three basic augmentation operations as opposed to a large set of complex augmentation operations in existing methods. These results underline the importance of magnitude range of augmentation operations in automatic augmentation. We also showed that \method can be seamlessly integrated with other tasks and achieve similar or better performance than existing methods. 

\paragraph{Future work} We plan to apply \method to learn the range of magnitudes for complex augmentation operations (e.g., geometric transformations) using different image similarity functions (e.g., SSIM). In addition to learning the range of magnitudes of each augmentation operation, we plan to apply \method to learn how to compose different augmentation operations with a constant search time.

\section*{Acknowledgements}
We are grateful to Mehrdad Farajtabar, Hadi Pour Ansari, Farzad Abdolhosseini, and members of Apple's machine intelligence and neural design team for their helpful comments. We are also thankful to Apple's infrastructure and open-source teams for their help with training infrastructure and open-source release of the code and pre-trained models.

\appendix

\section{Ablations on the ImageNet dataset}
\label{sec:appendix_ablations}

In this section, we study different components of \method using ResNet-50. For learning augmentation policy, we anneal the target image similarity (PSNR) value $\Delta$ from 40 to 5.

\paragraph{Effect of different curriculum} We trained \method with two curriculum's: (1) linear and (2) cosine. We found that cosine curriculum delivers 0.1-0.2\% better performance than linear. Therefore, we use cosine curriculum. 

\begin{figure}[!ht]
    \centering
    \includegraphics[width=0.5\textwidth]{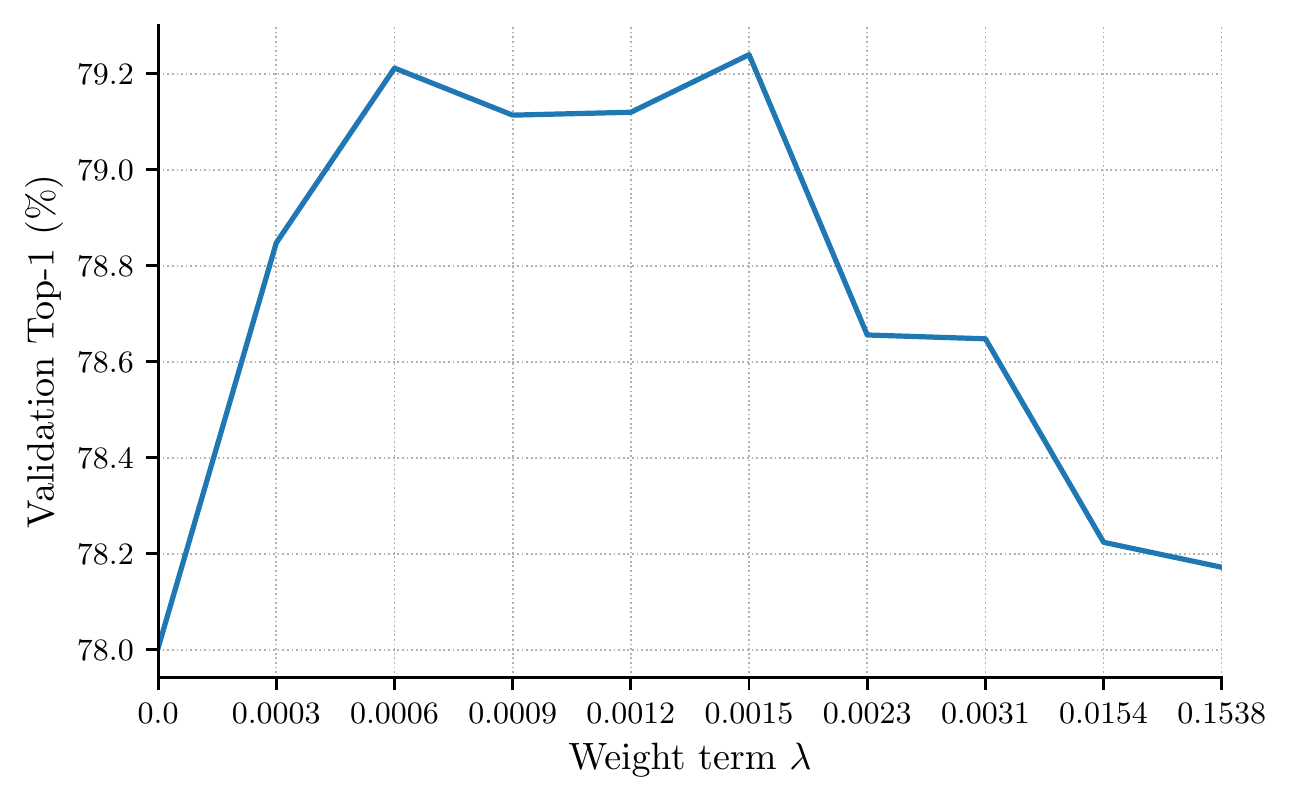}
    \caption{\textbf{Effect of weight term, $\lambda$, on ResNet-50's performance on the ImageNet dataset.}}
    \label{fig:ablation-lambda}
\end{figure}

\paragraph{Effect of $\lambda$} The weight term, $\lambda$, in \cref{eq:optim_objective} allows \method to balance the trade-off between augmentation loss $\Ls_{\text{ra}}$ and empirical loss $\mathcal{L}_{\text{task}}$. To study its impact, we vary the value of $\lambda$ from 0.0 to 0.15. Empirical results in \cref{fig:ablation-lambda} shows that the good range for $\lambda$  is between 0.0006 and $0.002$. In our experiments, we use $\lambda=0.0015$.

\begin{figure}[b!]
    \centering
        \resizebox{0.8\columnwidth}{!}{
            \begin{tabular}{cccc}
                \toprule[1.5pt]
                \textbf{$\Delta$} & \textbf{Brightness} & \textbf{Contrast} & \textbf{Noise} \\
                \midrule[1.25pt]
                $40 \rightarrow 5$ & \raisebox{-0.5\totalheight}{\includegraphics[width=0.3\columnwidth]{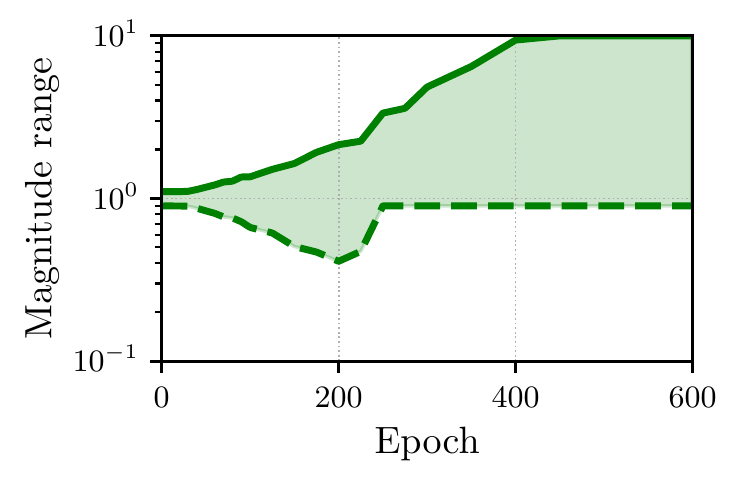}}
                & \raisebox{-0.5\totalheight}{\includegraphics[width=0.3\columnwidth]{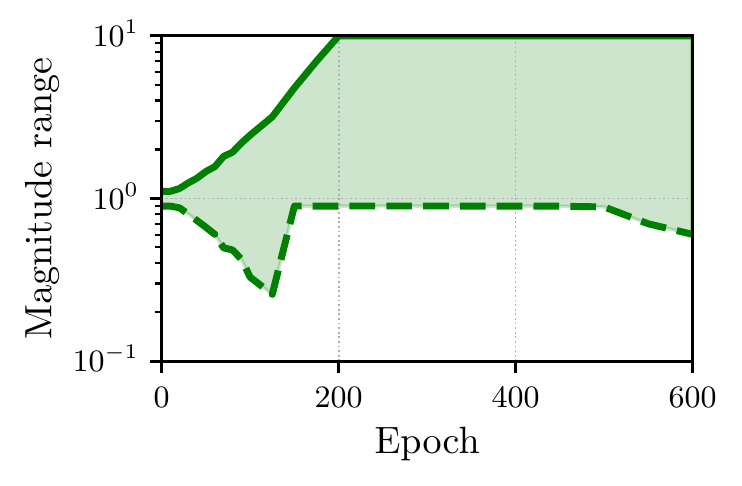}}
                & \raisebox{-0.5\totalheight}{\includegraphics[width=0.3\columnwidth]{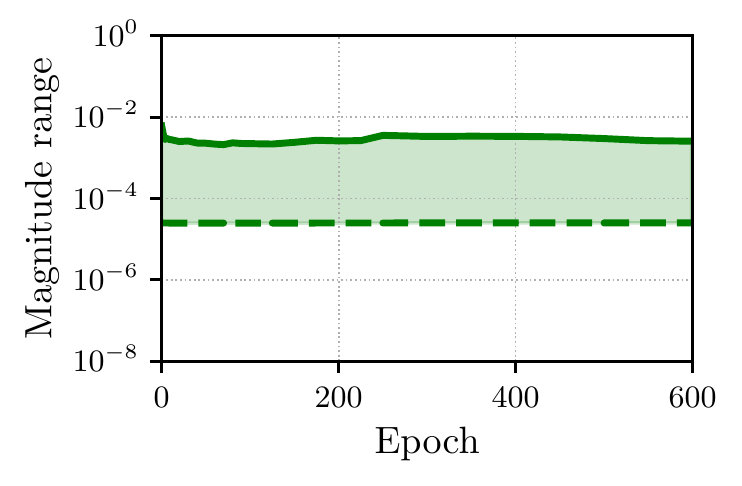}} \\
                \multicolumn{4}{c}{Joint loss} \\
                \midrule
                $40 \rightarrow 5$ & \raisebox{-0.5\totalheight}{\includegraphics[width=0.3\columnwidth]{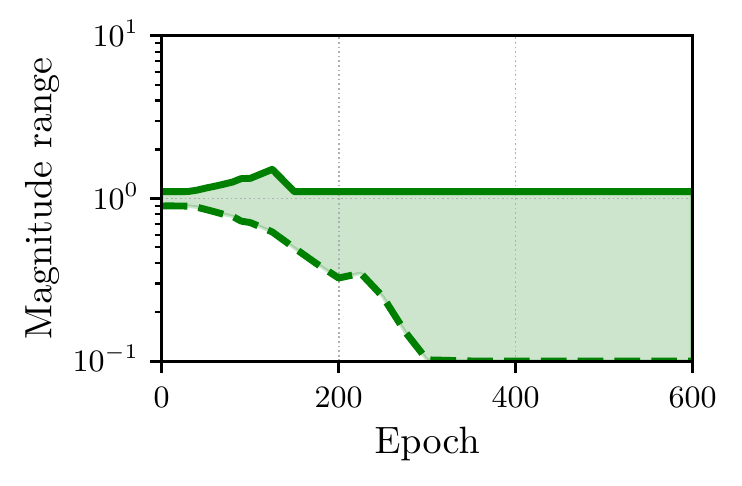}}
                & \raisebox{-0.5\totalheight}{\includegraphics[width=0.3\columnwidth]{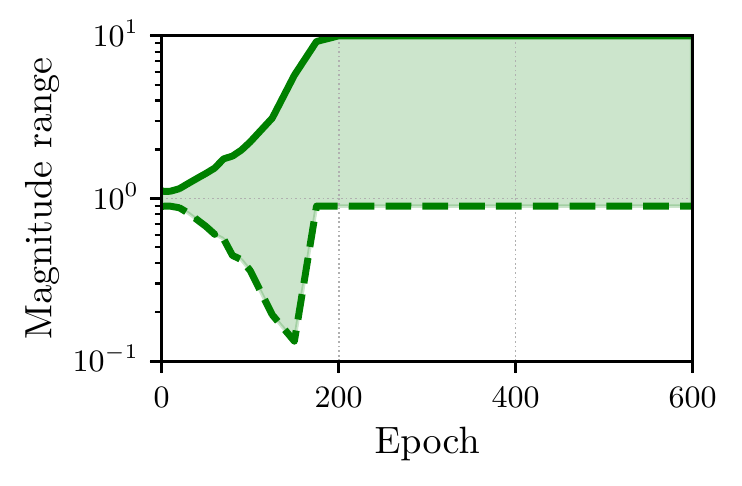}}
                & \raisebox{-0.5\totalheight}{\includegraphics[width=0.3\columnwidth]{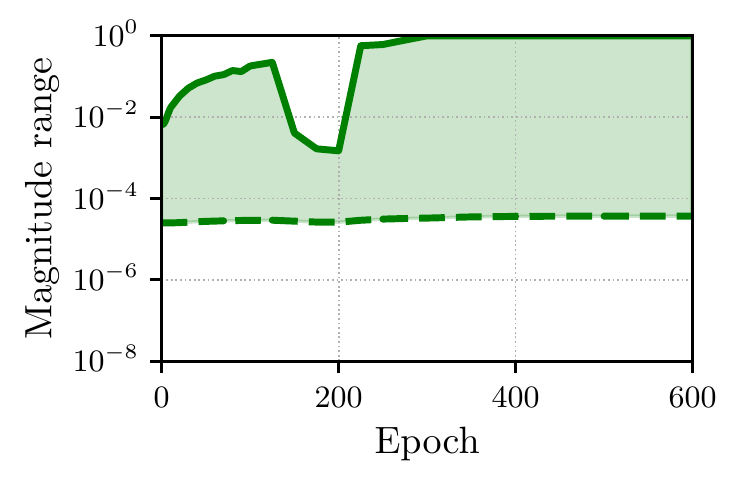}} \\
                \multicolumn{4}{c}{Independent loss} \\
                \bottomrule[1.5pt]
            \end{tabular}
        }
    \caption{\textbf{Joint vs. independent optimization}. The effect of learning magnitude ranges by jointly optimizing the loss terms $\Ls_{\text{ra}}$ and $\Ls_{\text{task}}$ (top row) compared to only optimizing $\Ls_{\text{ra}}$ (bottom row) is shown for ResNet-50 on the ImageNet dataset. Training ResNet-50 with the joint loss leads to smaller magnitudes of noise, and improves validation accuracy by approximately $1\%$ on the ImageNet dataset.}
    \label{fig:ablation-loss-joint-independent-ranges}
\end{figure}

\paragraph{Effect of joint vs. independent optimization} An expected behavior for learning model-specific augmentation policy using \method is that task-specific loss $\Ls_{\text{task}}$ in Eq. \ref{eq:optim_objective} should contribute towards policy learning. To validate it, ResNet-50 is trained independently\footnote{The augmented image is detached before feeding to the model.} as well as jointly on the ImageNet dataset. We found that the top-1 accuracy of ResNet-50 dropped by about 1\% when it is trained independently. This is likely because independent training allowed \method to produce augmented images with more additive Gaussian noise (\cref{fig:ablation-loss-joint-independent-ranges}), resulting in performance drop. This concurs with our observations in \cref{sec:imagenet_neural_aug}, especially \cref{fig:aug_effect_perf_char,fig:learned_range_r50}, where we found that sampling augmented images from wider magnitude range for additive Gaussian noise operation  dropped ResNet-50's performance on the ImageNet dataset. A plausible explanation is that PSNR is more sensitive to noise operation \citep{hore2010image}, allowing \method to learn wider magnitude ranges for noise operation when trained independently as compared to joint training. Overall, these results suggest that joint training helps in learning model-specific policy.

\section{Visualization of augmented samples}
\label{sec:appendix_sample_vis}
\begin{figure}[!ht]
    \centering
    \begin{subfigure}[b]{\textwidth}
        \centering
        \resizebox{0.8\columnwidth}{!}{
            \begin{tabular}{cccc}
                \toprule[1.5pt]
                \textbf{$\Delta$} & \textbf{Brightness} & \textbf{Contrast} & \textbf{Noise} \\
                \midrule[1.25pt]
                $40 \rightarrow 5$ & \raisebox{-0.5\totalheight}{\includegraphics[width=0.3\columnwidth]{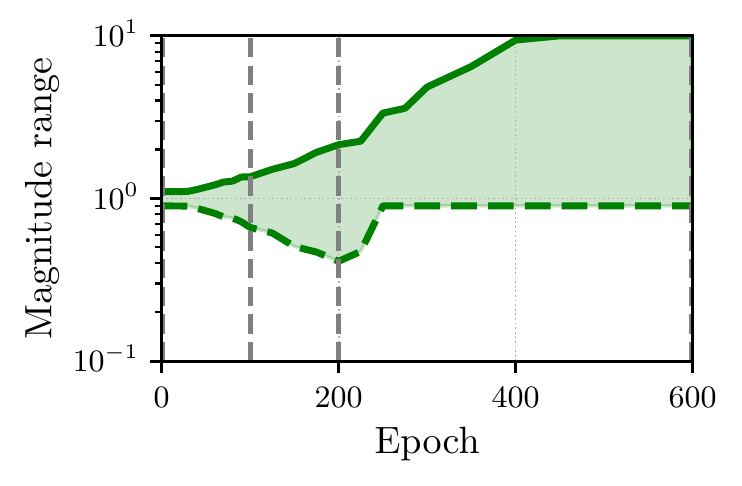}}
                & \raisebox{-0.5\totalheight}{\includegraphics[width=0.3\columnwidth]{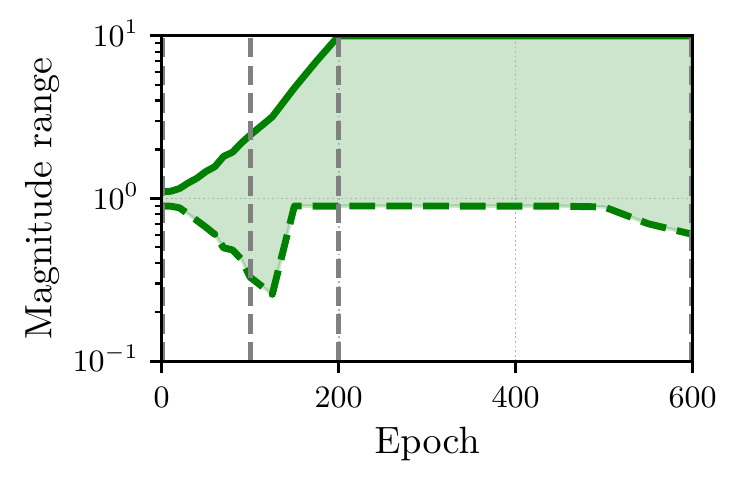}}
                & \raisebox{-0.5\totalheight}{\includegraphics[width=0.3\columnwidth]{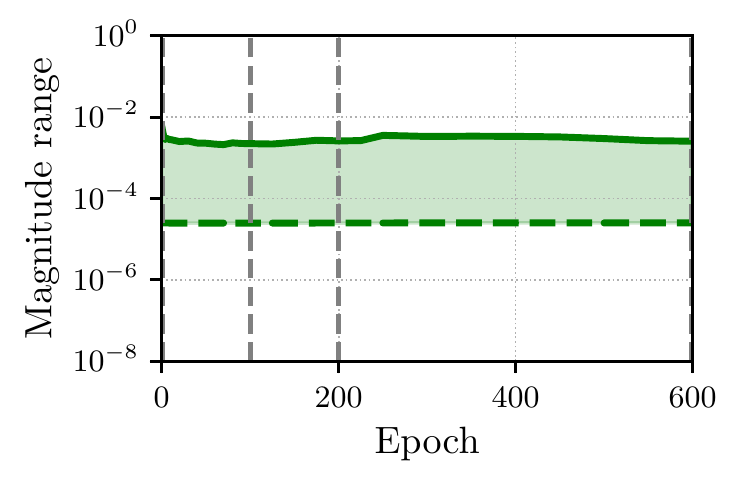}} \\
                \bottomrule[1.5pt]
            \end{tabular}
        }
        \caption{}
        \label{fig:augmentations-epochs-ranges}
        \vspace*{1em}
    \end{subfigure}
    \begin{subfigure}[b]{\textwidth}
        \centering
        \resizebox{0.9\columnwidth}{!}{
            \begin{tabular}{ccccc}
                \includegraphics[width=0.3\columnwidth]{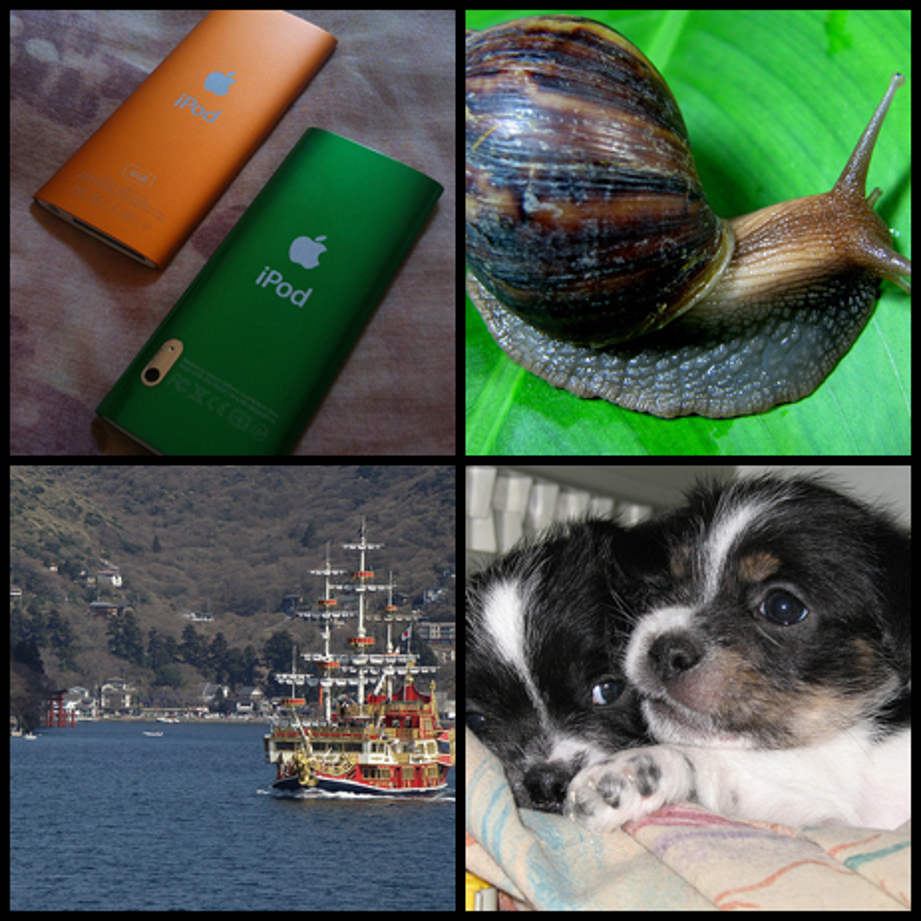}
                & \includegraphics[width=0.3\columnwidth]{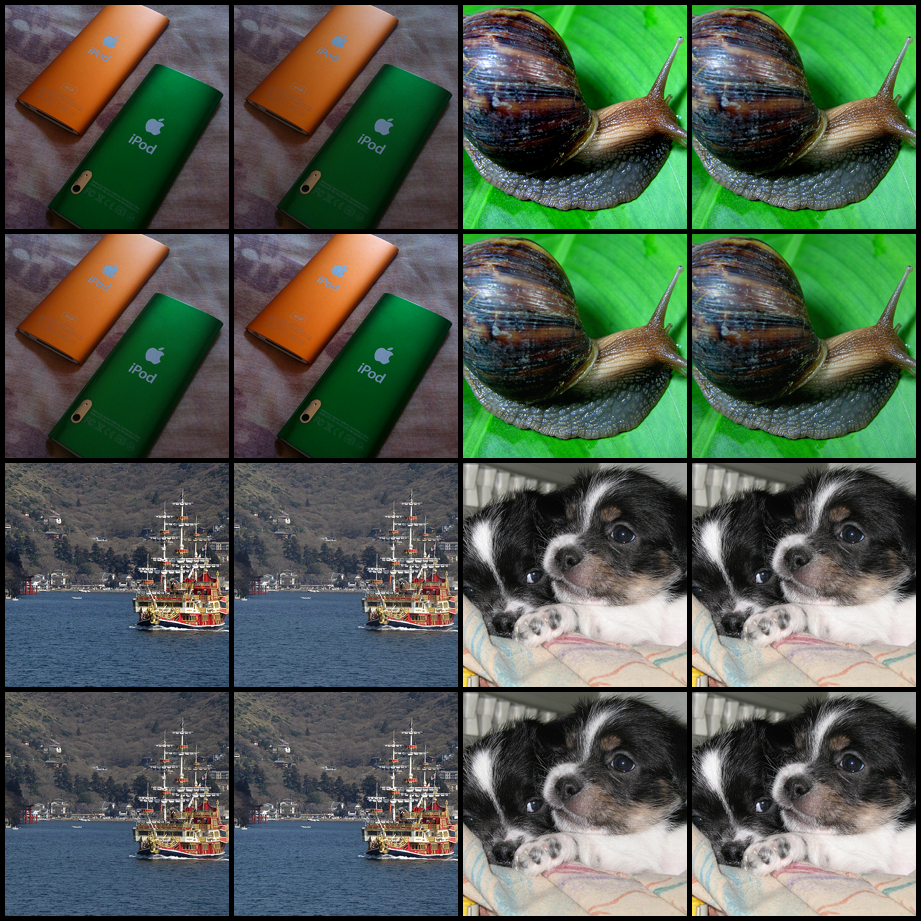}
                & \includegraphics[width=0.3\columnwidth]{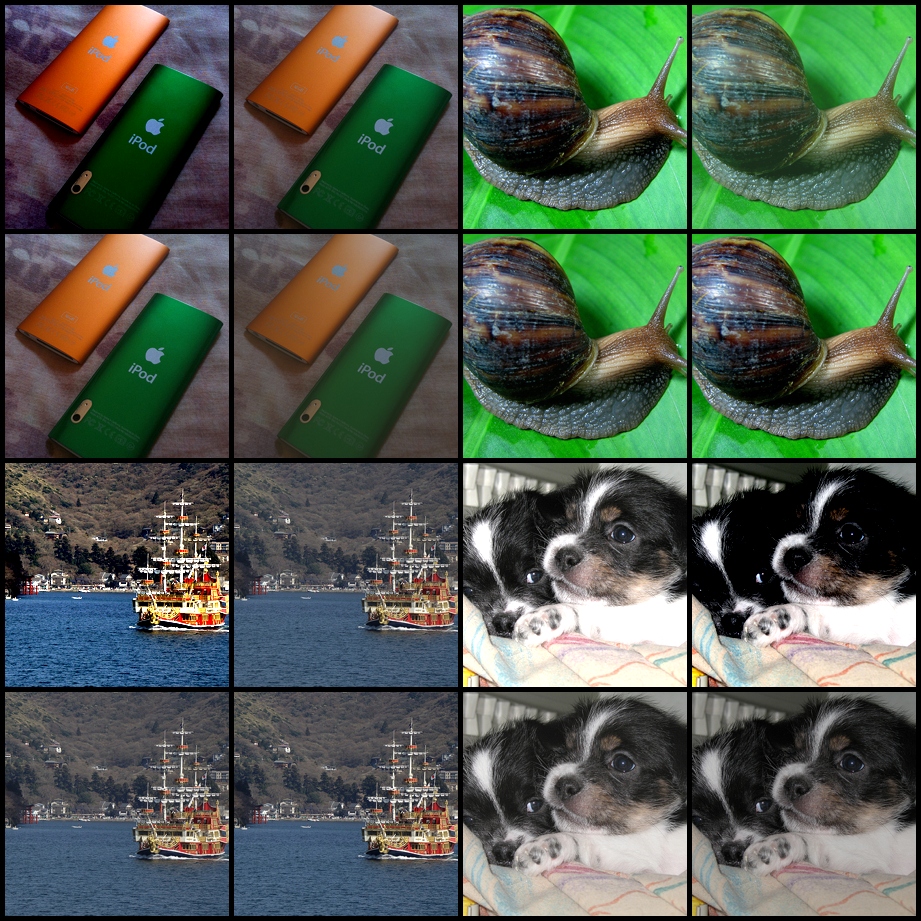}
                & \includegraphics[width=0.3\columnwidth]{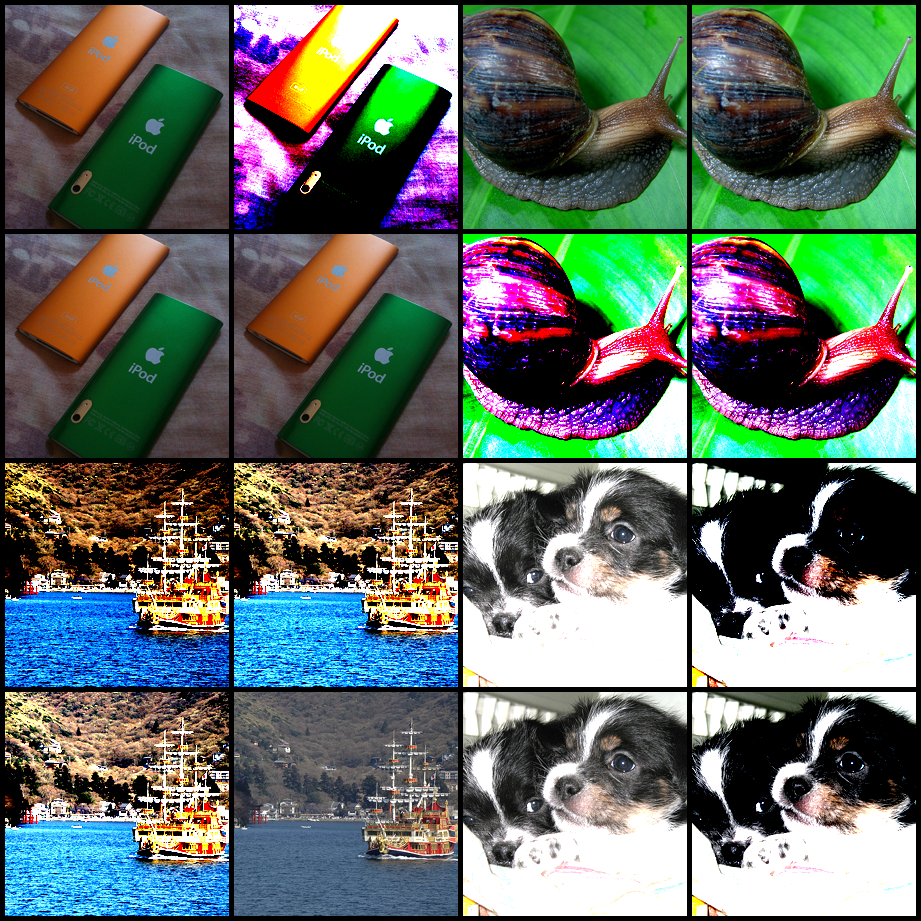}
                & \includegraphics[width=0.3\columnwidth]{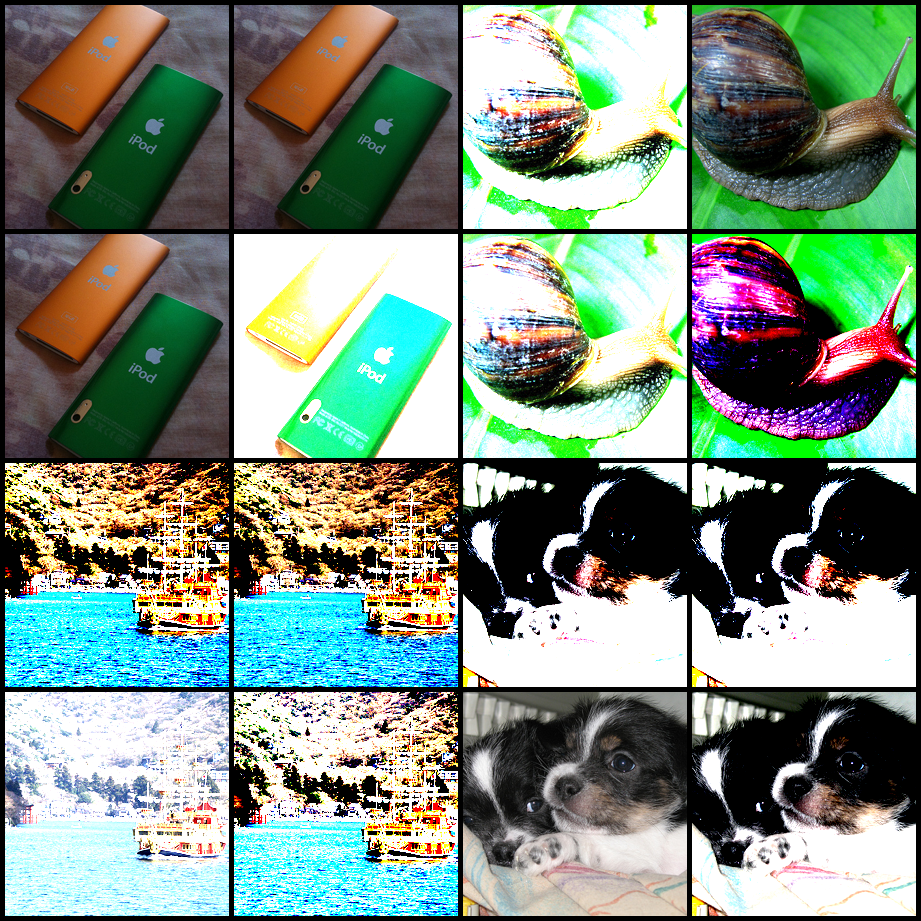} \\
                \textbf{Input images} & \textbf{Epoch 1} & \textbf{Epoch 100} & \textbf{Epoch 200} & \textbf{Epoch 600} \\
            \end{tabular}
        }
        \caption{}
        \label{fig:augmentations-epochs-images}
    \end{subfigure}
    \caption{\textbf{Visualization of augmented samples.} (a) Learned magnitude ranges of different augmentations when ResNet-50 is trained jointly with \method using a cosine curriculum. (b) Four image samples visualized at different epoch intervals. For illustration purposes, we visualize four random augmented samples produced by \method for each image. }
    \label{fig:augmentations_epochs}
\end{figure}

\cref{fig:augmentations_epochs} visualizes augmented samples produced by \method at different stages of training ResNet-50 with curriculum. We can see that the range of magnitudes (\cref{fig:augmentations-epochs-ranges}) and diversity of augmented samples (\cref{fig:augmentations-epochs-images}) increases as training progresses.

\section{Learned Magnitude Ranges for CLIP with \method}
\begin{figure}[!ht]
    \centering
    \resizebox{0.7\columnwidth}{!}{
        \begin{tabular}{ccc}
            \toprule[1.5pt]
            \textbf{Brightness} & \textbf{Contrast} & \textbf{Noise} \\
            \midrule[1.25pt]
            \raisebox{-0.5\totalheight}{\includegraphics[width=0.3\columnwidth]{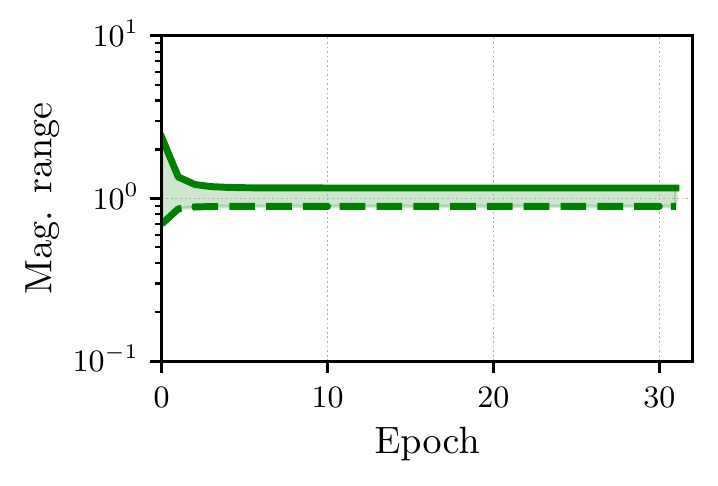}}
            & \raisebox{-0.5\totalheight}{\includegraphics[width=0.3\columnwidth]{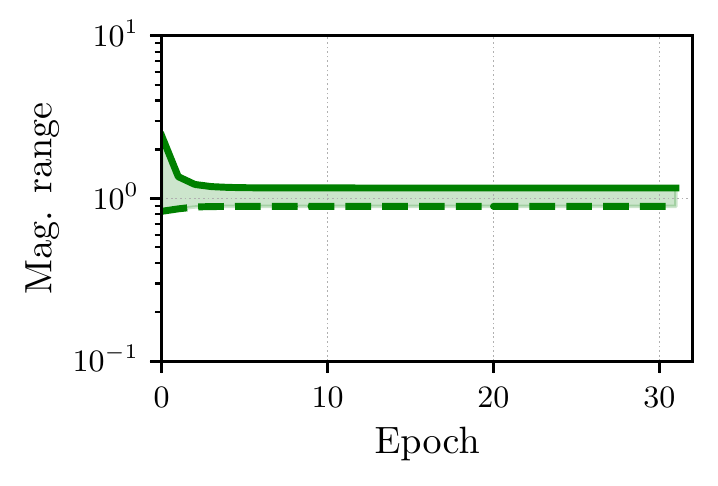}}
            & \raisebox{-0.5\totalheight}{\includegraphics[width=0.3\columnwidth]{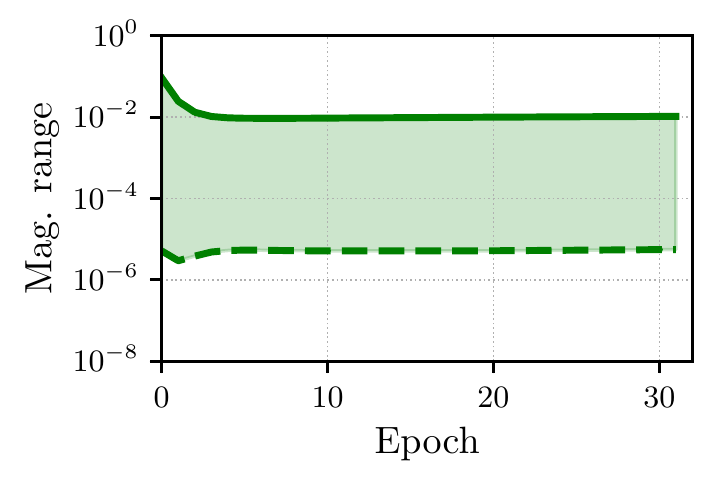}} \\
            \bottomrule[1.5pt]
        \end{tabular}
    }
    \caption{\textbf{Learned magnitude ranges of different augmentations for CLIP with ViT-B/16}. Target PSNR $\Delta$ is annealed from 40 to 20.}
    \label{fig:learned-ranges-clip}
\end{figure}

\cref{fig:learned-ranges-clip} shows the learned magnitude ranges of different augmentation operations for training the CLIP model with \method. Unlike image classification (\cref{sec:imagenet_neural_aug}) and semantic segmentation (\cref{ssec:semantic_seg}) results, the CLIP model uses little augmentation. This is expected because image-text pair datasets are orders of magnitude larger than classification and segmentation datasets.

\section{Training Details}
\subsection{Training Hyper-parameters}
\label{sec:appendix_training_details}
\cref{tab:task_training_procedure} summarizes training recipe used for training different models across different tasks.

\begin{table}[!ht]
    \centering
    \begin{subtable}[b]{\columnwidth}
        \resizebox{\columnwidth}{!}{
        \begin{tabular}{lccccccccc}
            \toprule[1.5pt]
             & & \multicolumn{2}{c}{\bfseries Non-mobile models} & && & \multicolumn{2}{c}{\bfseries Mobile models} & \\
             \cmidrule[1pt]{2-5} \cmidrule[1pt]{7-10}
             &  \textbf{ResNet-50} & \textbf{ResNet-101} & \textbf{EfficientNet} & \textbf{SwinTransformer} && \textbf{MobileViT} & \textbf{MobileNetv1} & \textbf{MobileNetv2} & \textbf{MobileNetv3} \\
             \midrule[1pt]
             Epochs & 600 & 600 & 400 & 300 && 300 & 300 & 300 & 300 \\
             Batch size & 1024 & 1024 & 2048 & 1024 && 1024 & 512 & 1024 & 2048 \\
             Data sampler & MSc-VBS & MSc-VBS & MSc-VBS & SSc-FBS && MSc-VBS & MSc-VBS & MSc-VBS & MSc-VBS \\
             \midrule
             Max. LR & $0.4$ & $0.4$ & $0.8$ & $10^{-3}$ && $2 \times 10^{-3}$ & $0.4$ & $0.4$ & $0.8$ \\
             Min. LR & $2 \times 10^{-4}$ & $2 \times 10^{-4}$ & $4 \times 10^{-4}$ & $10^{-5}$ && $2 \times 10^{-4}$ & $2 \times 10^{-4}$ & $2 \times 10^{-4}$ & $4 \times 10^{-4}$ \\
             Warmup init. LR & 0.05 & 0.05 & 0.1 & $10^{-6}$ && $2 \times 10^{-4}$ & $0.05$ & $0.05$ & $0.1$ \\
             Warmup epochs & 5 & 5 & 5 & 20 && 16 & 3 & 6 & 5 \\
             LR Annealing & cosine & cosine & cosine & cosine && cosine & cosine & cosine & cosine \\
             Weight decay & $4 \times 10^{-5}$ & $4 \times 10^{-5}$ & $4 \times 10^{-5}$ & $5 \times 10^{-2}$ && 0.01 & $4 \times 10^{-5}$ & $4 \times 10^{-5}$ & $4 \times 10^{-5}$ \\
             Optimizer & SGD & SGD & SGD & AdamW && AdamW & SGD & SGD & SGD \\
             Momentum & 0.9 & 0.9 & 0.9 & \xmarkApp && \xmarkApp & 0.9 & 0.9 & 0.9\\
             \midrule
             Label smoothing $\epsilon$ & 0.1 & 0.1 & 0.1 & 0.1 && 0.1 & 0.1 & 0.1 & 0.1 \\
             Stoch. Depth & \xmarkApp & \xmarkApp & \xmarkApp & 0.3 && \xmarkApp & \xmarkApp & \xmarkApp & \xmarkApp \\
             Grad. clipping & \xmarkApp & \xmarkApp & \xmarkApp & 5.0 && \xmarkApp & \xmarkApp & \xmarkApp & \xmarkApp \\
             \midrule[1.25pt]
             \# parameters & 25.6 M & 44.5 M & 12.3 M & 49.6 M && 5.6 M & 4.2 M & 3.5 M  & 5.4 M \\
             \# FLOPs & 4.0 G & 7.7 G & 1.9 G & 8.8 G && 2.0 G & 579 M & 314 M & 220 M \\
            \bottomrule[1.5pt]
        \end{tabular}
    }
    \caption{Image classification on ImageNet}
    \label{tab:imagenet_training_procedure}
    \end{subtable}
    \vfill
    \begin{subtable}[b]{0.5\columnwidth}
        \centering
        \resizebox{0.9\columnwidth}{!}{
        \centering
        \begin{tabular}{lcc}
            \toprule[1.5pt]
            \textbf{Hyperparameter} & \textbf{ADE20k} & \textbf{PASCAL VOC} \\
            \midrule
             Epochs & 50 & 50 \\
             Batch size & 16 & 128 \\
             Data sampler & SSc-FBS & SSc-FBS \\ 
             Max. LR & $0.02$ & 0.02\\
             Min. LR & $10^{-4}$ & $10^{-4}$\\
             LR Annealing & cosine & cosine \\
             Weight decay & $10^{-4}$ & $10^{-4}$ \\
             Optimizer & SGD & SGD \\
            \bottomrule[1.5pt]
        \end{tabular}
        }
        \caption{Semantic segmentation with DeepLabv3}
        \label{tab:segmentation_training_procedure}
    \end{subtable}
    \hfill
    \begin{subtable}[b]{0.45\columnwidth}
        \centering
        \resizebox{0.6\columnwidth}{!}{
        \centering
        \begin{tabular}{lc}
            \toprule[1.5pt]
            \textbf{Hyperparameter} & \textbf{COCO} \\
            \midrule
             Epochs & 100 \\
             Batch size & 32 \\
             Data sampler & MSc-VBS \\
             Warm-up iterations & 500 \\
             Warm-up init. LR & 0.001 \\
             LR & 0.1 \\
             LR annealing & Multi-step \\
             LR annealing factor & 0.1 \\
             Annealing epochs & [60, 84 ] \\
             Weight decay & 0.2 \\
             Optimizer & SGD \\
            \bottomrule[1.5pt]
        \end{tabular}
        }
        \caption{Object detection with Mask R-CNN}
        \label{tab:od_train_procedure}
    \end{subtable}
    \vfill
    \begin{subtable}[b]{\columnwidth}
        \centering
        \resizebox{0.7\columnwidth}{!}{
        \centering
        \begin{tabular}{lccc}
            \toprule[1.5pt]
            \textbf{Hyperparameter} & \textbf{CLIP-VIT-B/16} & \textbf{CLIP-VIT-B/16} & \textbf{CLIP-VIT-H/16}\\
            \midrule
            \# image-text pairs & 302M & 1.1B & 1.1B \\
             Epochs/iterations & 32 & 150k & 120k \\
             Batch size & 32k & 65k & 132k \\
             Data sampler & MSc-VBS & MSc-VBS & MSc-VBS \\
             Warm-up iterations & 2000 & 1000 & 1000 \\
             Warm-up init. LR & $10^{-6}$ & $10^{-6}$ & $10^{-6}$\\
             Max. LR & $5^{-4}$ & $5^{-4}$ & $5^{-4}$ \\
             Min. LR & $10^{-6}$ & $10^{-6}$ & $10^{-6}$\\
             LR Annealing & cosine & cosine & cosine \\
             Weight decay & 0.2 & 0.2 & 0.2 \\
             Optimizer & AdamW & AdamW & AdamW\\
            \bottomrule[1.5pt]
        \end{tabular}
        }
        \caption{CLIP models}
        \label{tab:clip_training_procedure}
    \end{subtable}
    \caption{\textbf{Hyper-parameters used for training models on different tasks.} Here, SSc-FBS and MSc-VBS refers to single-scale fixed batch size and multi-scale variable batch size data samplers respectively \citep{mehta2022cvnets}. We found that the performance of CLIP models with smaller (32k) and larger (65k or 132k) batch sizes is similar. We used large batch sizes for faster training.}
    \label{tab:task_training_procedure}
\end{table}

\subsection{Bounds for Augmentation operations}
\label{sec:app_bounds}
\cref{tab:app_clip_bounds} shows the clipping bounds that \method uses to prevent training instability.

\begin{table}[!ht]
    \centering
    \resizebox{0.75\columnwidth}{!}{
    \begin{tabular}{lccc}
        \toprule[1.5pt]
        \multirow{2}{*}{\textbf{Operation}} & \multicolumn{2}{c}{\textbf{Clipping bounds in \method}} & \textbf{Magnitude range in AutoAugment}\\
        \cmidrule[1pt]{2-3}
        & \textbf{\hspace{20pt} Min. \hspace{20pt}} & \textbf{\hspace{20pt} Max. \hspace{20pt}} & \textbf{(reference)}\\
        \midrule[1pt]
        Brightness & 0.1 & 10.0 & [0.1 - 1.9] \\
        Contrast & 0.1 & 10.0 & [0.1 - 1.9] \\
        Noise (std. dev) & 0 & 1.0 & --\\
        \bottomrule[1.5pt]
    \end{tabular}
    }
    \caption{\textbf{Clipping bounds used in \method for preventing training instability.} The range of magnitudes for augmentation operations  in AutoAugment is also given as a reference.}
    \label{tab:app_clip_bounds}
\end{table}

\bibliography{iclr2023_conference}
\bibliographystyle{iclr2023_conference}

\end{document}